\documentclass[10pt,twocolumn,letterpaper]{article}

\usepackage{iccv}
\usepackage{times}
\usepackage{epsfig}
\usepackage{graphicx}
\usepackage{amsmath}
\usepackage{amssymb}
\usepackage[dvipsnames]{xcolor}
\usepackage[utf8]{inputenc} %
\usepackage[T1]{fontenc}    %
\usepackage{url}            %
\usepackage{booktabs}       %
\usepackage{amsfonts}       %
\usepackage{nicefrac}       %
\usepackage{microtype}      %
\usepackage{multirow}
\usepackage{bbm}
\usepackage{rotating}

\usepackage[pagebackref=true,breaklinks=true,letterpaper=true,colorlinks,bookmarks=false]{hyperref}

\DeclareMathOperator*{\argmin}{arg\,min}

\definecolor{green}{RGB}{0,150,10}

\newcommand*{\pd}[3][]{\ensuremath{\frac{\partial^{#1} #2}{\partial #3}}}

\newcommand{\figLabel}{Fig.~}
\newcommand{\eqLabel}[1]{{Eq (#1)}}
\newcommand{\secLabel}{Section~}

\newcommand{\mysection}[1]{\noindent\textbf{#1.}}
\newcommand{\supp}{\textbf{Appendix\xspace} }
\newcommand{\specialcell}[2][c]{%
  \begin{tabular}[#1]{@{}c@{}}#2\end{tabular}}

\iccvfinalcopy %

\def\iccvPaperID{1133} %

\ificcvfinal\fi

\begin{document}

\title{MVTN: Multi-View Transformation Network for 3D Shape Recognition}
\author{Abdullah Hamdi \quad\quad Silvio Giancola \quad\quad Bernard Ghanem\\  
King Abdullah University of Science and Technology (KAUST), Thuwal, Saudi Arabia\\
\small{\{abdullah.hamdi, silvio.giancola, bernard.ghanem\}@kaust.edu.sa}
}

\maketitle
\ificcvfinal\thispagestyle{empty}\fi

\begin{abstract}
Multi-view projection methods have demonstrated their ability to reach state-of-the-art performance on 3D shape recognition. Those methods learn different ways to aggregate information from multiple views. However, the camera view-points for those views tend to be heuristically set and fixed for all shapes. To circumvent the lack of dynamism of current multi-view methods, we propose to learn those view-points. In particular, we introduce the Multi-View Transformation Network (MVTN) that regresses optimal view-points for 3D shape recognition, building upon advances in differentiable rendering. As a result, MVTN can be trained end-to-end along with any multi-view network for 3D shape classification. We integrate MVTN in a novel adaptive multi-view pipeline that can render either 3D meshes or point clouds. MVTN exhibits clear performance gains in the tasks of 3D shape classification and 3D shape retrieval without the need for extra training supervision. In these tasks, MVTN achieves state-of-the-art performance on ModelNet40, ShapeNet Core55, and the most recent and realistic ScanObjectNN dataset (up to 6\% improvement). Interestingly, we also show that MVTN can provide network robustness against rotation and occlusion in the 3D domain. The code is available at \url{https://github.com/ajhamdi/MVTN}.
\end{abstract}
\section{Introduction} \label{sec:introduction}
Given its success in the 2D realm, deep learning naturally expanded to the 3D vision domain. In 3D, deep networks achieve impressive results in classification, segmentation, and detection. 3D deep learning pipelines operate directly on 3D data, commonly represented as point clouds \cite{pointnet,pointnet++,dgcn}, meshes \cite{meshnet,meshcnn}, or voxels \cite{voxnet,minkosky,sparseconv}. However, other methods choose to represent 3D information by rendering multiple 2D views of objects or scenes \cite{mvcnn}. Such multi-view methods are more similar to a human approach, where the human visual system is fed with streams of rendered images instead of more elaborate 3D representations.

\begin{figure}[t]
    \centering
    \includegraphics[trim= 5cm 3cm 8cm 3cm , clip,width=\linewidth ]{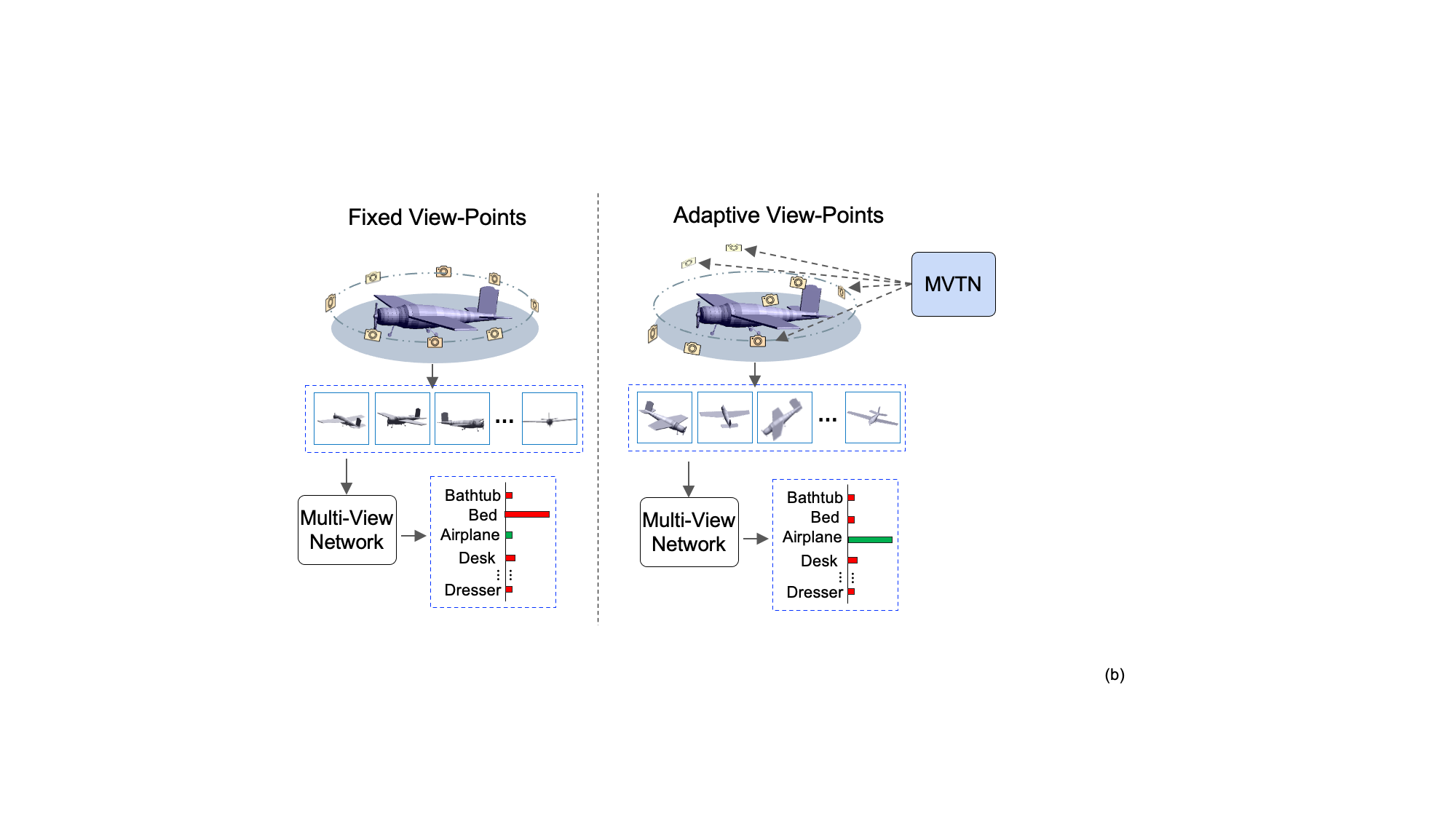}
    \caption{ \small \textbf{Multi-View Transformation Network (MVTN).} We propose a differentiable module that predicts the best view-points for a task-specific multi-view network. MVTN is trained jointly with this network without any extra training supervision, while improving the performance on 3D classification and shape retrieval.}
    \label{fig:pullingFigure}
\end{figure}

Recent developments in multi-view methods show impressive performance, and in many instances, achieve state-of-the-art results in 3D shape classification and segmentation \cite{mvrotationnet,mvviewgcn,mvvirtualsceneseg,mvshapeseg,mvsceneseg}. Multi-view approaches bridge the gap between 2D and 3D learning by solving a 3D task using 2D convolutional architectures. These methods render several views for a given 3D shape and leverage the rendered images to solve the end task. As a result, they build upon the recent advances in 2D grid-based deep learning and leverage larger image datasets for pre-training (\eg ImageNet~\cite{IMAGENET}) to compensate for the general scarcity of labeled 3D datasets.
However, the manner of choosing the rendering view-points for such methods remains mostly unexplored.
Current methods rely on heuristics like random sampling in scenes \cite{mvvirtualsceneseg} or predefined canonical view-points in oriented datasets \cite{mvviewgcn}. There is no evidence suggesting that such heuristics are empirically the best choice. To address this shortcoming, we propose to learn better view-points by introducing a Multi-View Transformation Network (MVTN). 
As shown in \figLabel{\ref{fig:pullingFigure}}, MVTN learns to regress view-points, renders those views with a differentiable renderer, and trains the downstream task-specific network in an end-to-end fashion, thus leading to the most suitable views for the task. 
MVTN is inspired by the Spatial Transformer Network (STN) \cite{stn}, which was developed for the 2D image domain. Both MVTN and STN learn spatial transformations for the input without leveraging any extra supervision nor adjusting the learning process. 

The paradigm of perception by predicting the best environment parameters that generated the image is called  
Vision as Inverse Graphics (VIG) \cite{old-vision1,vig-cinvg,vig-nsd,vig-reinforce,vig-inverse-render-net}.
One approach to VIG is to make the rendering process invertible or differentiable \cite{vig-open-dr,vig-nmr,soft-rasterizer,vig-bid-r,vig-monte-carlo-raytrace}. In this paper, MVTN takes advantage of differentiable rendering \cite{vig-nmr,soft-rasterizer,pytorch3d}. With such a renderer, models can be trained end-to-end for a specific target 3D vision task, with the view-points (\ie camera poses) 
being inferred by MVTN in the same forward pass.
To the best of our knowledge, we are the first to integrate a learnable approach to view-point prediction in multi-view methods by using a differentiable renderer and establishing an end-to-end pipeline that works for \textit{both} mesh and 3D point cloud classification and retrieval. 

\vspace{2pt}\noindent\textbf{Contributions:} \textbf{(i)} We propose a Multi-View Transformation Network (MVTN) that regresses better view-points for multi-view methods. Our MVTN leverages a differentiable renderer that enables end-to-end training for 3D shape recognition tasks. 
\textbf{(ii)} Combining MVTN with multi-view approaches leads to state-of-the-art results in 3D classification and shape retrieval on standard benchmarks ModelNet40 \cite{modelnet}, ShapeNet Core55 \cite{shapenet,shrek17}, and ScanObjectNN \cite{scanobjectnn}. 
\textbf{(iii)} Additional analysis shows that MVTN improves the robustness of multi-view approaches to rotation and occlusion, making MVTN more practical for realistic scenarios, where 3D models are not perfectly aligned or partially cropped.

\section{Related Work} \label{sec:related}
\mysection{Deep Learning on 3D Data}
PointNet \cite{pointnet} paved the way as the first deep learning algorithm to operate directly on 3D point clouds. PointNet computes point features independently and aggregates them using an order invariant function like max-pooling. Subsequent works focused on finding neighborhoods of points to define point convolutional operations \cite{pointnet++,dgcn,pc_li2018pointcnn,pc_landrieu2018large,pc_landrieu2019point,pc_wang2018sgpn}.   
Voxel-based deep networks allow for 3D CNNs yet suffer from cubic memory complexity \cite{voxnet,minkosky,sparseconv}. Several recent works combine point cloud representations with other 3D modalities like voxels \cite{pvoxelcnn} or multi-view images \cite{pvnet,mvpnet}. In this paper, we leverage a point encoder to predict the optimal view-points, from which images are rendered and fed to a multi-view network.

\mysection{Multi-View 3D Shape Classification}
The first work on using 2D images to recognize 3D objects was proposed by Bradski~\etal~\cite{bradski1994recognition}. Twenty years later and after the success of deep learning in 2D vision tasks, MVCNN \cite{mvcnn} emerged as the first use of deep 2D CNNs for 3D object recognition. The original MVCNN uses max pooling to aggregate features from different views. Several follow-up works propose different strategies to assign weights to views to perform weighted average pooling of view-specific features \cite{mvnhbn,mvrelations,mvgvcnn,mvvram}. RotationNet \cite{mvrotationnet} classifies the views and the object jointly. Equivariant MV-Network \cite{mvequivariant} uses a rotation equivariant convolution operation on multi-views by utilizing rotation group convolutions \cite{groupconv}. The more recent work of ViewGCN \cite{mvviewgcn} utilizes dynamic graph convolution operations to adaptively pool features from different fixed views for the task of 3D shape classification. All these previous methods rely on fixed rendered datasets of 3D objects. The work of \cite{mvvram} attempts to select views adaptively through reinforcement learning and RNNs, but this comes with limited success and an elaborate training process. In this paper, we propose a novel MVTN framework for predicting optimal view-points in a multi-view setup. This is done by jointly training MVTN with a multi-view task-specific network, without the need for any extra supervision nor adjustment to the learning process.

\begin{figure*}
    \centering
    \includegraphics[trim= 0cm 2.4cm 0cm 5.5cm , clip,width=0.99\linewidth]{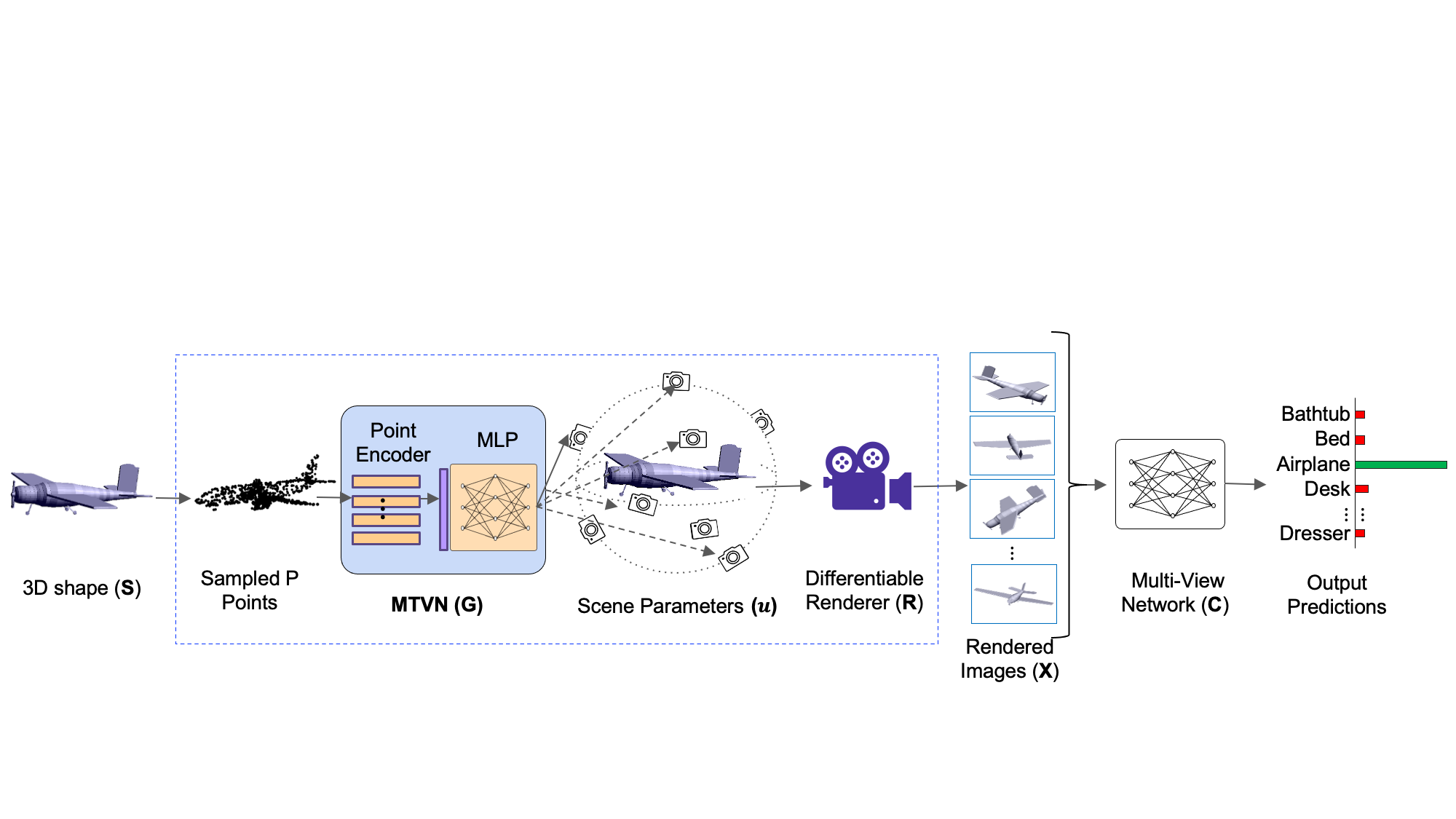}
    \caption{\textbf{End-to-End Learning Pipeline for Multi-View Recognition.} To learn adaptive scene parameters $\mathbf{u}$ that maximize the performance of a multi-view network $\mathbf{C}$ for every 3D object shape $\mathbf{S}$, we use a differentiable renderer $\mathbf{R}$. MVTN extracts coarse features from $\mathbf{S}$ by a point encoder and regresses the adaptive scene parameters for that object. In this example, the parameters $\mathbf{u}$ are the azimuth and elevation angles of cameras pointing towards the center of the object. The MVTN pipeline is optimized end-to-end for the task loss. 
    }
    \label{fig:pipeline}
\end{figure*}

\mysection{3D Shape Retrieval}
Early methods in the literature compare the distribution of hand-crafted descriptors to retrieve similar 3D shapes. Those shape signatures could either represent geometric~\cite{osada2002shape} or visual~\cite{chen2003visual} cues. Traditional geometric methods would estimate distributions of certain characteristics (\eg distances, angles, areas, or volumes) to measure the similarity between shapes \cite{akgul20093d,chaudhuri2010data,bronstein2011shape}.
Gao~\etal~\cite{gao2011camera} use multiple camera projections, and Wu~\etal~\cite{wu20153d} use a voxel grid to extract analogous model-based signatures. Su~\etal~\cite{mvcnn} introduce a deep learning pipeline for multi-view classification, with aggregated features achieving high retrieval performance. They use a low-rank Mahalanobis metric atop extracted multi-view features to improve retrieval performance. This seminal work on multi-view learning is extended for retrieval with
volumetric-based descriptors~\cite{qi2016volumetric},
hierarchical view-group architectures~\cite{mvgvcnn}, and 
triplet-center loss~\cite{he2018triplet}.
Jiang~\etal~\cite{mlvcnn} investigate better views for retrieval using many loops of circular cameras around the three principal axes.
However, these approaches consider fixed camera view-points compared to MVTN's learnable ones.

\mysection{Vision as Inverse Graphics (VIG)}
A key issue in VIG is the non-differentiability of the classical graphics pipeline. 
Recent VIG approaches focus on making the graphics operations differentiable, allowing gradients to flow from the image to the rendering parameters directly \cite{vig-open-dr,vig-nmr,soft-rasterizer,vig-monte-carlo-raytrace,semantic-robustness}.
NMR \cite{vig-nmr} approximates non-differentiable rasterization by smoothing the edge rendering, where SoftRas \cite{soft-rasterizer} assigns a probability for all mesh triangles to every pixel in the image. Synsin \cite{synsin} proposes an alpha-blending mechanism for differentiable point cloud rendering. The Pytorch3D \cite{pytorch3d} renderer improves the speed and modularity of SoftRas and Synsin and allows for customized shaders and point cloud rendering. MVTN harnesses advances in differentiable rendering to train jointly with the multi-view network in an end-to-end fashion. 
Using both mesh and point cloud differentiable rendering enables MVTN to work on 3D CAD models and the more accessible 3D point cloud data.

\section{Methodology} \label{sec:methodology}
We illustrate our proposed multi-view pipeline using MVTN in \figLabel{\ref{fig:pipeline}}. MVTN is a generic module that learns camera view-point transformations for specific 3D multi-view tasks, \eg 3D shape classification. In this section, we review a generic framework for common multi-view pipelines, introduce MVTN details, and present an integration of MVTN for 3D shape classification and retrieval.

\subsection{Overview of Multi-View 3D Recognition}
\vspace{-4pt}
3D multi-view recognition defines $M$ different images $\{\mathbf{x}_i\}_{i=1}^M$ rendered from multiple view-points of the same shape $\mathbf{S}$. The views are fed into the same backbone network $\mathbf{f}$ that extracts discriminative features per view. These features are then aggregated among views to describe the entire shape and used for downstream tasks such as classification or retrieval. 
Specifically, a multi-view network $\mathbf{C}$ with parameters $\boldsymbol{\theta}_{\mathbf{C}}$
operates on an input set of images $\mathbf{X} \in \mathbb{R}^{M  \times h\times w \times c }$ to obtain a softmax probability vector for the shape $\mathbf{S}$.

\mysection{Training Multi-View Networks}
The simplest deep multi-view classifier is MVCNN, where $\mathbf{C} = \text{MLP}\left( \max_{i} \mathbf{f}(\mathbf{x}_i)\right) $ with $\mathbf{f} : \mathbb{R}^{h \times w \times c} \rightarrow{\mathbb{R}^{d}}$ being a 2D CNN backbone (\eg ResNet \cite{resnet}) applied individually on each rendered image. A more recent method like ViewGCN would be described as $\mathbf{C} = \text{MLP}\left( \text{cat}_{\text{GCN}}\left( \mathbf{f}(\mathbf{x}_i)\right)\right) $, where $\text{cat}_{\text{GCN}}$ is an aggregation of views' features learned from a graph convolutional network. In general, learning a task-specific multi-view network on a labeled 3D dataset is formulated as:
\begin{equation}
\begin{aligned} 
 &\argmin_{\boldsymbol{\theta}_{\mathbf{C}}}~~ \sum_{n}^{N} L~\big( \mathbf{C} (\mathbf{X}_n)~,~y_n \big) \\ =~
 &\argmin_{\boldsymbol{\theta}_{\mathbf{C}}} \sum_{n}^{N} L~\Big( \mathbf{C} \big(\mathbf{R}(\mathbf{S}_n,\mathbf{u}_0)\big)~,~y_n \Big),
\label{eq:mv-objectiive}
\end{aligned} 
\end{equation}
\noindent where $L$ is a task-specific loss defined over $N$ 3D shapes in the dataset,
$y_n$ is the label for the $n^{\text{th}}$ 3D shape $\mathbf{S}_n$, and $\mathbf{u}_0 \in \mathbb{R}^{\tau}$ is a set of $\tau$ fixed scene parameters for the entire dataset. These parameters represent properties that affect the rendered image, including camera view-point, light, object color, and background. $\mathbf{R}$ is the renderer that takes as input a shape $\mathbf{S}_n$ and the parameters $\mathbf{u}_0$ to produce $M$ multi-view images $\mathbf{X}_n$ per shape. 
In our experiments, we choose the scene parameters $\mathbf{u}$ to be the azimuth and elevation angles of the camera view-points pointing towards the object center, thus setting $\tau = 2M$.

\mysection{Canonical Views}
Previous multi-view methods rely on scene parameters $\mathbf{u}_0$ that are pre-defined for the entire 3D dataset. In particular, the fixed camera view-points are usually selected based on the alignment of the 3D models in the dataset. The most common view configurations are \textit{circular} that aligns view-points on a circle around the object~\cite{mvcnn,mvnhbn} and \textit{spherical} that aligns equally spaced view-points on a sphere surrounding the object~\cite{mvviewgcn,mvrotationnet}. 
Fixing those canonical views for all 3D objects can be misleading for some classes. For example, looking at a bed from the bottom could confuse a 3D classifier.
In contrast, MVTN learns to regress per-shape view-points, as illustrated in \figLabel{\ref{fig:views-types}}.

\subsection{Multi-View Transformation Network (MVTN)}
\vspace{-4pt}
Previous multi-view methods take the multi-view image $\mathbf{X}$ as the only representation for the 3D shape, where $\mathbf{X}$ is rendered using fixed scene parameters $\mathbf{u}_0$. In contrast, we consider a more general case, where $\mathbf{u}$ is \textit{variable} yet within bounds $\pm \mathbf{u}_{\text{bound}}$.
Here, $\mathbf{u}_{\text{bound}}$ is positive and it defines the permissible range for the scene parameters.
We set $\mathbf{u}_{\text{bound}} $ to  $180^\circ$ and $90^\circ$ for each azimuth and elevation angle.

\begin{figure} [t] 
\tabcolsep=0.03cm
\begin{tabular}{c|c|c}  
\textbf{Circular} & \textbf{Spherical}  &    \textbf{MVTN} \\  %

 \includegraphics[trim= 4cm 2.7cm 4cm 2.2cm , clip, width = 0.333\linewidth]{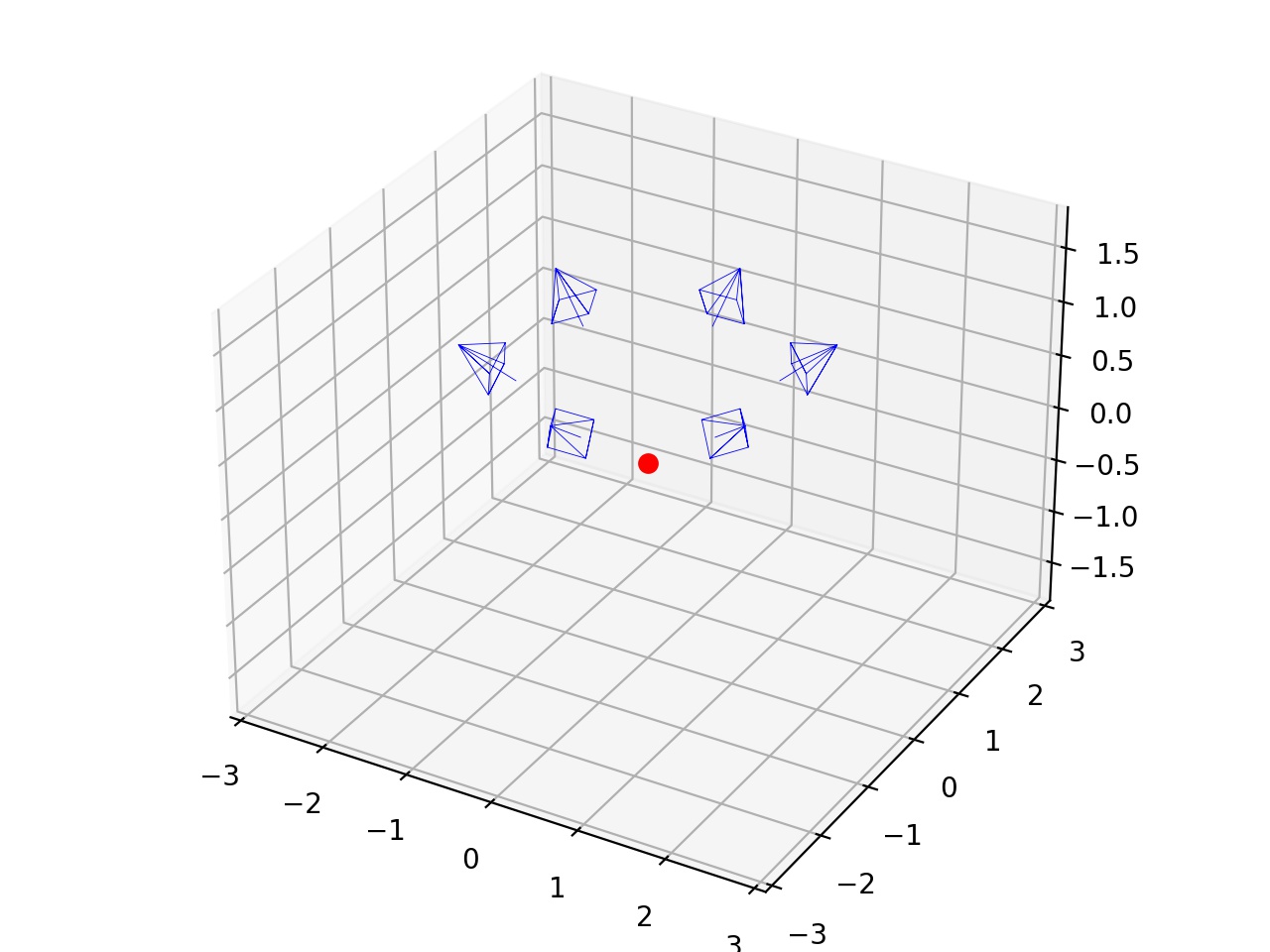} & \includegraphics[trim= 4cm 2.7cm 4cm 2.2cm , clip, width = 0.333\linewidth]{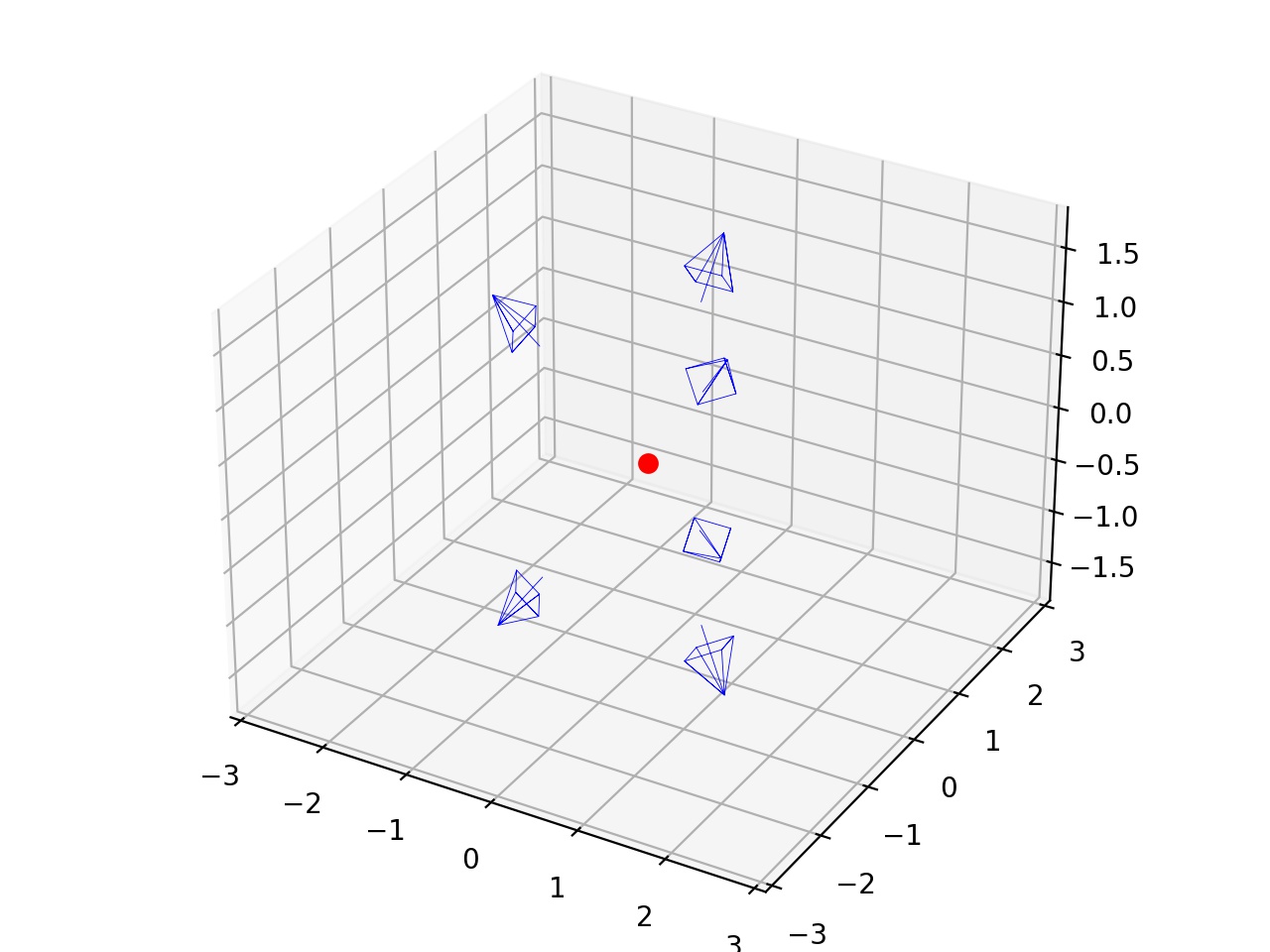} 
&\includegraphics[trim= 4cm 2.7cm 4cm 2.2cm , clip, width = 0.333\linewidth]{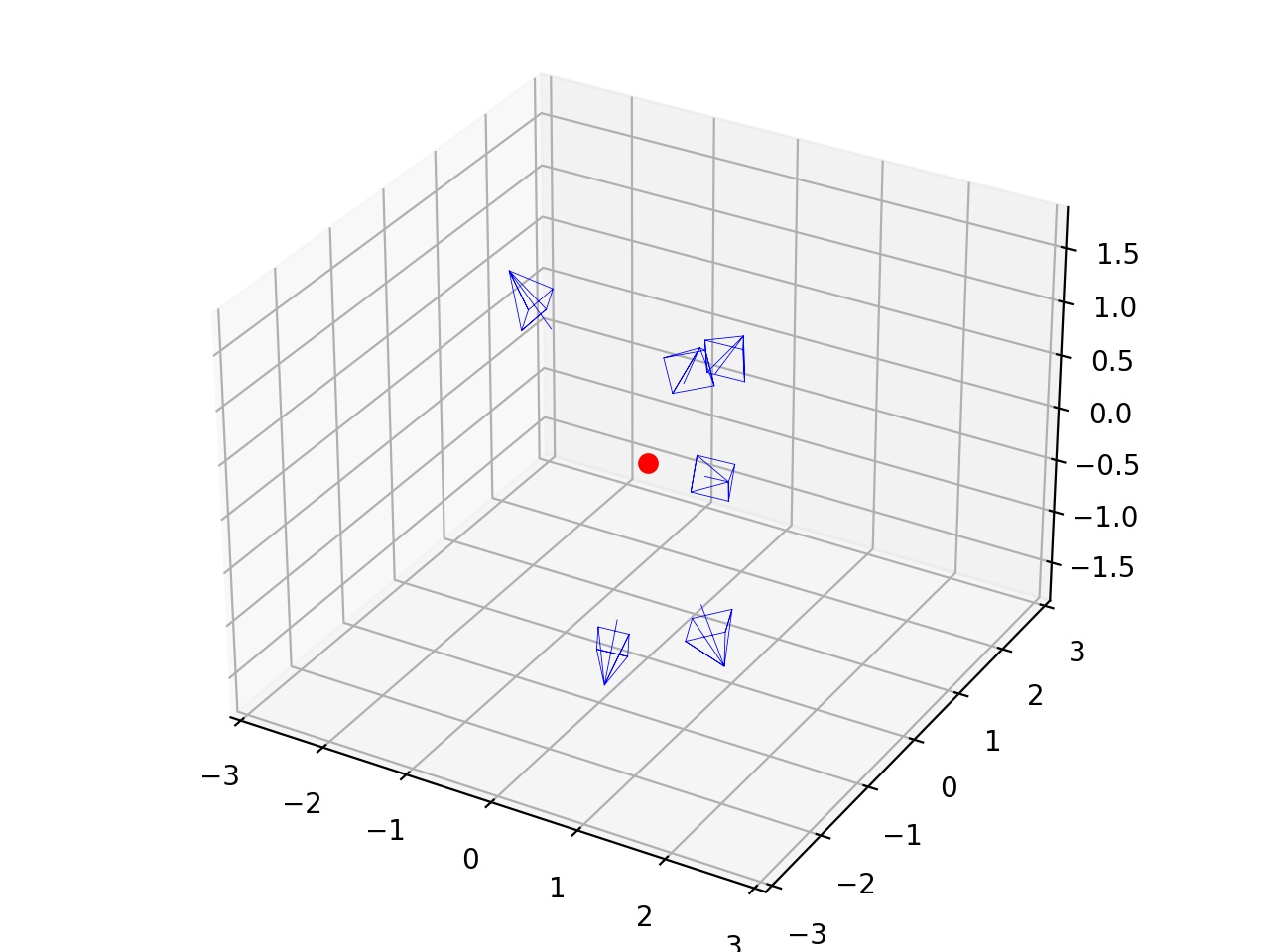} \\
  \includegraphics[trim= 0cm 0cm 24cm 0cm , clip, width = 0.333\linewidth]{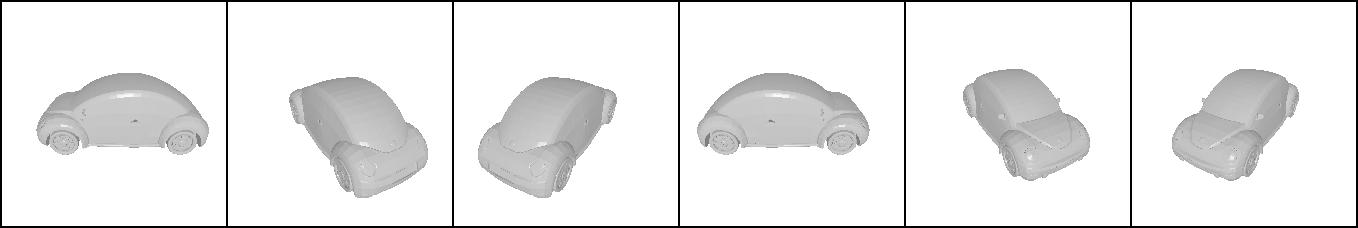} & \includegraphics[trim= 0cm 0cm 24cm 0cm , clip, width = 0.333\linewidth]{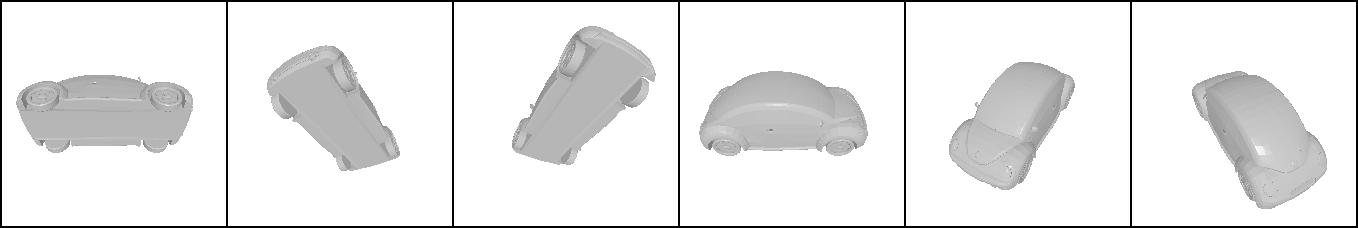} & 
\includegraphics[trim= 0cm 0cm 24cm 0cm , clip, width = 0.333\linewidth]{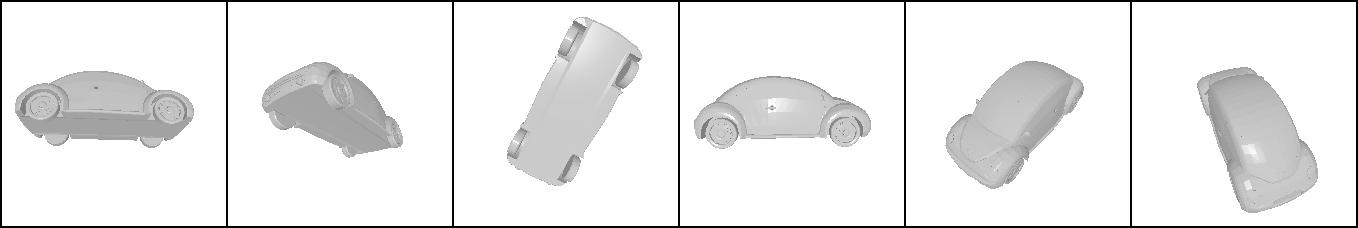}  \\
  \includegraphics[trim= 24cm 0cm 0cm 0cm , clip, width = 0.333\linewidth]{images/qualitative_views/circular_rend.jpg} & \includegraphics[trim= 24cm 0cm 0cm 0cm , clip, width = 0.333\linewidth]{images/qualitative_views/spherical_rend.jpg} & 
\includegraphics[trim= 24cm 0cm 0cm 0cm , clip, width = 0.333\linewidth]{images/qualitative_views/mvt_spherical_rend.jpg}  \\ 

\end{tabular}
\caption{\small \textbf{Multi-View Camera Configurations}: The view setups commonly used in the multi-view literature are circular \cite{mvcnn} or spherical \cite{mvviewgcn,mvrotationnet}. Our MVTN learns to predict specific view-points for each object shape at inference time. The shape's center is shown as a red dot, and the view-points as blue cameras with their mesh renderings shown at the bottom.
}
    \label{fig:views-types}
\end{figure}

\begin{figure}[t]
    \centering
    \includegraphics[width=0.98\linewidth]{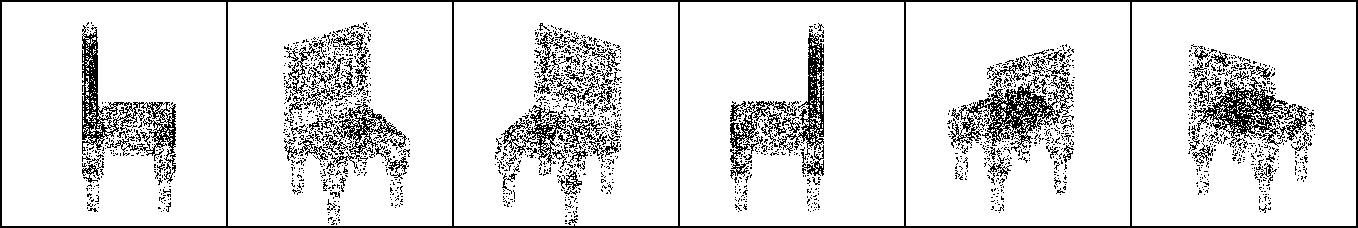}\\
    \includegraphics[width=0.98\linewidth]{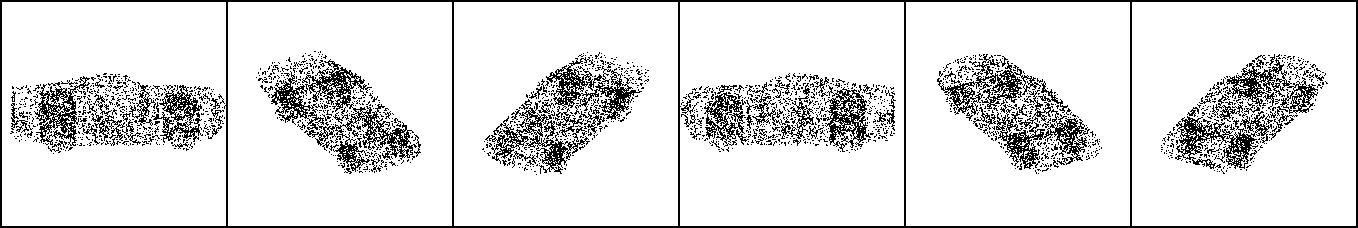} \\
    \includegraphics[width=0.98\linewidth]{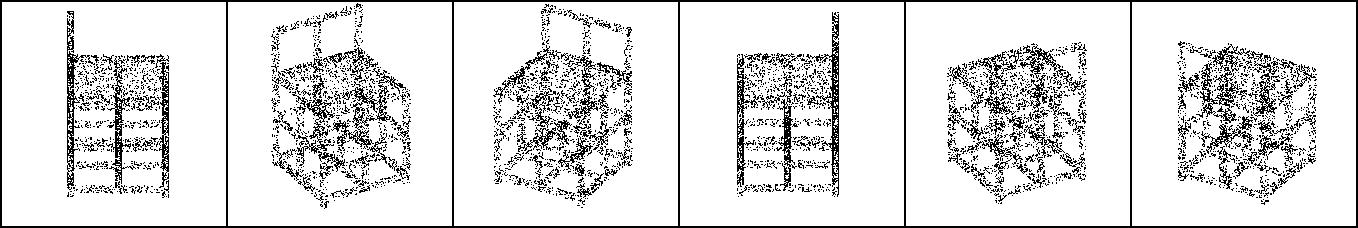}
    \caption{\textbf{Multi-View Point Cloud Renderings.} We show some examples of point cloud renderings used in our pipeline. Note how point cloud renderings offer more information about content hidden from the camera view-point (\eg car wheels from the occluded side), which can be useful for recognition.}
    \label{fig:point-rendring}
\end{figure}

\vspace{1pt}\mysection{Differentiable Renderer}
A renderer $\mathbf{R}$ takes a 3D shape $\mathbf{S}$ (mesh or point cloud) and scene parameters $\mathbf{u}$ as inputs, and outputs the corresponding $M$ rendered images $\{\mathbf{x}_i\}_{i=1}^M$. 
Since $\mathbf{R}$ is differentiable, gradients $\pd{\mathbf{x}_i}{\mathbf{u}}{}$ can propagate backward from each rendered image to the scene parameters, thus establishing a framework that suits end-to-end deep learning pipelines.
When $\mathbf{S}$ is represented as a 3D mesh, $\mathbf{R}$ has two components: a \textit{rasterizer} and a \textit{shader}. First, the rasterizer transforms meshes from the world to view coordinates given the camera view-point and assigns faces to pixels. Using these face assignments, the shader creates multiple values for each pixel then blends them. On the other hand, if $\mathbf{S}$ is represented by a 3D point cloud, $\mathbf{R}$ would use an alpha-blending mechanism instead \cite{synsin}.
\figLabel{\ref{fig:views-types}} and \figLabel{\ref{fig:point-rendring}} illustrate examples of mesh and point cloud renderings used in MVTN. %

\vspace{1pt}\mysection{View-Points Conditioned on 3D Shape}
We design $\mathbf{u}$ to be a function of the 3D shape by learning a Multi-View Transformation Network (MVTN), denoted as $\mathbf{G} \in \mathbb{R}^{P\times 3} \rightarrow{\mathbb{R}^{\tau}} $ and parameterized by $\boldsymbol{\theta}_{\mathbf{G}}$, where $P$ is the number of points sampled from shape $\mathbf{S}$. 
Unlike \eqLabel{\ref{eq:mv-objectiive}} that relies on constant rendering parameters, MVTN predicts $\mathbf{u}$ adaptively for each object shape $\mathbf{S}$ and is optimized along with the classifier $ \mathbf{C}$. The pipeline is trained end-to-end to minimize the following loss on a dataset of N objects:
\begin{equation}
\begin{aligned} 
 \argmin_{\boldsymbol{\theta}_{\mathbf{C}}, \boldsymbol{\theta}_{\mathbf{G}}} &\sum_{n}^{N} L~\Big( \mathbf{C} \big(\mathbf{R}(\mathbf{S}_n,\mathbf{u}_n)\big)~,~y_n \Big), \\& \text{s. t.} \quad \mathbf{u}_n = ~ \mathbf{u}_{\text{bound}}.\text{tanh}\big( \mathbf{G}(\mathbf{S}_n)\big)
\label{eq:mvt-objective}
\end{aligned} 
\end{equation}
Here, $\mathbf{G}$ encodes a 3D shape to predict its optimal view-points for the task-specific multi-view network $\mathbf{C}$. Since the goal of $\mathbf{G}$ is only to predict view-points and not classify objects (as opposed to $\mathbf{C}$), its architecture is designed to be simple and light-weight. 
As such, we use a simple point encoder (\eg shared MLP as in PointNet \cite{pointnet}) that processes $P$ points from $\mathbf{S}$ and produces coarse shape features of dimension $b$. %
Then, a shallow MLP regresses the scene parameters $\mathbf{u}_n$ from the global shape features.
To force the predicted parameters $\mathbf{u}$ to be within a permissible range $\pm\mathbf{u}_{\text{bound}}$, we use a hyperbolic tangent function scaled by $\mathbf{u}_{\text{bound}}$.

\mysection{MVTN for 3D Shape Classification} \label{sec:mmtd-cls}
To train MVTN for 3D shape classification, 
we define a cross-entropy loss in \eqLabel{\ref{eq:mvt-objective}},
yet other losses and regularizers can be used here as well. %
The multi-view network ($\mathbf{C}$) and the MVTN ($\mathbf{G}$) are trained jointly on the same loss. 
One merit of our multi-view pipeline is its ability to seamlessly handle 3D point clouds, which is absent in previous multi-view methods. When $\mathbf{S}$ is a 3D point cloud, we simply define $\mathbf{R}$ as a differentiable point cloud renderer.

\vspace{1pt}\mysection{MVTN for 3D Shape Retrieval} \label{sec:mmtd-retr}
The shape retrieval task is defined as follows: given a query shape $\mathbf{S}_q$, find the most similar shapes in a broader set of size $N$.
For this task, we follow the retrieval setup of MVCNN~\cite{mvcnn}. In particular, we consider the deep feature representation of the last layer before the classifier in $\mathbf{C}$. We project those features into a more expressive space using LFDA reduction \cite{sugiyama2007dimensionality} and consider the reduced feature as the signature to describe a shape. At test time, shape signatures are used to retrieve (in order) the most similar shapes in the training set.

\section{Experiments} \label{sec:experiments}
We evaluate MVTN for the tasks of 3D shape classification and retrieval on  ModelNet40 \cite{modelnet}, ShapeNet Core55 \cite{shapenet}, and the more realistic ScanObjectNN \cite{scanobjectnn}.
\subsection{Datasets}
\vspace{-4pt}
\mysection{ModelNet40}
ModelNet40 \cite{modelnet} is composed of 12,311 3D objects (9,843/2,468 in training/testing) labelled with 40 object classes. %
Since we render 3D models in the forward pass, we limit the number of triangles in the meshes due to hardware constraints. 
In particular, we simplify the meshes to 20k vertices using the official Blender API \cite{blender,mesh-simplify}.

\mysection{ShapeNet Core55}
ShapeNet Core55 is a subset of ShapeNet \cite{shapenet} comprising 51,162 3D mesh objects labelled with 55 object classes. The training, validation, and test sets consist of 35764, 5133, and 10265 shapes, respectively. It is designed for the shape retrieval challenge SHREK \cite{shrek17}. 

\mysection{ScanObjectNN}
ScanObjectNN \cite{scanobjectnn} is a recently released point cloud dataset for 3D classification that is more realistic and challenging than ModelNet40, since it includes background and considers occlusions. The dataset is composed of 2902 point clouds divided into 15 object categories. We consider its three main variants: object only, object with background, and the hardest perturbed variant (PB\_T50\_RS variant). These variants are used in the 3D Scene Understanding Benchmark associated with the ScanObjectNN dataset.
This dataset offers a more challenging setup than ModelNet40 and tests the generalization capability of 3D deep learning model in more realistic scenarios.

\subsection{Metrics}
\vspace{-4pt}
\mysection{Classification Accuracy} 
The standard evaluation metric in 3D classification is accuracy. We report overall accuracy (percentage of correctly classified test samples) and average per-class accuracy (mean of all true class accuracies).

\mysection{Retrieval mAP} 
Shape retrieval is evaluated by mean Average Precision (mAP) over test queries. For every query shape $\mathbf{S}_q$ from the test set, AP is defined as $AP= \frac{1}{\text{GTP}} \sum_{n}^{N}\frac{\mathbbm{1}(\mathbf{S}_n)}{n} $, where $GTP$ is the number of ground truth positives, $N$ is the size of the ordered training set, and $\mathbbm{1}(\mathbf{S}_n) = 1$ if the shape $\mathbf{S}_n$ is from the same class label of query $\mathbf{S}_q$. We average the retrieval AP over the test set to measure retrieval mAP.

\subsection{Baselines}
\vspace{-4pt}
\mysection{Voxel Networks}
We choose VoxNet \cite{voxnet}, DLAN \cite{dlanretr}, and 3DShapeNets \cite{modelnet} as baselines that use voxels.

\mysection{Point Cloud Networks}
We select PointNet \cite{pointnet}, PointNet++ \cite{pointnet++}, DGCNN \cite{dgcn}, PVNet \cite{pvnet}, and KPConv \cite{kpconv} as baselines that use point clouds.
These methods leverage different convolution operators on point clouds by aggregating local and global point information.

\mysection{Multi-view Networks}
We compare against MVCNN \cite{mvcnn}, RotationNet \cite{mvrotationnet}, GVCNN \cite{mvgvcnn} and ViewGCN \cite{mvviewgcn} as representative multi-view methods.
These methods are limited to meshes, pre-rendered from canonical view-points.
 \subsection{MVTN Details}
\vspace{-4pt}
\mysection{Rendering}
We choose the differentiable mesh and point cloud renderers $\mathbf{R}$ from Pytorch3D \cite{pytorch3d} in our pipeline for their speed and compatibility with Pytorch libraries \cite{paszke2017pytorch}. We show examples of the rendered images for meshes (\figLabel{\ref{fig:views-types}}) and point clouds (\figLabel{\ref{fig:point-rendring}}).
Each rendered image has a size of 224$\times$224. 
For ModelNet40, we use the differentiable \textit{mesh} renderer. We direct the light randomly and assign a random color for the object for augmentation purposes in training. In testing, we keep a fixed light pointing towards the object center and color the object white for stable performance. For ShapeNet Core55 and ScanObjectNN, we use the differentiable \textit{point cloud} renderer using 2048 and 5000 points, respectively. Point cloud rendering offers a light alternative to mesh rendering when the mesh contains a large number of faces that hinders training the MVTN pipeline.

\begin{table}[t]
\tabcolsep=0.07cm
    \centering
\resizebox{0.98\linewidth}{!}{\begin{tabular}{rccc}
\toprule
 &   & \multicolumn{2}{c}{Classification Accuracy} \\
\multicolumn{1}{c}{Method}       & Data Type & (\textbf{Per-Class})   & (\textbf{Overall}) \\ \midrule
VoxNet \cite{voxnet}     & Voxels                 & 83.0 & 85.9      \\
PointNet \cite{pointnet} &  Points                  &       86.2 & 89.2      \\
PointNet++ \cite{pointnet++}   & Points                & - & 91.9      \\
PointCNN \cite{pc_li2018pointcnn}  & Points            &   88.1   & 91.8      \\
DGCNN \cite{dgcn}             & Points                &  90.2     & 92.2      \\
SGAS \cite{sgas}             & Points                &  -     & 93.2      \\
KPConv\cite{kpconv}  &  Points & -  & 92.9 \\ 
PTransformer\cite{pointtransformer}  &  Points & \textbf{90.6} & \textbf{93.7}  \\ \midrule
MVCNN  \cite{mvcnn}         & 12 Views                   & 90.1  & 90.1 \\
GVCNN \cite{mvgvcnn}         & 12 Views                   & 90.7 & 93.1 \\
ViewGCN \cite{mvviewgcn}  & 20 Views   & \textbf{96.5} & \textbf{97.6} \\ 

\midrule
ViewGCN \cite{mvviewgcn}$^*$& 12 views &    90.7   &93.0 \\
ViewGCN \cite{mvviewgcn}$^*$& 20 views &    91.3   &93.3 \\
MVTN (ours)$^*$  & 12 Views       & 92.0 & \textbf{93.8} \\
MVTN (ours)$^*$  & 20 Views       & \textbf{92.2} & 93.5 \\
\bottomrule
\end{tabular}
}
\vspace{2pt}
    \caption{\textbf{3D Shape Classification on ModelNet40}. We compare MVTN against other methods in 3D classification on ModelNet40 \cite{modelnet}. $^*$ indicates results from our rendering setup (differentiable pipeline), while other multi-view results are reported from pre-rendered views. \textbf{Bold} denotes the best result in its setup.}
    \label{tab:ModelNet40-cls}
\end{table}
\begin{table}[t]
\tabcolsep=0.08cm
    \centering
\resizebox{0.9\linewidth}{!}{\begin{tabular}{rccc}
\toprule
 &  \multicolumn{3}{c}{Classification Overall Accuracy } \\
\multicolumn{1}{c}{Method}& \textbf{OBJ\_BG}  & \textbf{OBJ\_ONLY} & \textbf{Hardest}  \\ \midrule
3DMFV \cite{3Dmfv} &  68.2                  &  73.8  &  63.0  \\
PointNet \cite{pointnet}   & 73.3                &   79.2  & 68.0 \\
SpiderCNN \cite{pc_xu2018spidercnn}&    77.1                &    79.5   &  73.7    \\
PointNet ++ \cite{pointnet++}            & 82.3 & 84.3  &   77.9   \\
PointCNN \cite{pc_li2018pointcnn}  & 86.1 & 85.5  & 78.5 \\
DGCNN \cite{dgcn}  & 82.8 & 86.2  & 78.1 \\ 
SimpleView \cite{simpleview}& - & - & 79.5 \\
BGA-DGCNN \cite{scanobjectnn}   & - & - & 79.7 \\
BGA-PN++ \cite{scanobjectnn}   & - & - & 80.2 \\
\midrule
MVTN (ours)  & \textbf{92.6}    & \textbf{92.3} & \textbf{82.8} \\
\bottomrule
\end{tabular}
}
\vspace{2pt}
    \caption{\textbf{3D Point Cloud Classification on ScanObjectNN}. We compare the performance of MVTN in 3D point cloud classification on three different variants of ScanObjectNN \cite{scanobjectnn}. The variants include object with background, object only, and the hardest variant.}
    \label{tab:Scanobjectnn}
\end{table}

\mysection{View-Point Prediction}
As shown in \eqLabel{\ref{eq:mvt-objective}}, the MVTN $\mathbf{G}$ network learns to predict the view-points directly (\textit{MVTN-direct}). Alternatively, MVTN can learn relative offsets \wrt initial parameters $\mathbf{u}_0$. In this case, we concatenate the point features extracted in $\mathbf{G}$ with $\mathbf{u}_0$ to predict the offsets to apply on $\mathbf{u}_0$.
The learned view-points $\mathbf{u}_n$ in \eqLabel{\ref{eq:mvt-objective}} are defined as: $~ \mathbf{u}_n = \mathbf{u}_0 + \mathbf{u}_{\text{bound}}.\text{tanh}\big( \mathbf{G}(\mathbf{u}_0~,~\mathbf{S}_n)\big)$. 
We take $\mathbf{u}_0$ to be the circular or spherical configurations commonly used in multi-view classification pipelines~\cite{mvcnn,mvrotationnet,mvviewgcn}.
We refer to these learnable variants as \textit{MVTN-circular} and \textit{MVTN-spherical}, accordingly. 
For MVTN-circular, the initial elevations for the views are  30$^\circ$, and the azimuth angles are equally distributed over 360$^\circ$ following \cite{mvcnn}. For MVTN-spherical, we follow the method from \cite{spherical-config} that places equally-spaced view-points on a sphere for an arbitrary number of views, which is similar to the ``dodecahedral'' configuration in ViewGCN. 

\begin{table}[t]
\tabcolsep=0.07cm
    \centering
\resizebox{0.98\linewidth}{!}{\begin{tabular}{rccc}
\toprule
 &   & \multicolumn{2}{c}{ Shape Retrieval (mAP)} \\
\multicolumn{1}{c}{Method} & Data Type  & \textbf{ModelNet40}  & \textbf{ShapeNet Core} \\ \midrule
LFD     \cite{lfd}                &  Voxels                 & 40.9 & - \\ 
3D ShapeNets \cite{modelnet} &  Voxels                 & 49.2 & - \\
Densepoint\cite{densepoint}  &  Points & 88.5 & - \\ 
PVNet\cite{pvnet}  &  Points & 89.5 & - \\ 
MVCNN  \cite{mvcnn}         & 12 Views                   & 80.2 & 73.5 \\
GIFT \cite{giftretr}&    20 Views                &     - & 64.0      \\
MVFusionNet \cite{mvfusionnet}            & 12 Views                &      - & 62.2      \\
ReVGG \cite{shrek17}  & 20 Views            &     - & 74.9      \\
RotNet \cite{mvrotationnet}  & 20 Views   & - & 77.2 \\ 
ViewGCN \cite{mvviewgcn}  & 20 Views   & - &  78.4 \\ 
MLVCNN \cite{mlvcnn} &   24 Views               & 92.2      & -\\
\midrule
MVTN (ours)  & 12 Views         & \textbf{92.9} &  \textbf{82.9} \\
\bottomrule
\end{tabular}
}
\vspace{2pt}
    \caption{\textbf{3D Shape Retrieval}. We benchmark the shape retrieval mAP of  MVTN on ModelNet40 \cite{modelnet} and ShapeNet Core55 \cite{shapenet,shrek17}. MVTN achieves the best retrieval performance among recent state-of-the-art methods on both datasets with only 12 views.}
     \label{tab:retrieval}
\end{table}

\mysection{Architecture}
We select MVCNN \cite{mvcnn}, RotationNet \cite{mvrotationnet}, and the more recent ViewGCN \cite{mvviewgcn} as our multi-view networks of choice in the MVTN pipeline.
In our experiments, we select PointNet \cite{pointnet} as the 3D point encoder network $\mathbf{G}$ and experiment with DGCNN in Section \ref{sec:ablation}.
We sample $P = 2048$ points from each mesh as input to the point encoder and use a 5-layer MLP for the regression network, which takes as input the point features extracted by the point encoder of size $b=40$.
All MVTN variants and the baseline multi-view networks use ResNet-18 \cite{resnet} pre-trained on ImageNet \cite{IMAGENET} for the multi-view backbone in $C$, with output features of size $d=1024$. The main classification and retrieval results are based on MVTN-spherical with ViewGCN \cite{mvviewgcn} as the multi-view network $\mathbf{C}$, unless otherwise specified as in \secLabel{\ref{sec:exp-robust}} and \secLabel{\ref{sec:ablation}}. %

\mysection{Training Setup}
To avoid gradient instability introduced by the renderer, we use gradient clipping in the MVTN network $\mathbf{G}$. We clip the gradient updates such that the $\ell_2$ norm of the gradients does not exceed 30. We use a learning rate of $0.001$ but refrain from fine-tuning the hyper-parameters introduced in MVCNN~ \cite{mvcnn} and View-GCN~\cite{mvviewgcn}. More details about the training procedure are in the \supp\hspace{-2pt}.

\begin{figure} [t] 
\tabcolsep=0.03cm
\resizebox{0.98\linewidth}{!}{
\begin{tabular}{c|ccccc}

\includegraphics[width = 0.16666666666666666\linewidth]{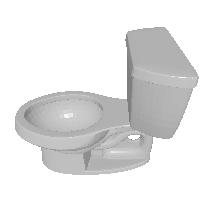} &
\includegraphics[width = 0.16666666666666666\linewidth]{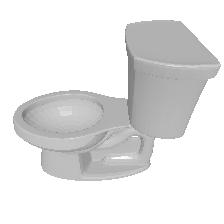} &
\includegraphics[width = 0.16666666666666666\linewidth]{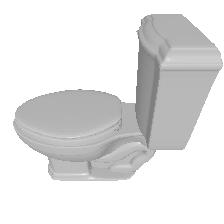} &
\includegraphics[width = 0.16666666666666666\linewidth]{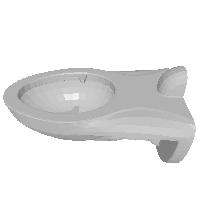} &
\includegraphics[width = 0.16666666666666666\linewidth]{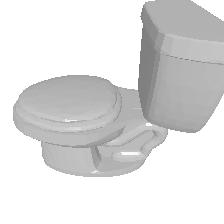} &
\includegraphics[width = 0.16666666666666666\linewidth]{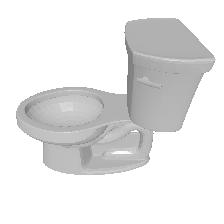} \\ \midrule

\includegraphics[width = 0.16666666666666666\linewidth]{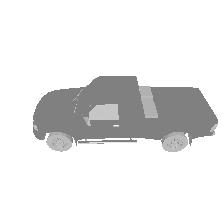} &
\includegraphics[width = 0.16666666666666666\linewidth]{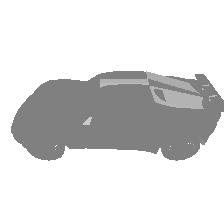} &
\includegraphics[width = 0.16666666666666666\linewidth]{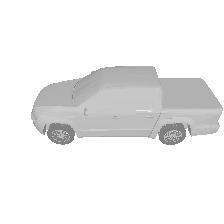} &
\includegraphics[width = 0.16666666666666666\linewidth]{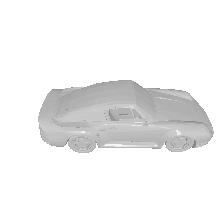} &
\includegraphics[width = 0.16666666666666666\linewidth]{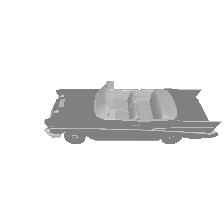} &
\includegraphics[width = 0.16666666666666666\linewidth]{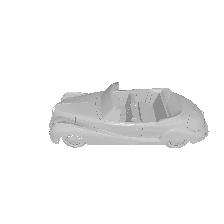} \\ \midrule

\includegraphics[width = 0.16666666666666666\linewidth]{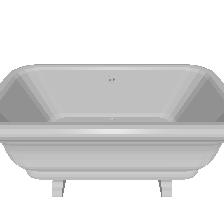} &
\includegraphics[width = 0.16666666666666666\linewidth]{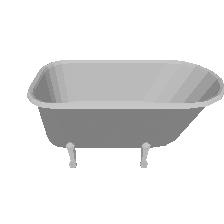} &
\includegraphics[width = 0.16666666666666666\linewidth]{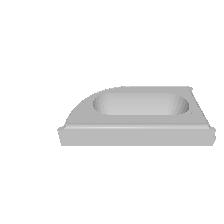} &
\includegraphics[width = 0.16666666666666666\linewidth]{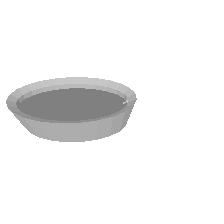} &
\includegraphics[width = 0.16666666666666666\linewidth]{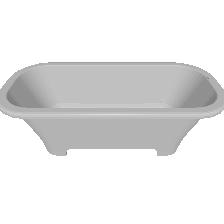} &
\includegraphics[width = 0.16666666666666666\linewidth]{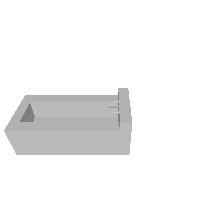} \\ \midrule 

\includegraphics[width = 0.16666666666666666\linewidth]{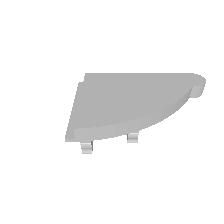} &
\includegraphics[width = 0.16666666666666666\linewidth]{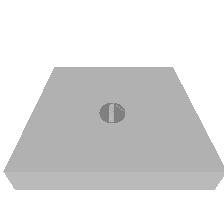} &
\includegraphics[width = 0.16666666666666666\linewidth]{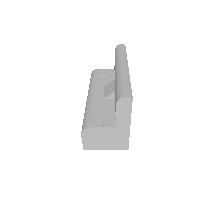} &
\fbox{ \includegraphics[width = 0.16666666666666666\linewidth]{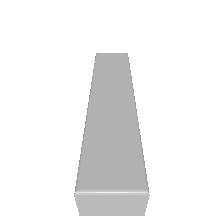}} &
\includegraphics[width = 0.16666666666666666\linewidth]{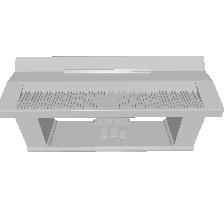} &
\includegraphics[width = 0.16666666666666666\linewidth]{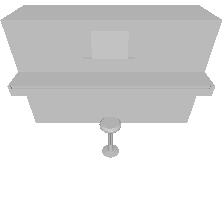} \\ \bottomrule
\end{tabular}
}
\vspace{2pt}
\caption{\small \textbf{Qualitative Examples for Object Retrieval}: \textit{(left):} we show some query objects from the test set. \textit{(right)}: we show top five retrieved objects by our MVTN from the training set. Images of negative retrieved objects are framed.}
    \label{fig:imgs-retr}
\end{figure}
\section{Results} \label{sec:results}
The main results of MVTN are summarized in Tables \ref{tab:ModelNet40-cls}, \ref{tab:Scanobjectnn}, \ref{tab:retrieval} and \ref{tbl:y-robustness}. We achieve state-of-the-art performance in 3D classification on ScanObjectNN by a large margin (up to 6\%) and achieve a competitive test accuracy of \textbf{93.8}\% on ModelNet40. On shape retrieval, we achieve state-of-the-art performance on both ShapeNet Core55 (\textbf{82.9} mAP) and ModelNet40 (\textbf{92.9} mAP). 
Following the common practice, we report the best results out of four runs in benchmark tables, but detailed results are in \supp\hspace{-2pt}.

\subsection{3D Shape Classification} \label{sec:exp-classification}
\vspace{-4pt}
Table \ref{tab:ModelNet40-cls} compares the performance of MVTN against other methods on ModelNet40 \cite{modelnet}. Our MVTN achieves a competitive test accuracy of 93.8\% compared to all previous methods. ViewGCN~\cite{mvviewgcn} achieves higher classification performance by relying on higher quality images from a more advanced yet non-differentiable OpenGL \cite{opengl} renderer. For a fair comparison, we report with an $^*$ the performance of ViewGCN using images generated by the renderer used in MVTN. Using the same rendering process, regressing views with MVTN improves the classification performance of the baseline ViewGCN at 12 and 20 views. We believe future advances in differentiable rendering would bridge the gap between our rendered images and the original high-quality pre-rendered ones.

Table \ref{tab:Scanobjectnn} reports the classification accuracy of a 12 view MVTN on the realistic ScanObjectNN benchmark \cite{scanobjectnn}. MVTN improves performance on different variants of the dataset. The most difficult variant of ScanObjectNN (PB\_T50\_RS) includes challenging scenarios of objects undergoing translation and rotation. Our MVTN achieves state-of-the-art results (+2.6\%) on this variant, highlighting the merits of MVTN for realistic 3D point cloud scans. Also, note how adding background points (in OBJ\_BG) does not hurt MVTN, contrary to most other classifiers. .

\subsection{3D Shape Retrieval} 
\label{sec:exp-retr}
Table \ref{tab:retrieval} reports the retrieval mAP of MVTN compared with recent methods on ModelNet40 \cite{modelnet} and ShapeNet Core55 \cite{shapenet}. The results of the latter methods are taken from \cite{mlvcnn,mvviewgcn,pvnet}. 
MVTN achieves state-of-the-art retrieval performance ($92.9\%$ mAP) on ModelNet40. It also improves the state-of-the-art by a large margin in ShapeNet, while only using 12 views. It is important to note that the baselines in Table \ref{tab:retrieval} include strong and recent methods trained specifically for retrieval, such as MLVCNN \cite{mlvcnn}.
 \figLabel{\ref{fig:imgs-retr}} shows qualitative examples of objects retrieved using MVTN.

\begin{table}[t]

\tabcolsep=0.3cm
\centering
\resizebox{0.98\linewidth}{!}{
\begin{tabular}{rcccc} 
\toprule
& \multicolumn{3}{c}{Rotation Perturbations Range} \\
\multicolumn{1}{c}{Method}  &$0^\circ$ & $\pm90^\circ$ & $\pm180^\circ$ \\ 
\midrule
PointNet \cite{pointnet}&  88.7  &  42.5   &  38.6 \\
PointNet ++ \cite{pointnet++} &  88.2   & 47.9   & 39.7   \\
RSCNN \cite{rspointcloud}  &  90.3 & 90.3    &  90.3  \\ \midrule
MVTN (ours) &  \textbf{91.7} & \textbf{90.8}  & \textbf{91.2}  \\

\bottomrule
\end{tabular}
}
\vspace{2pt}
\caption{\small \textbf{Rotation Robustness on ModelNet40.} 
At test time, we randomly rotate objects in ModelNet40 around the Y-axis (gravity) with different ranges and report the overall accuracy. MVTN displays strong robustness to such Y-rotations.}
\label{tbl:y-robustness}
\end{table}
\subsection{Rotation Robustness} \label{sec:exp-robust}
\vspace{-4pt}
A common practice in 3D shape classification literature is to test the robustness of trained models to perturbations at test time. Following the same setup as \cite{rspointcloud,sada}, we perturb the shapes with random rotations around the Y-axis (gravity-axis) contained within $\pm90^\circ$ and $\pm180^\circ$. 
We repeat the inference ten times for each setup and report the average performance in Table \ref{tbl:y-robustness}. The MVTN-circular variant (with MVCNN) reaches state-of-the-art performance in rotation robustness (91.2\% test accuracy) compared to more advanced methods trained in the same setup. The baseline RSCNN \cite{rspointcloud} is a strong baseline designed to be invariant to translation and rotation. In contrast, MVTN is learned in a simple setup with MVCNN without targeting rotation invariance.

\subsection{Occlusion Robustness} \label{sec:occlusion}
\begin{figure} [] 
\tabcolsep=0.03cm
\resizebox{0.99\linewidth}{!}{
\begin{tabular}{c|ccccc}

 & \multicolumn{5}{c}{Occlusion Ratio} \\ 
 dir.& 0.1 & 0.2 & 0.3 & 0.5 & 0.75 \\ \midrule

+X &
\includegraphics[trim= 0cm 2cm 0cm 2cm , clip, width = 0.19\linewidth]{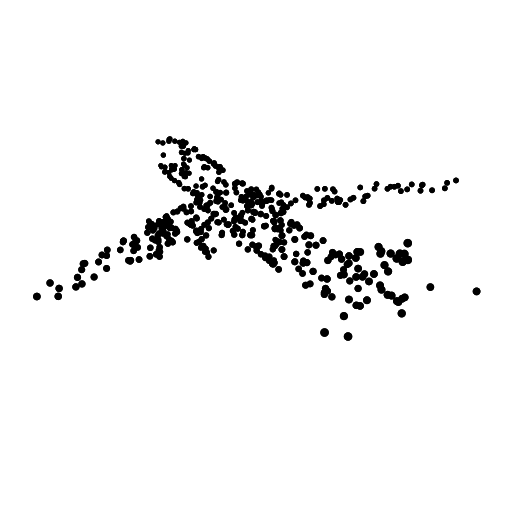} &
\includegraphics[trim= 0cm 2cm 0cm 2cm , clip, width = 0.19\linewidth]{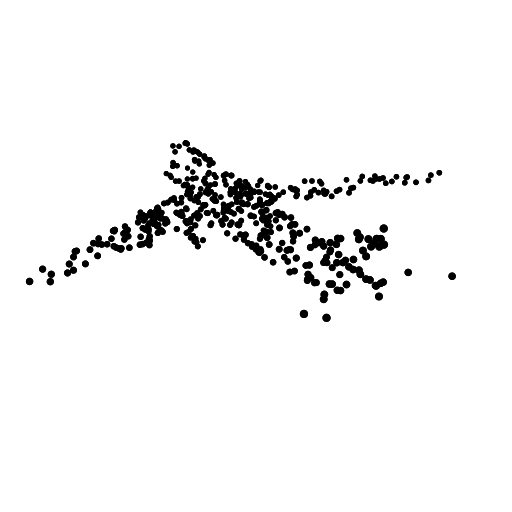} &
\includegraphics[trim= 0cm 2cm 0cm 2cm , clip, width = 0.19\linewidth]{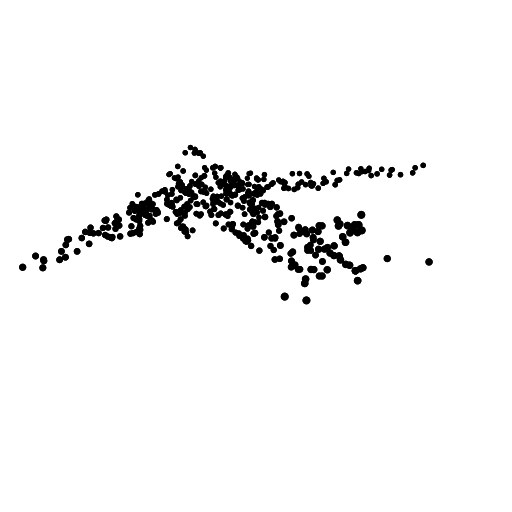} &
\includegraphics[trim= 0cm 2cm 0cm 2cm , clip, width = 0.19\linewidth]{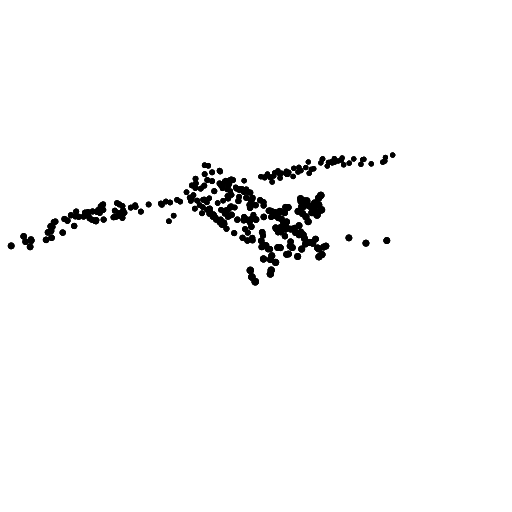} &
\includegraphics[trim= 0cm 2cm 0cm 2cm , clip, width = 0.19\linewidth]{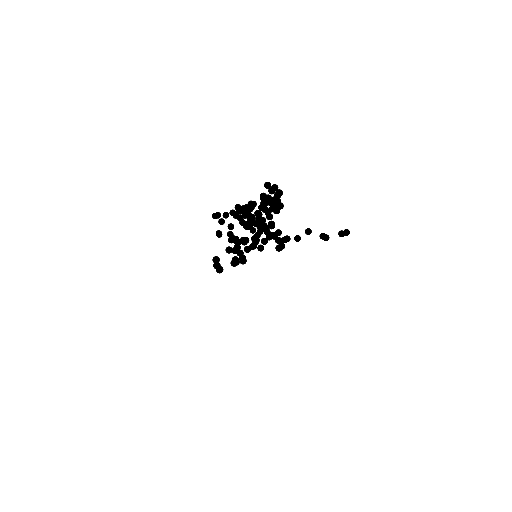} \\ \hline

-X &
\includegraphics[trim= 0cm 2cm 0cm 2cm , clip, width = 0.19\linewidth]{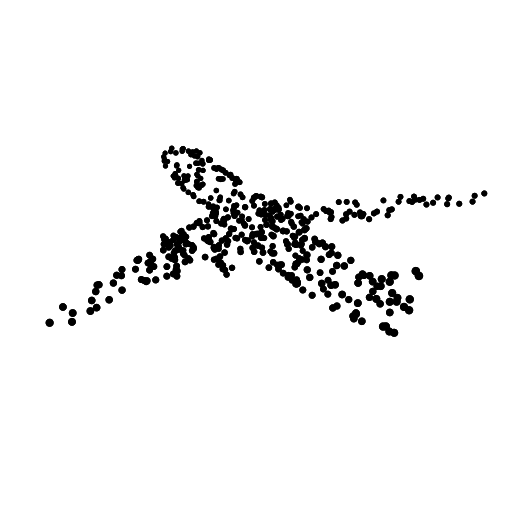} &
\includegraphics[trim= 0cm 2cm 0cm 2cm , clip, width = 0.19\linewidth]{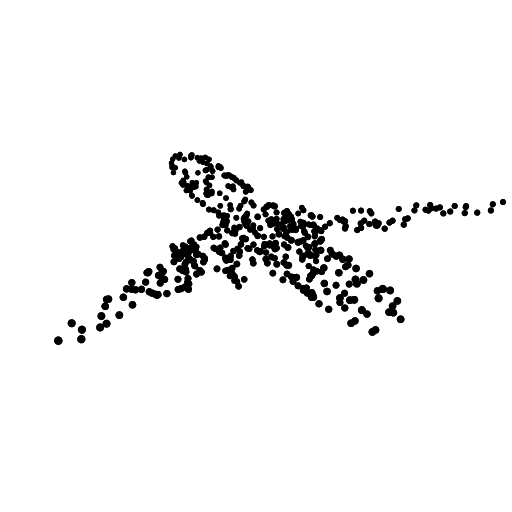} &
\includegraphics[trim= 0cm 2cm 0cm 2cm , clip, width = 0.19\linewidth]{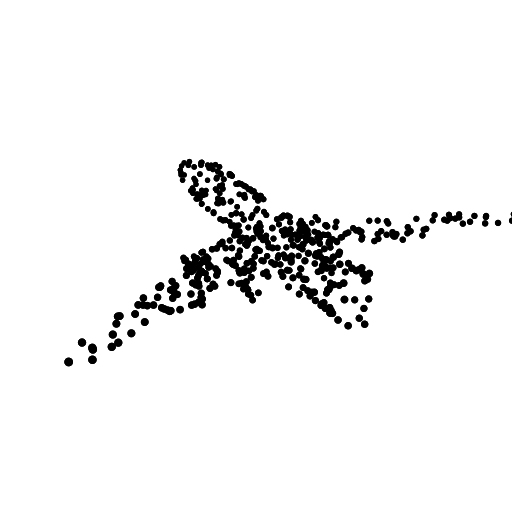} &
\includegraphics[trim= 0cm 2cm 0cm 2cm , clip, width = 0.19\linewidth]{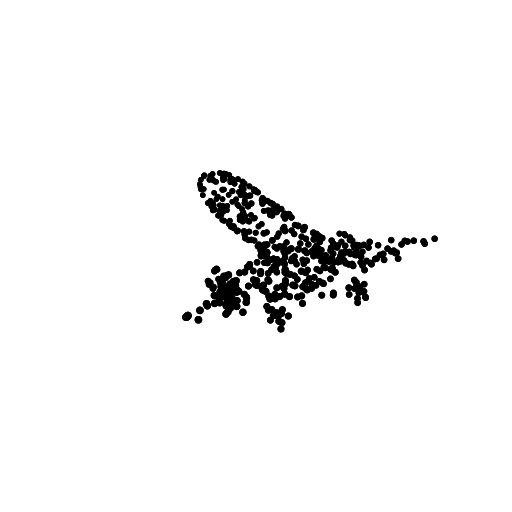} &
\includegraphics[trim= 0cm 2cm 0cm 2cm , clip, width = 0.19\linewidth]{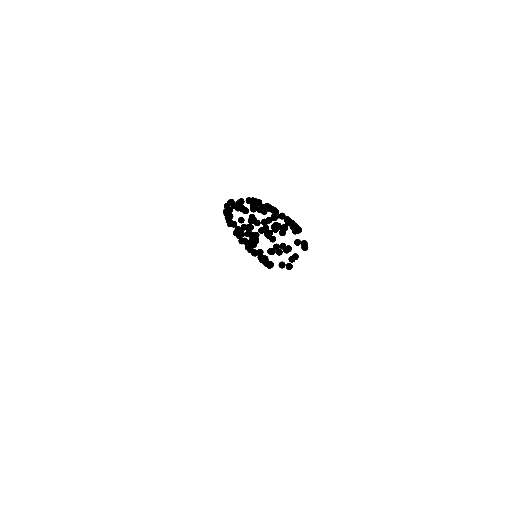} \\ \hline

+Z &
\includegraphics[trim= 0cm 2cm 0cm 2cm , clip, width = 0.19\linewidth]{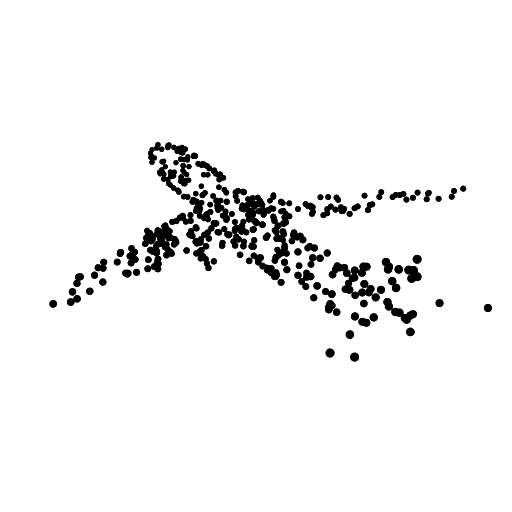} &
\includegraphics[trim= 0cm 2cm 0cm 2cm , clip, width = 0.19\linewidth]{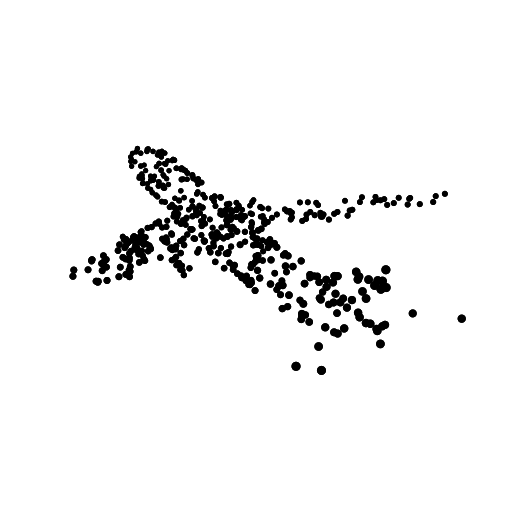} &
\includegraphics[trim= 0cm 2cm 0cm 2cm , clip, width = 0.19\linewidth]{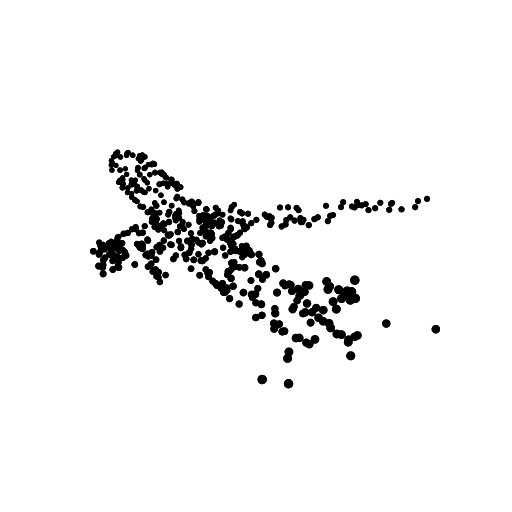} &
\includegraphics[trim= 0cm 2cm 0cm 2cm , clip, width = 0.19\linewidth]{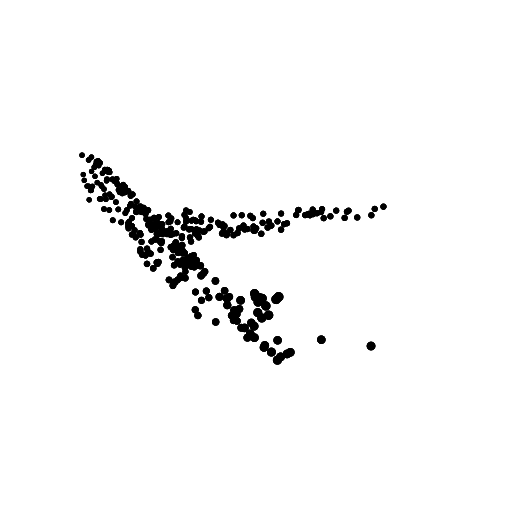} &
\includegraphics[trim= 0cm 2cm 0cm 2cm , clip, width = 0.19\linewidth]{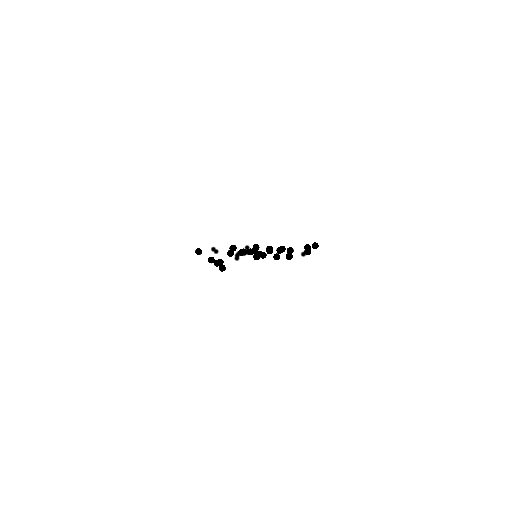} \\ \hline

-Z &
\includegraphics[trim= 0cm 2cm 0cm 2cm , clip, width = 0.19\linewidth]{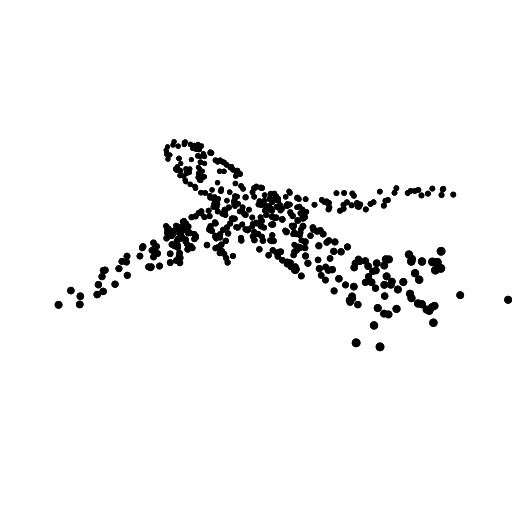} &
\includegraphics[trim= 0cm 2cm 0cm 2cm , clip, width = 0.19\linewidth]{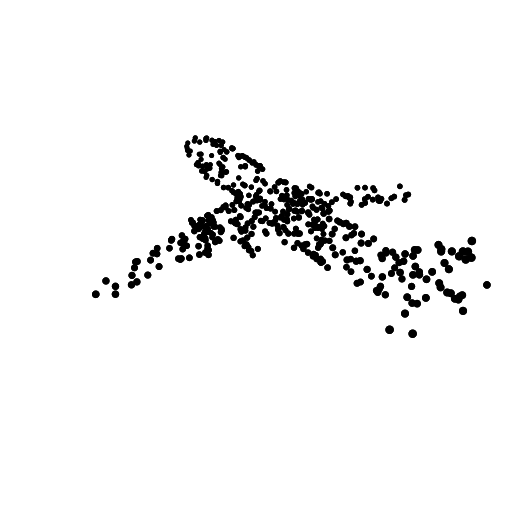} &
\includegraphics[trim= 0cm 2cm 0cm 2cm , clip, width = 0.19\linewidth]{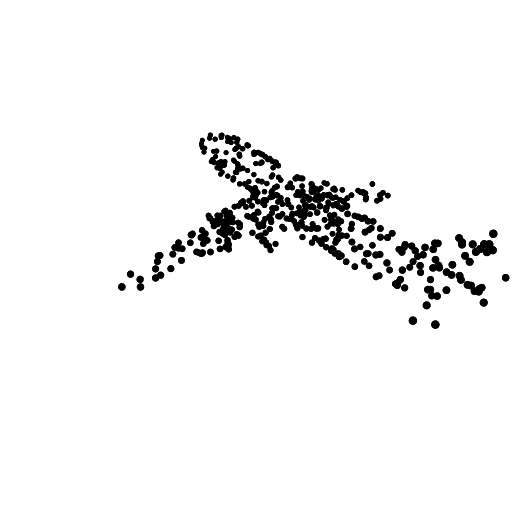} &
\includegraphics[trim= 0cm 2cm 0cm 2cm , clip, width = 0.19\linewidth]{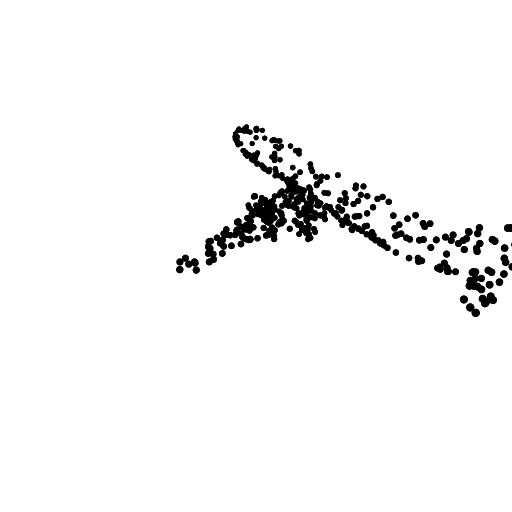} &
\includegraphics[trim= 0cm 2cm 0cm 2cm , clip, width = 0.19\linewidth]{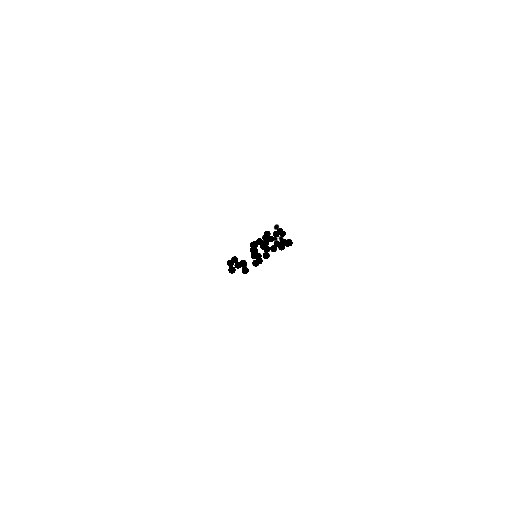} \\ \bottomrule
\end{tabular}
}
\vspace{2pt}
\caption{\small \textbf{Occlusion of 3D Objects}: We simulate realistic occlusion scenarios in 3D point clouds by cropping a percentage of the object along canonical directions. Here, we show an object occluded with different ratios and from different directions. 
}
    \label{fig:occlusion-qual}
\end{figure}

\vspace{-4pt}
To test the usefulness of MVTN in a realistic scenario, we investigate the common problem of occlusion in 3D computer vision, especially in 3D point cloud scans. Various factors lead to occlusion, including the view angle to the object, the sensor's sampling density (\eg LiDAR), or the presence of noise in the sensor. In such realistic scenarios, deep learning models typically fail. To quantify this occlusion effect due to the viewing angle of the 3D sensor in our setup of 3D classification, we simulate realistic occlusion by cropping the object from canonical directions. We train PointNet \cite{pointnet}, DGCNN \cite{dgcn}, and MVTN on the ModelNet40 point cloud dataset. Then, at test time, we crop a portion of the object (from 0\% occlusion ratio to 100\%) along the $\pm$X, $\pm$Y, and $\pm$Z directions. \figLabel{\ref{fig:occlusion-qual}} shows examples of this occlusion effect with different occlusion ratios. In all robustness experiments, the studied transformations (rotation or occlusion) happen only in test time. All the methods compared, including MVTN, are trained naturally without any augmentation by those transformations. We report the average test accuracy of the six cropping directions for the baselines and MVTN in Table \ref{tbl:occlusion}. Note how MVTN achieves high test accuracy even when large portions of the object are cropped. Interestingly, MVTN outperforms PointNet \cite{pointnet} by 13\% in test accuracy when half of the object is occluded. This result is significant, given that PointNet is well-known for its robustness \cite{pointnet,advpc}. 

\begin{table}[t]

\tabcolsep=0.15cm
\centering
\resizebox{0.93\linewidth}{!}{
\begin{tabular}{rcccccc} 
\toprule
& \multicolumn{6}{c}{Occlusion Ratio} \\
\multicolumn{1}{c}{Method} & 0 & 0.1 &0.2 & 0.3 & 0.5 & 0.75 \\ 
\midrule
PointNet \cite{pointnet}& 89.1 &  88.2 &  86.1 &  81.6 &  53.5 &  4.7\\
DGCNN \cite{dgcn} &  92.1 &  77.1 &  74.5 &  71.2 &  30.1 &  4.3\\ \midrule
MVTN (ours)& \textbf{92.3} &  \textbf{90.3} &  \textbf{89.9} &  \textbf{88.3} &  \textbf{67.1} &  \textbf{9.5}   \\
\bottomrule
\end{tabular}
}
\vspace{2pt}
\caption{\small\textbf{Occlusion Robustness of 3D Methods.} We report the test accuracy on point cloud ModelNet40 for different occlusion ratios of the data to measure occlusion robustness of different 3D methods. MVTN achieves 13\% better accuracy than PointNet (a robust network) when half of the object is occluded.}
\label{tbl:occlusion}
\end{table}

\section{Analysis and Insights} \label{sec:analysis}
\subsection{Ablation Study} \label{sec:ablation}
\vspace{-4pt}
This section performs a comprehensive ablation study on the different components of MVTN and their effect on the overall test accuracy on ModelNet40 \cite{modelnet}.

\begin{figure}[t]
    \centering
    \includegraphics[,trim=0 0 1.5cm 1.8cm, clip,width=0.95\linewidth]{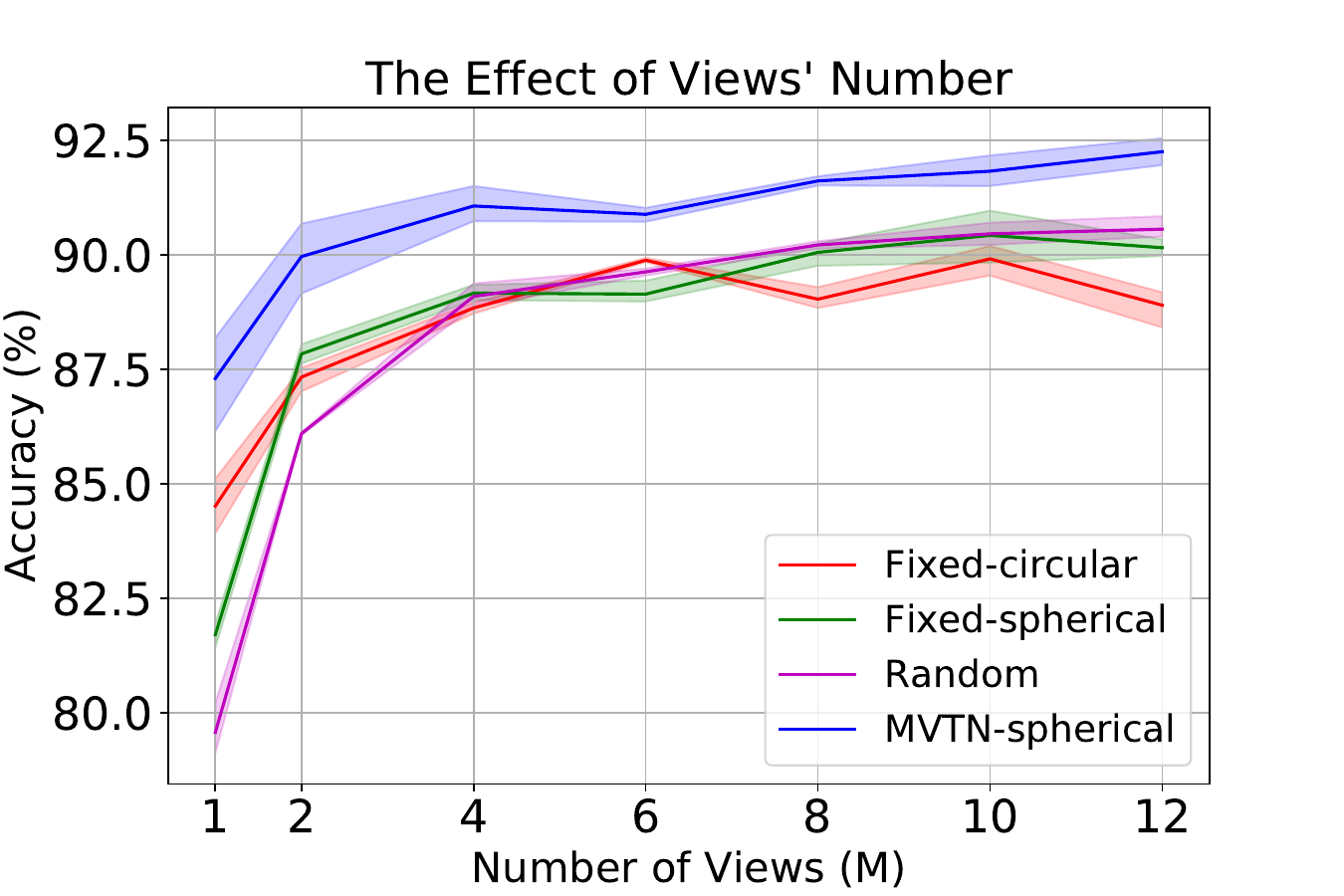}
    \caption{\textbf{Effect of the Number of Views.} We plot the test accuracy \vs the number of views (M) used to train MVCNN on fixed, random, and learned MVTN view configurations. %
    We observe a consistent 2\% improvement with MVTN over a variety of views.}
    \label{fig:classification}
\end{figure}

\mysection{Number of Views} \label{sec:views}
We study the effect of the number of views $M$ on the performance of MVCNN when using fixed views (circular/spherical), learned views (MVTN), and random views.
The experiments are repeated four times, and the average test accuracies with confidence intervals are shown in  \figLabel{\ref{fig:classification}}. 
 The plots show how learned MVTN-spherical achieves consistently superior performance across a different number of views. 

\mysection{Choice of Backbone and Point Encoders}
In all of our main MVTN experiments, we use ResNet-18 as our backbone and PointNet as the point feature extractor. However, different choices could be made for both.
We explore using DGCNN \cite{dgcn} as an alternative point encoder and ResNet-34 as an alternative 2D backbone in ViewGCN. We report all MVTN ablation results in Table \ref{tbl:ablation}. We observe diminishing returns for making the CNN backbone and the shape feature extractor more complex in the MVTN setup, which justifies using the simpler combination in our main experiments

\begin{table}[t]
\footnotesize
\setlength{\tabcolsep}{6pt} %
\renewcommand{\arraystretch}{1.1} %
\centering
\resizebox{0.95\hsize}{!}{
\begin{tabular}{c|c|c|c} 
\toprule
 \multicolumn{1}{c|}{\textbf{Backbone}}& \multicolumn{1}{c|}{\textbf{Point}}& \multicolumn{1}{c|}{\textbf{MVTN}}&   \multicolumn{1}{c}{\textbf{Results}}\\
 \textbf{Network} & \textbf{Encoder}  & \textbf{Setup} &  \textbf{Accuracy}     \\
\midrule
 \multirow{4}{*}{ResNet-18} & \multirow{2}{*}{PointNet} &  circular  &  92.83 $\pm$ 0.06 \\ 
  & &  spherical  &  93.41 $\pm$ 0.13  \\ 
 & \multirow{2}{*}{DGCNN} & circular  &  93.03 $\pm$ 0.15  \\ 
 & &  spherical   &  93.26 $\pm$ 0.04 \\  \hline

   \multirow{4}{*}{ResNet-34} & \multirow{2}{*}{PointNet} &  circular &  92.72 $\pm$ 0.16 \\ 
   &  &  spherical  &  92.83 $\pm$ 0.12 \\ 
   & \multirow{2}{*}{DGCNN} & circular  & 92.72 $\pm$ 0.03 \\ 
   &  &  spherical   &  92.63 $\pm$ 0.15 \\ 
 \bottomrule
\end{tabular}
}
\vspace{2pt}
\caption{\small \textbf{Ablation Study}. We analyze the effect of ablating different MVTN components on test accuracy in ModelNet40. Namely, we observe that using deeper backbone CNNs or a more complex point encoder do not increase the test accuracy.}
\label{tbl:ablation}
\end{table}

\mysection{Choice of Multi-View Network}
MVTN integrates smoothly with different multi-view networks and always leads to performance boost. In Table \ref{tbl:combine}. we show the overall accuracies (averaged over four runs) on ModelNet40 of 12 views when fixed views are used versus when MVTN is used with different multi-view networks.

\begin{table}[t]

\tabcolsep=0.11cm
\centering
\resizebox{0.99\linewidth}{!}{
\begin{tabular}{c|c|c|c} 
\toprule
\textbf{View Selection} & \multicolumn{3}{c}{\textbf{Multi-View Networks}} \\
& MVCNN\cite{mvcnn}  &  RotNet\cite{mvrotationnet}  & ViewGCN\cite{mvviewgcn}\\
\midrule
fixed views & 90.4 &  91.6  & 93.0  \\ 
with MVTN& \textbf{92.6} &   \textbf{93.2} &  \textbf{93.8}    \\
\bottomrule
\end{tabular}
}
\vspace{2pt}
\caption{\small \textbf{Integrating MVTN with Multi-View Networks}.
We show overall classification accuracies on ModelNet40 with 12 views on different multi-view networks when fixed views are used versus when MVTN is used. 
}
\label{tbl:combine}
\end{table}

\mysection{Other Factors Affecting MVTN} 
We study the effect of the light direction in the renderer, the camera's distance to the object, and the object's color.
We also study the transferability of the learned views from one multi-view network to another, and the performance of MVTN variants. More details are provided in the \supp\hspace{-2pt}.

\subsection{Time and Memory Requirements}
\vspace{-4pt}
We compare the time and memory requirements of different parts of our 3D recognition pipeline. We record the number of floating-point operations (GFLOPs) and the time of a forward pass for a single input sample. In Table \ref{tbl:speed}, MVTN contributes negligibly to the time and memory requirements of the multi-view networks. 

\begin{table}[t]

\tabcolsep=0.11cm
\centering
\resizebox{0.98\linewidth}{!}{
\begin{tabular}{r|ccc} 
\toprule
\multicolumn{1}{c}{Network} & GFLOPs & Time (ms) &   Parameters \# (M)\\
\midrule
MVCNN \cite{mvcnn} &  43.72 &  39.89 & 11.20 \\
ViewGCN \cite{mvviewgcn} &  44.19& 26.06 &   23.56   \\ \midrule
MVTN module &  \textbf{1.78}& \textbf{4.24}   & \textbf{3.5} \\
\bottomrule
\end{tabular}
}
\vspace{2pt}
\caption{\small \textbf{Time and Memory Requirements}. We assess the contribution of the MVTN module to the time and memory requirements in the multi-view pipeline. We note that the MVTN's time and memory requirements are negligible.}
\label{tbl:speed}
\end{table}

\section{Conclusions and Future Work} \label{sec:conclusion}
 Current multi-view methods rely on fixed views aligned with the dataset. We propose MVTN that learns to regress view-points for any multi-view network in a fully differentiable pipeline. MVTN harnesses recent developments in differentiable rendering and does not require any extra training supervision. Empirical results highlight the benefits of MVTN in 3D classification and 3D shape retrieval.  Some possible future works for MVTN include extending it to other 3D vision tasks such as shape and scene segmentation. Furthermore, MVTN can include more intricate scene parameters different from the camera view-points, such as light and textures. 
 
 \mysection{Acknowledgments} This work was supported by the King Abdullah University of Science and Technology (KAUST) Office of Sponsored Research through the Visual Computing Center (VCC) funding.

{\small
\bibliographystyle{ieee_fullname}
\bibliography{egbib}
}
\clearpage \clearpage
\appendix

\section{Detailed Experimental Setup}
\subsection{Datasets}
\begin{figure*} [h] 
\centering
\includegraphics[width = 0.9\linewidth]{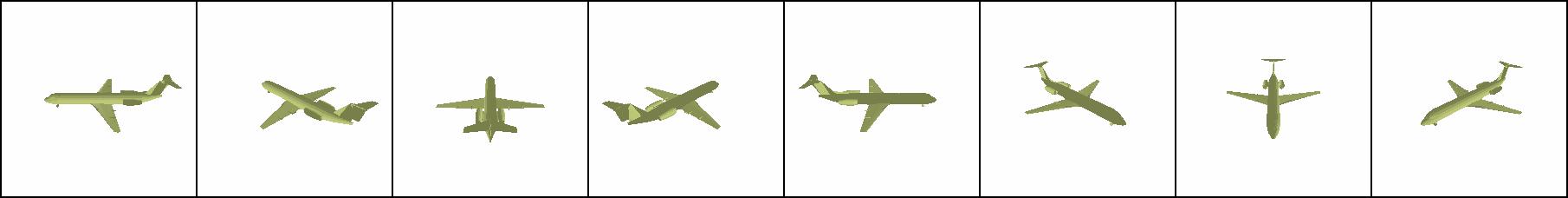} \\
\includegraphics[width = 0.9\linewidth]{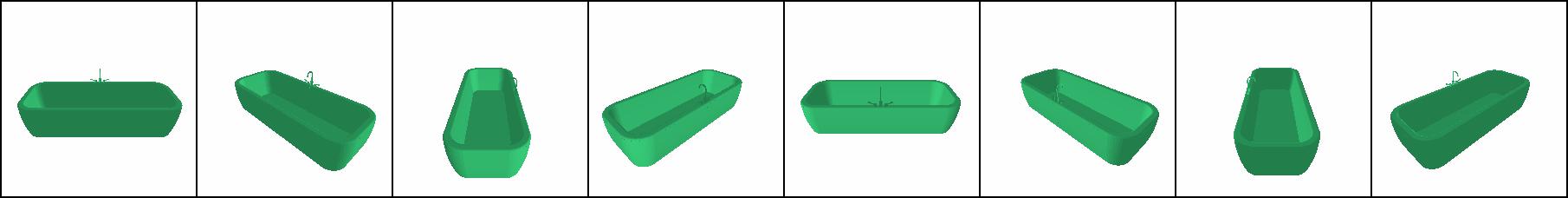} \\
\includegraphics[width = 0.9\linewidth]{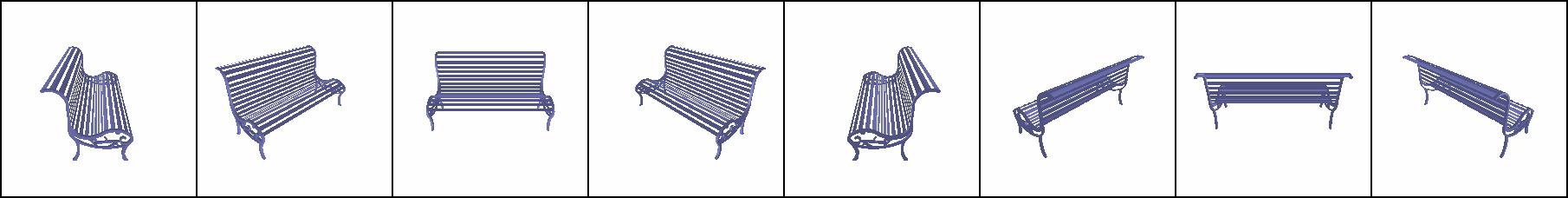} \\
\includegraphics[width = 0.9\linewidth]{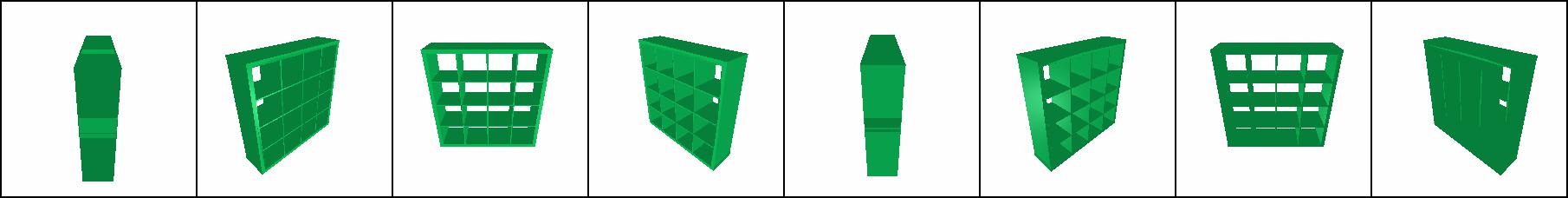} \\
\includegraphics[width = 0.9\linewidth]{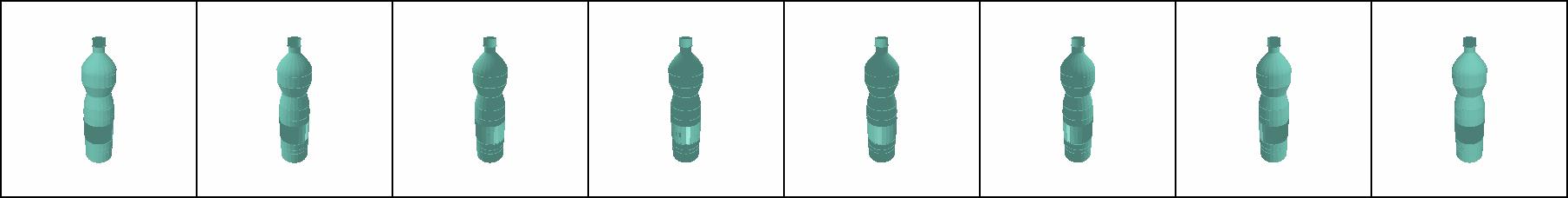} \\
\includegraphics[width = 0.9\linewidth]{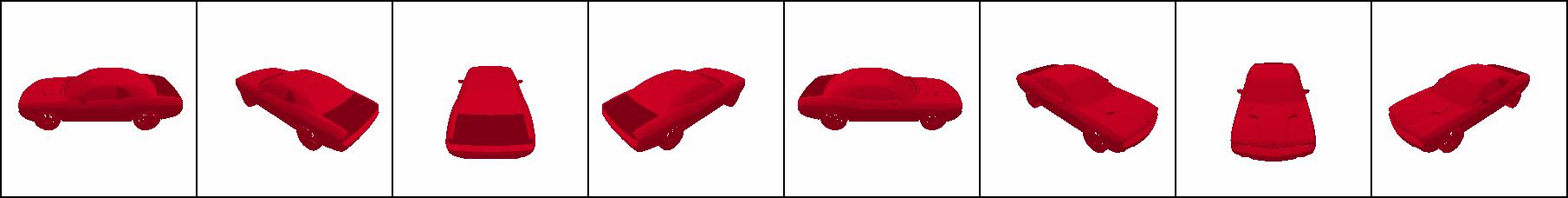} \\
\includegraphics[width = 0.9\linewidth]{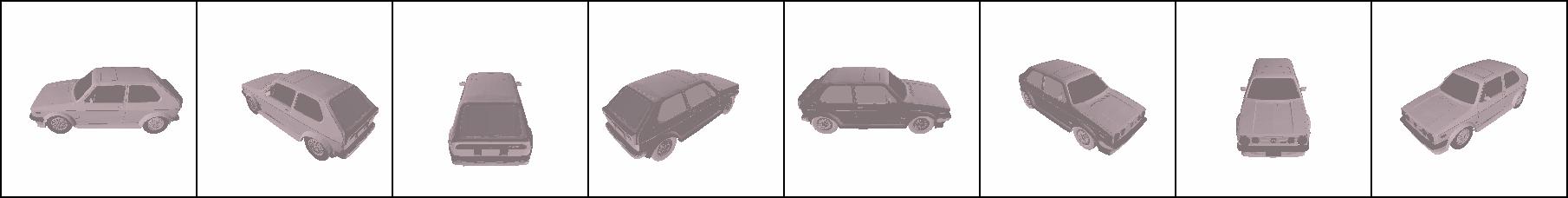} \\
\includegraphics[width = 0.9\linewidth]{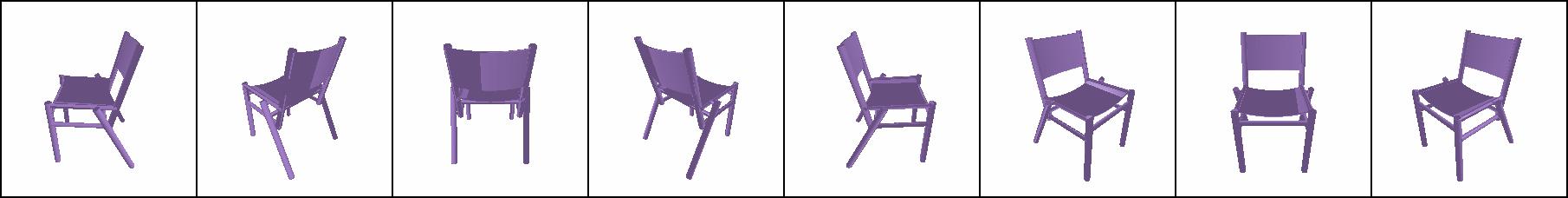} \\
\includegraphics[width = 0.9\linewidth]{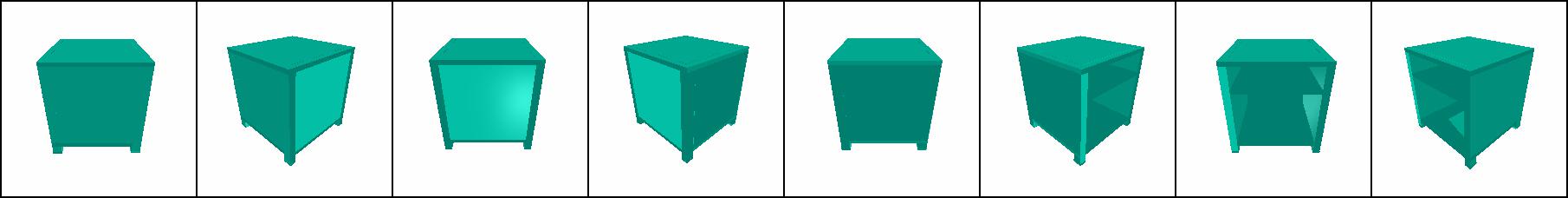} \\
\includegraphics[width = 0.9\linewidth]{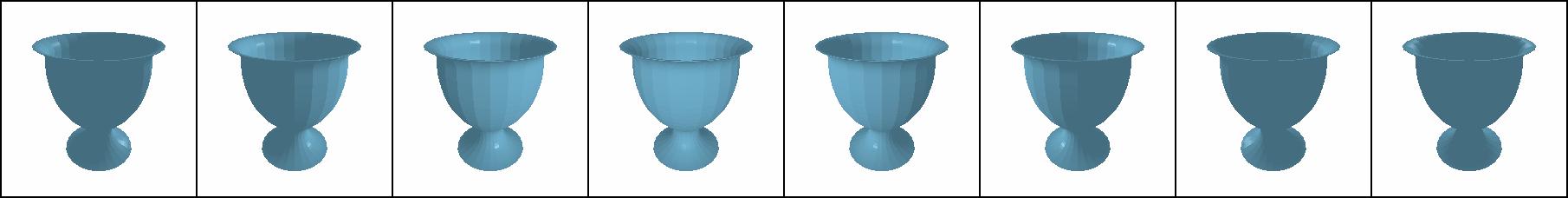} \\

\vspace{2pt}
\caption{\small \textbf{Training Data with Randomized Color and Lighting.} We show examples of mesh renderings of ModelNet40 used in training our MVTN. The color of the object and the light's direction are randomized during training for augmentation purposes and fixed in testing for stable performance. For this figure, eight circular views are shown for each 3D shape.  
}
    \label{fig:colorful-sup}
\end{figure*}

\mysection{ModelNet40}
We show in \figLabel{\ref{fig:colorful-sup}} examples of the mesh renderings of ModelNet40 used in training our MVTN. Note that the color of the object and the light direction are randomized in training for augmentation but are fixed in testing for stable performance.   

\begin{figure*}[t]
    \centering
    \includegraphics[width=0.98\linewidth]{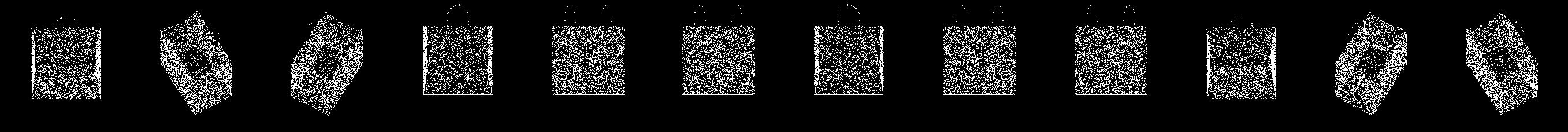}\\
    \includegraphics[width=0.98\linewidth]{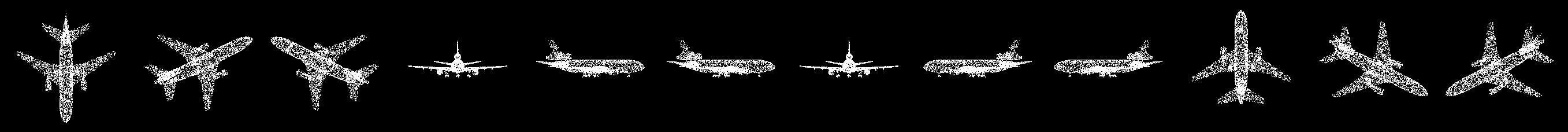}\\
    \includegraphics[width=0.98\linewidth]{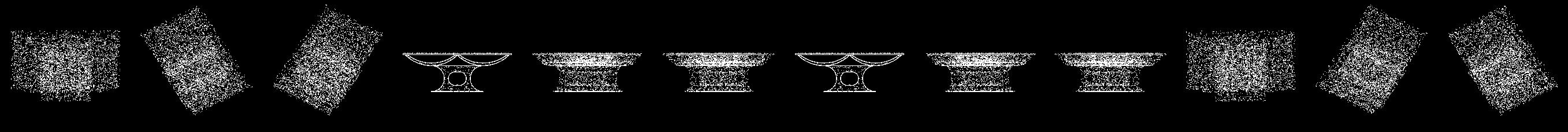}\\
    \includegraphics[width=0.98\linewidth]{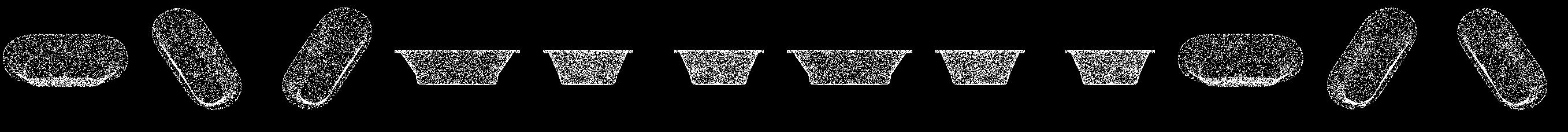}\\
    \includegraphics[width=0.98\linewidth]{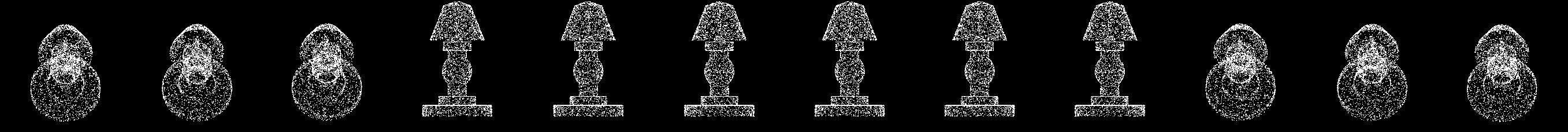}\\
    \includegraphics[width=0.98\linewidth]{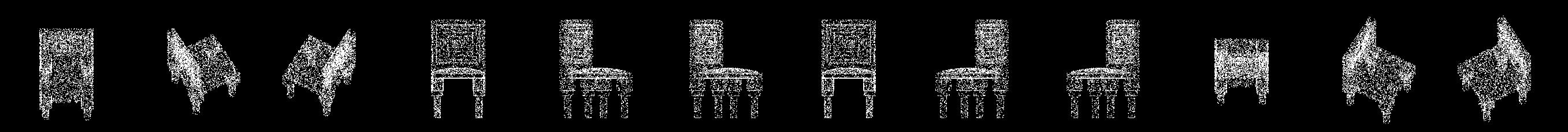}\\
    \includegraphics[width=0.98\linewidth]{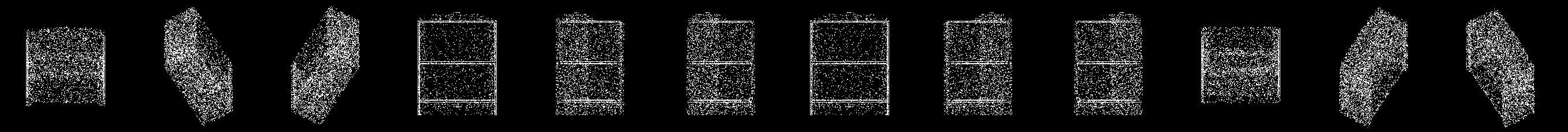}\\
    \includegraphics[width=0.98\linewidth]{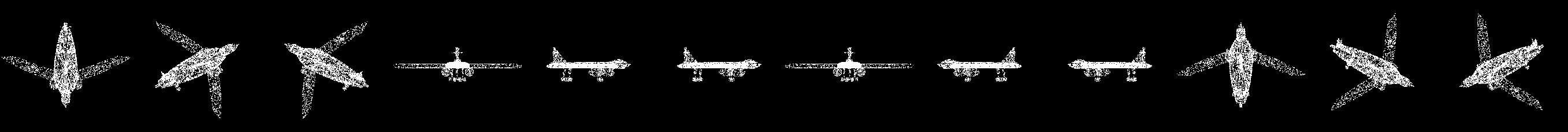}\\
    \includegraphics[width=0.98\linewidth]{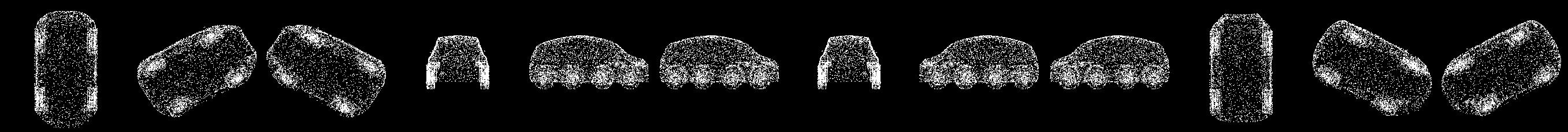}\\
    \includegraphics[width=0.98\linewidth]{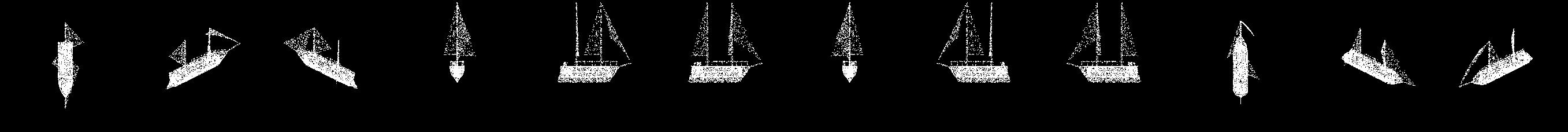}
    \caption{\textbf{ShapeNet Core55.} We show some examples of point cloud renderings of ShapeNet Core55 \cite{shapenet} used in training MVTN. Note how point cloud renderings offer more information about content hidden from the camera view-point (\eg car wheels from the occluded side), which can be useful for recognition. For this figure, 12 spherical views are shown for each 3D shape.}
    \label{fig:point-rendring-supp}
\end{figure*}

\mysection{ShapeNet Core55}
In \figLabel{\ref{fig:point-rendring-supp}}, we show  examples of the point cloud renderings of ShapeNet Core55 \cite{shapenet,shrek17} used in training MVTN. Note how point cloud renderings offer more information about content hidden from the camera view-point, which can be useful for recognition. White color is used in training and testing for all point cloud renderings. For visualization purposes, colors are inverted in the main paper examples (Fig. 4 in the main paper).

\begin{figure*} [h] 
\tabcolsep=0.03cm
\resizebox{0.98\linewidth}{!}{
\begin{tabular}{c|c}

\multirow{3}{*}{\textbf{Object Only}} &
\includegraphics[width = 0.7\linewidth]{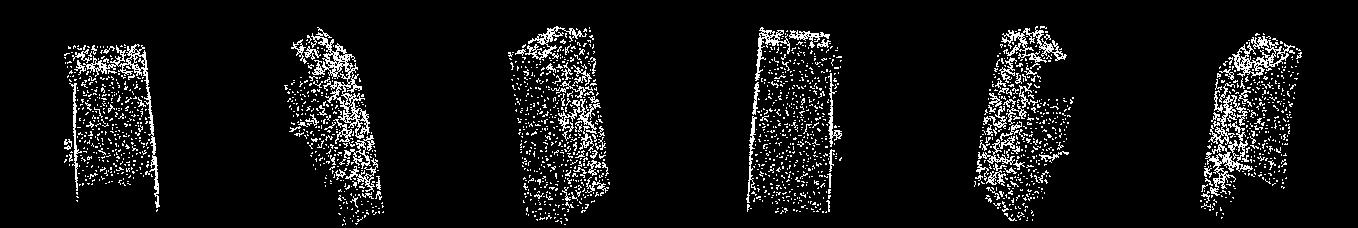} \\
&\includegraphics[width = 0.7\linewidth]{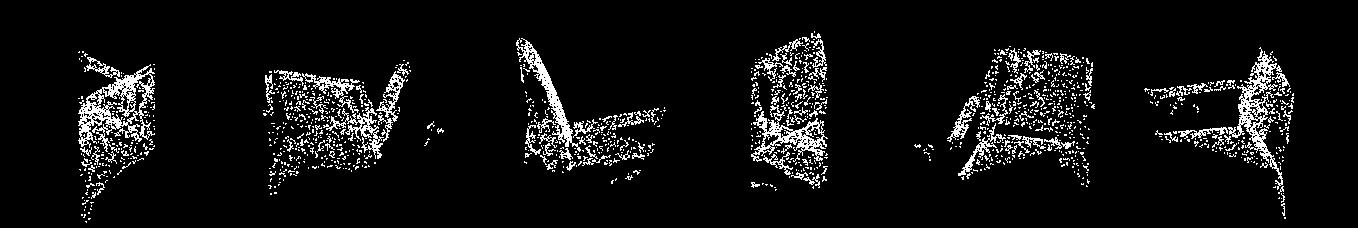} \\
&\includegraphics[width = 0.7\linewidth]{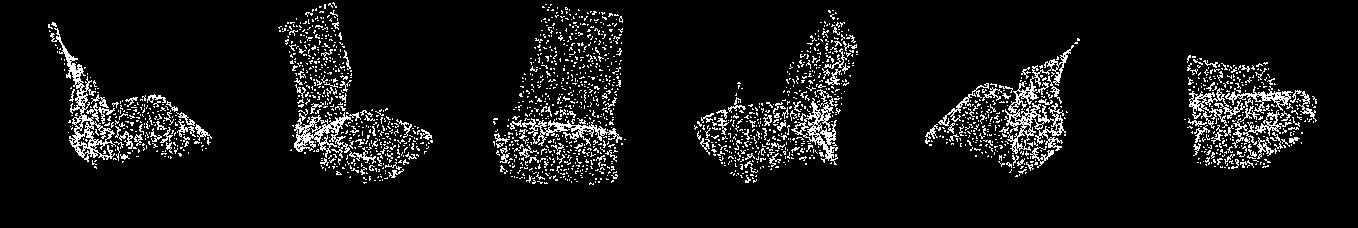} \\
\midrule
\multirow{3}{*}{\textbf{With Background}} &
\includegraphics[width = 0.7\linewidth]{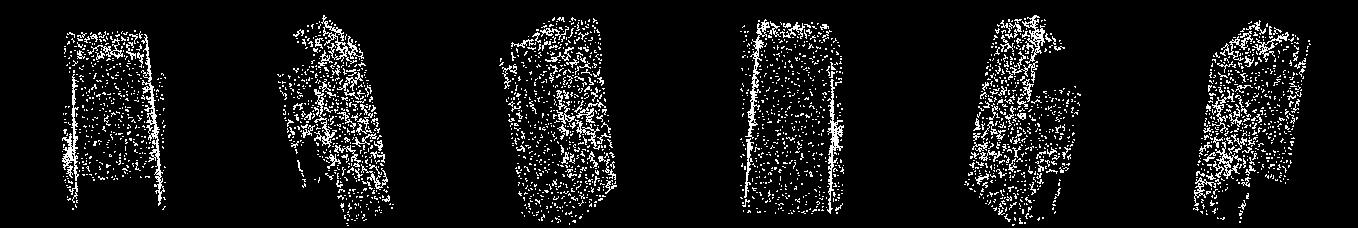} \\
&\includegraphics[width = 0.7\linewidth]{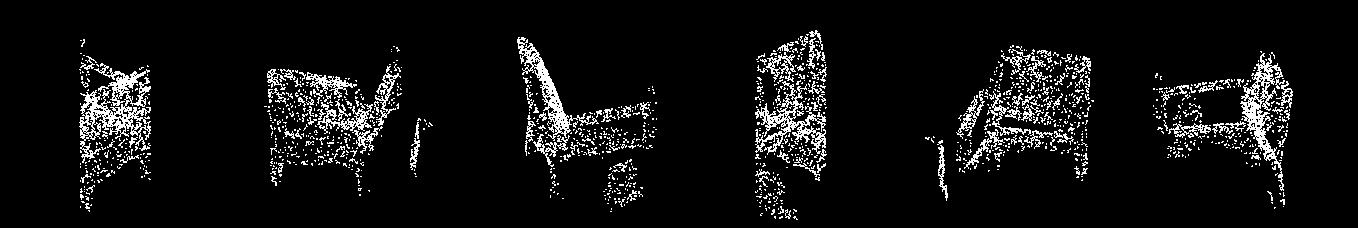} \\
&\includegraphics[width = 0.7\linewidth]{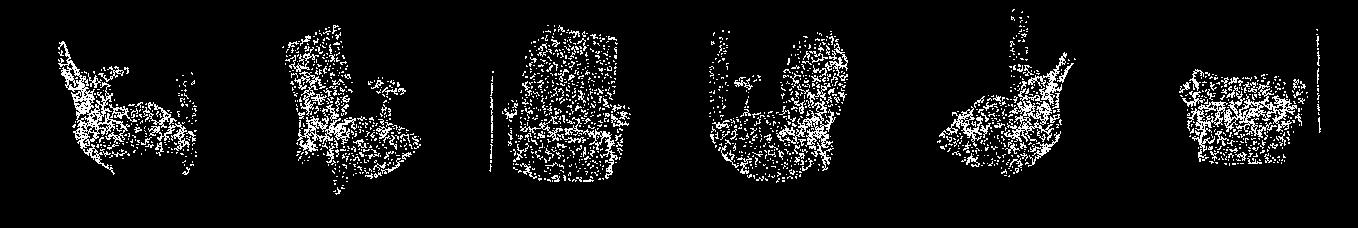} \\
\midrule
\multirow{3}{*}{\textbf{PB\_T50\_RS (Hardest)}} &
\includegraphics[width = 0.7\linewidth]{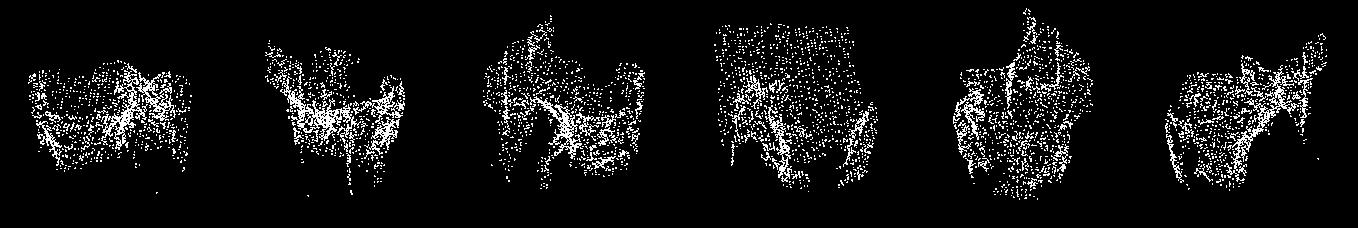} \\
&\includegraphics[width = 0.7\linewidth]{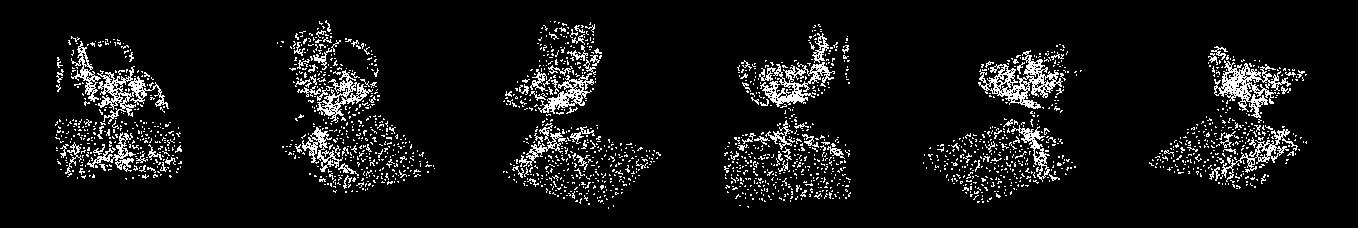} \\
&\includegraphics[width = 0.7\linewidth]{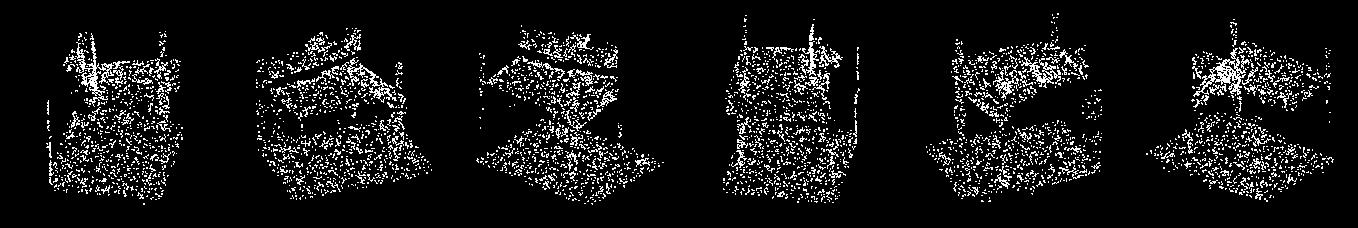} \\
\midrule

\end{tabular}
}
\vspace{2pt}
\caption{\small \textbf{ScanObjectNN Variants.} We show examples of point cloud renderings of different variants of the ScanObjectNN \cite{scanobjectnn} point cloud dataset used to train MVTN. The variants are: object only, object with background, and the hardest perturbed variant (with rotation and translation). For this figure, six circular views are shown for each 3D shape.     
}
    \label{fig:scanobjectnn-sup}
\end{figure*}

\mysection{ScanObjectNN}
ScanObjectNN \cite{scanobjectnn} has three main variants: object only, object with background, and the PB\_T50\_RS variant (hardest perturbed variant). \figLabel{\ref{fig:scanobjectnn-sup}} show examples of multi-view renderings of different samples of the dataset from its three variants. Note that adding the background points to the rendering gives some clues to our MVTN about the object, which explains why adding background improves the performance of MVTN in Table \ref{tab:Scanobjectnn-supp}. 

\subsection{MVTN Details}

\mysection{MVTN Rendering}
 Point cloud rendering offers a light alternative to mesh rendering in ShapeNet because its meshes contain large numbers of faces that hinders training the MVTN pipeline. Simplifying theses "high-poly" meshes (similar to ModelNet40) results in corrupted shapes that lose their main visual clues. Therefore, we use point cloud rendering for ShapeNet, allowing to process all shapes with equal memory requirements. Another benefit of point cloud rendering is making it possible to train MVTN with a large batch size on the same GPU (bath size of 30 on V100 GPU).    
 
\begin{figure*}
\centering
        \includegraphics[width=0.9\linewidth]{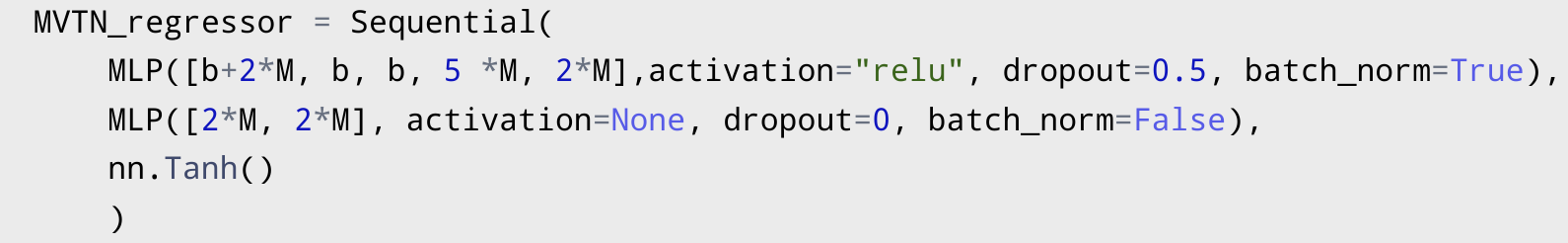} \vspace{8pt} \\
             \quad $b+2\times M$ \qquad\qquad \quad  $b$\qquad \qquad\qquad \qquad  $b$\qquad \qquad\qquad \quad $5\times M$ \qquad\qquad \qquad $2\times M$ \qquad \qquad  $2\times M$ \\
    \includegraphics[trim= 0 0 0 2cm , clip,width=0.9\linewidth]{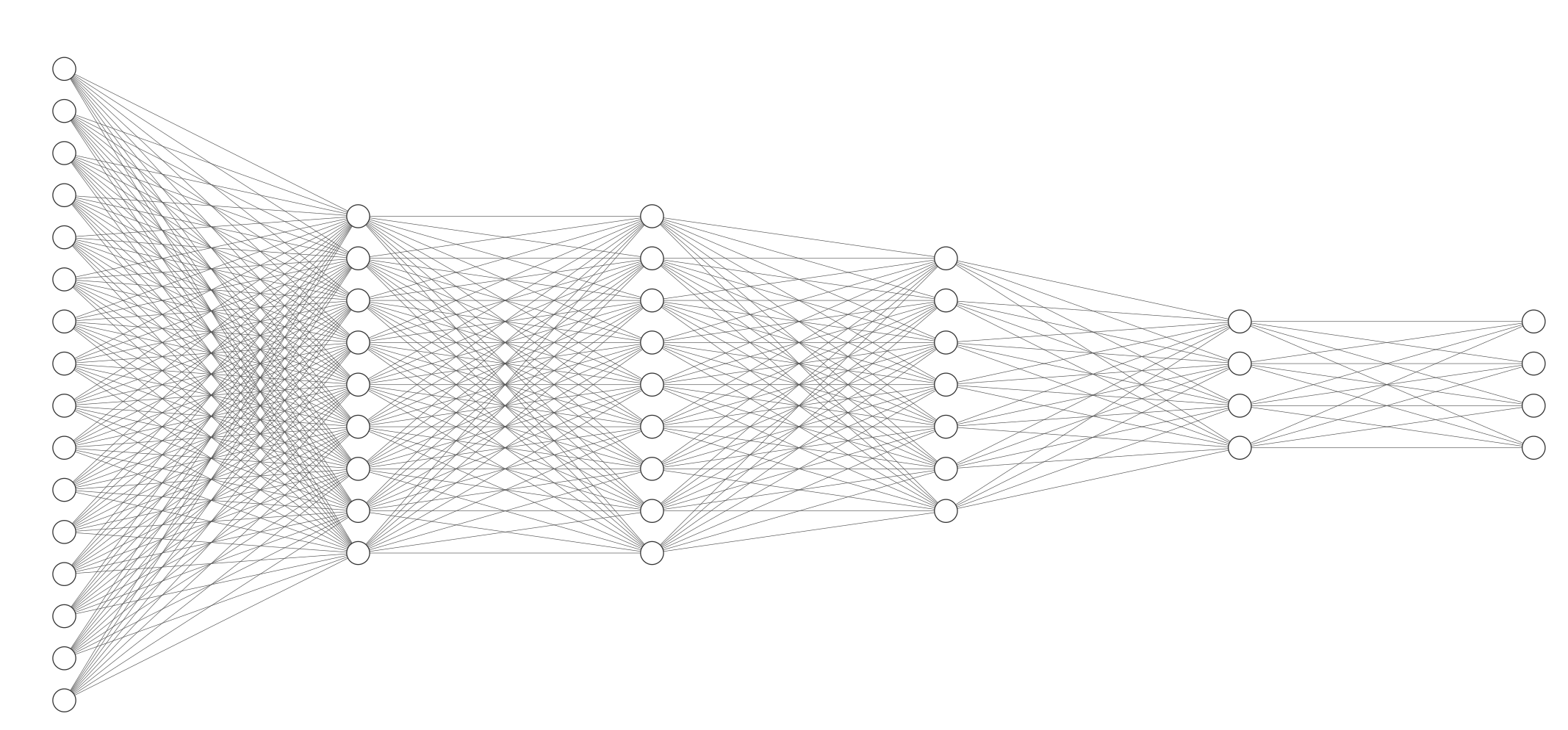} \\
    \caption{\textbf{MVTN Network Architecture.} We show a schematic and a code snippet for MVTN-spherical/MVTN-circular regression architectures used, where $b$ is the size of the point features extracted by the point encoder of MVTN and $M$ is the number of views learned. In most of our experiments, $b=40$, while the output is the azimuth and elevation angles for all the $M$ views used. The network is drawn using \cite{nndraw}}
    \label{fig:architecture-sup}
\end{figure*}

\mysection{MVTN Architecture}
We incorporate our MVTN into MVCNN \cite{mvcnn} and ViewGCN \cite{mvviewgcn}. 
In our experiments, we select PointNet \cite{pointnet} as the default point encoder of MVTN.
All MVTNs and their baseline multi-view networks use ResNet18 \cite{resnet} as backbone in our main experiments with output feature size $d=1024$. The azimuth angle maximum range ($\mathbf{u}_{\text{bound}}$) is $\frac{180^\circ}{M}$  for MVTN-circular and MVTN-spherical, while it is $180^\circ$ for MVTN-direct. On the other hand, the elevation angle maximum range ($\mathbf{u}_{\text{bound}}$) is $90^\circ$.
We use a 4-layer MLP for MVTN's regression network $\mathbf{G}$. For MVTN-spherical/MVTN-spherical, the regression network takes as input $M$ azimuth angles, $M$ elevation angles, and the point features of shape $\mathbf{S}$ of size $b=40$. The widths of the MVTN networks are illustrated in \figLabel{\ref{fig:architecture-sup}}.MVTN concatenates all of its inputs, and the MLP outputs the offsets to the initial $2\times M$ azimuth and elevation angles. The size of the MVTN network (with $b=40$) is $14 M^2 + 211 M + 3320$ parameters, where $M$ is the number of views. It is a shallow network of only around 9K parameters when $M=12$.  

\begin{figure*} [h] 
\tabcolsep=0.03cm
\begin{tabular}{c|ccccccc}
 & \textbf{1 view} & \textbf{2 views}& \textbf{4 views}& \textbf{6 views}& \textbf{8 views}& \textbf{10 views}& \textbf{12 views} \\
\textbf{(a)} & 
\includegraphics[trim= 4cm 2.7cm 4cm 2.2cm , clip, width = 0.135\linewidth]{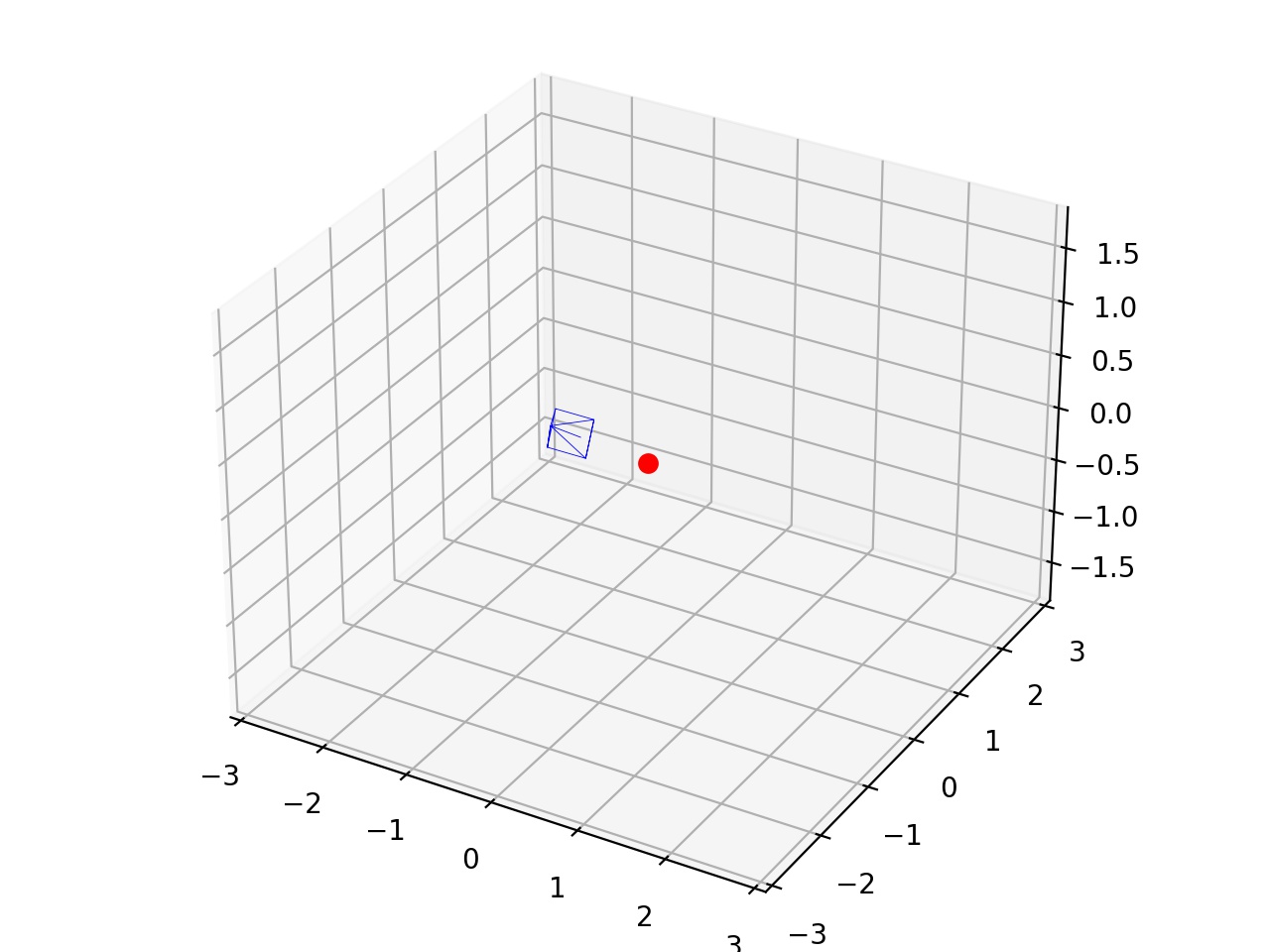} &
\includegraphics[trim= 4cm 2.7cm 4cm 2.2cm , clip, width = 0.135\linewidth]{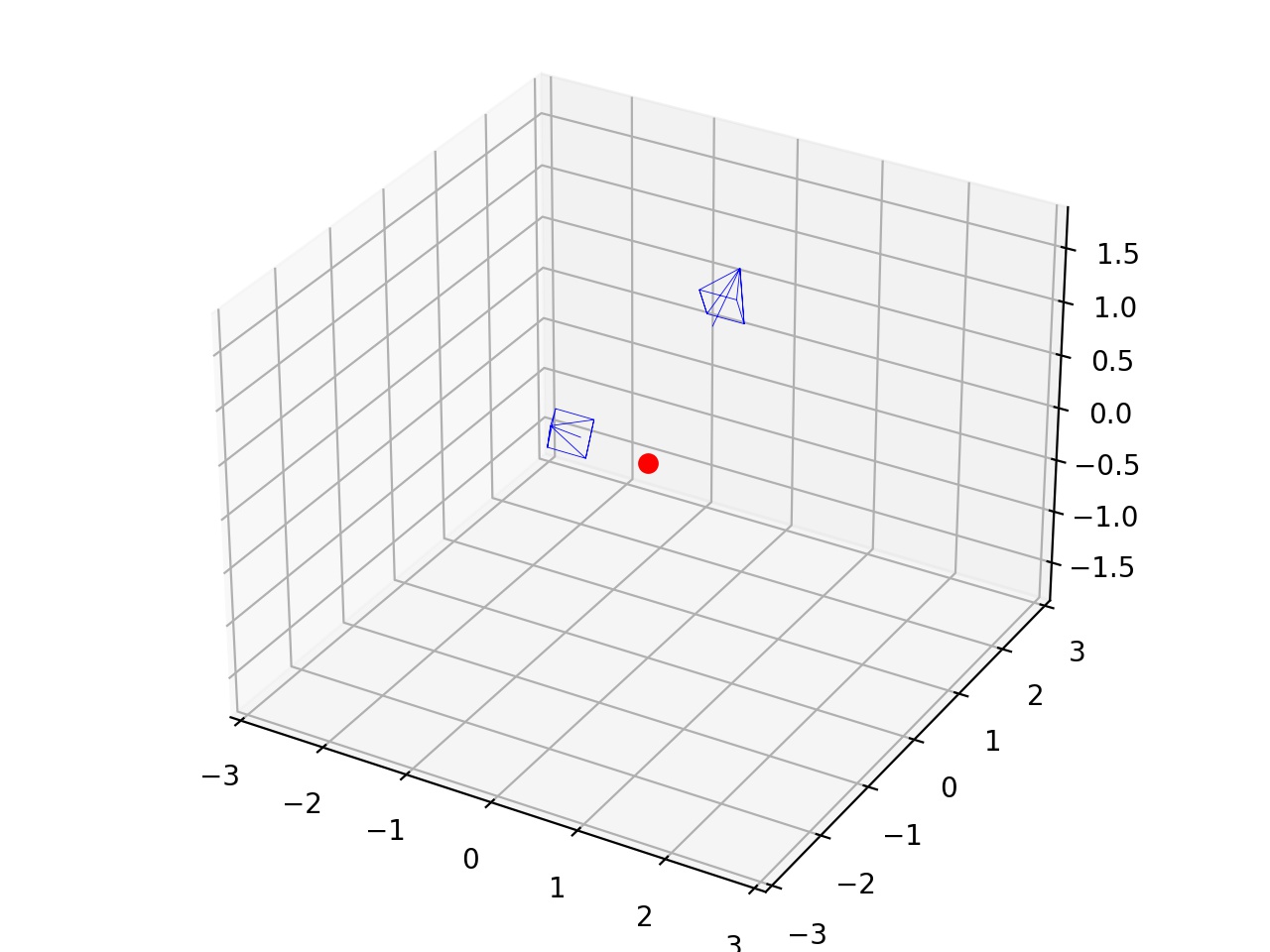} &
\includegraphics[trim= 4cm 2.7cm 4cm 2.2cm , clip, width = 0.135\linewidth]{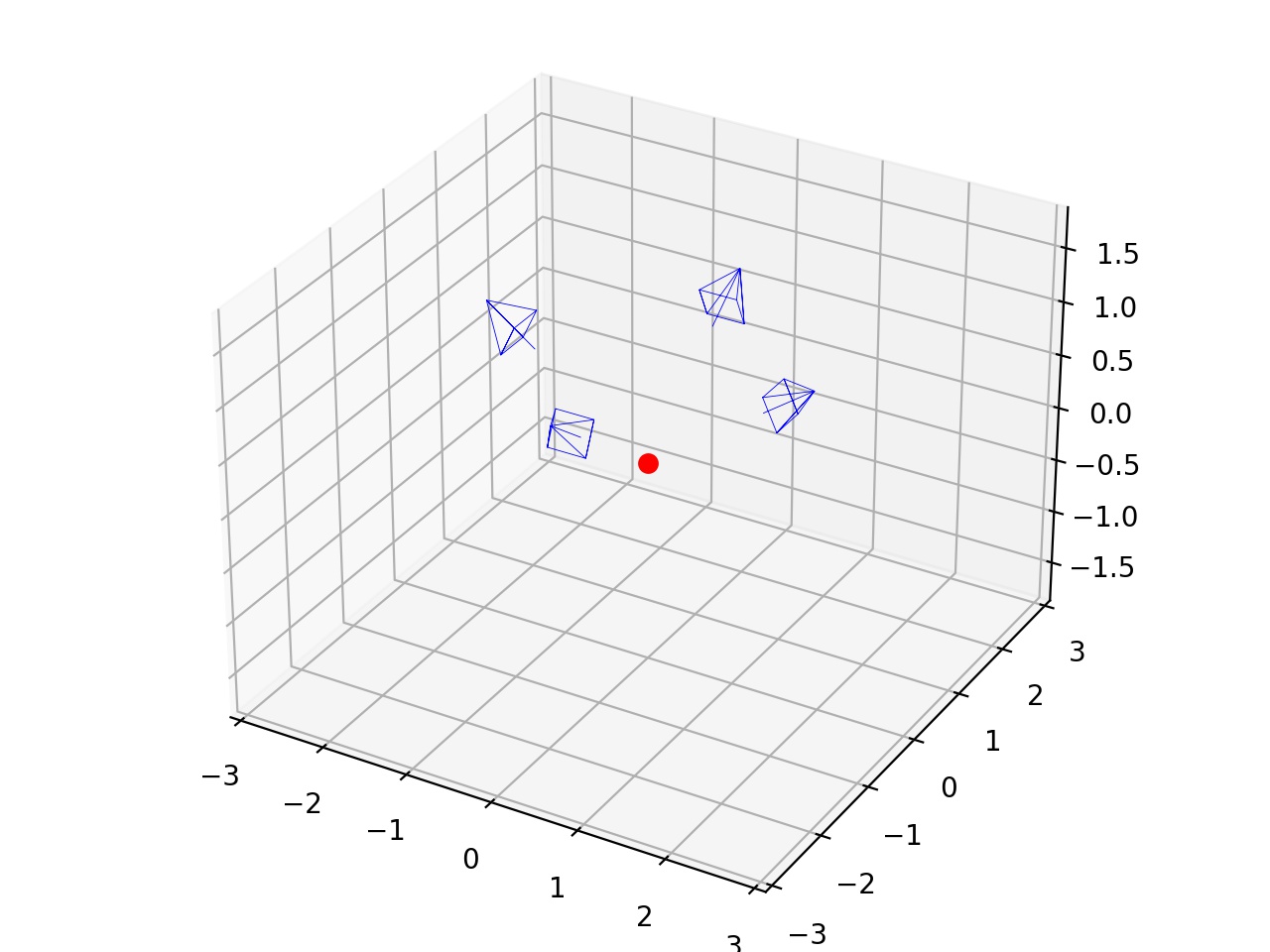} &
\includegraphics[trim= 4cm 2.7cm 4cm 2.2cm , clip, width = 0.135\linewidth]{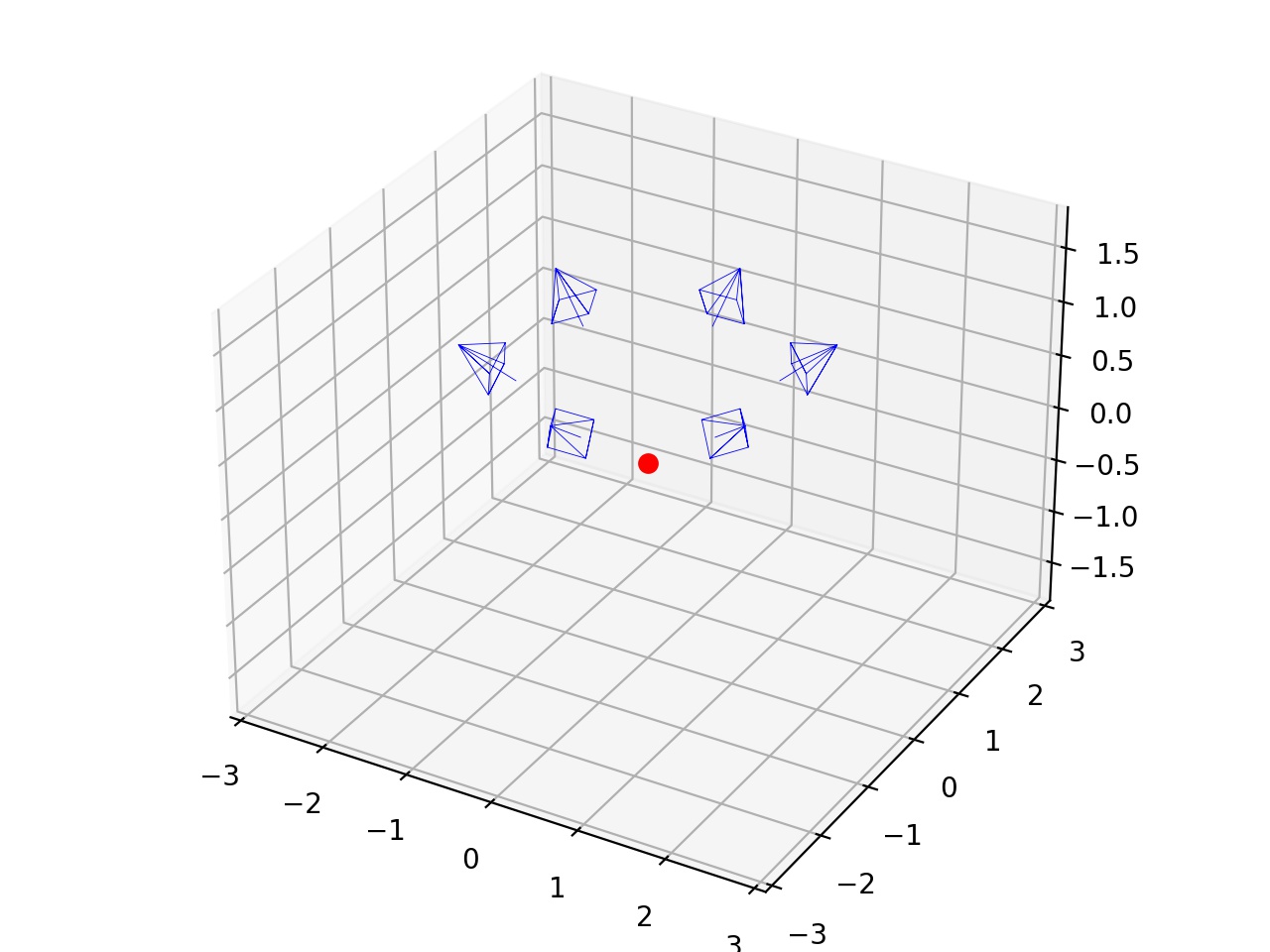} &
\includegraphics[trim= 4cm 2.7cm 4cm 2.2cm , clip, width = 0.135\linewidth]{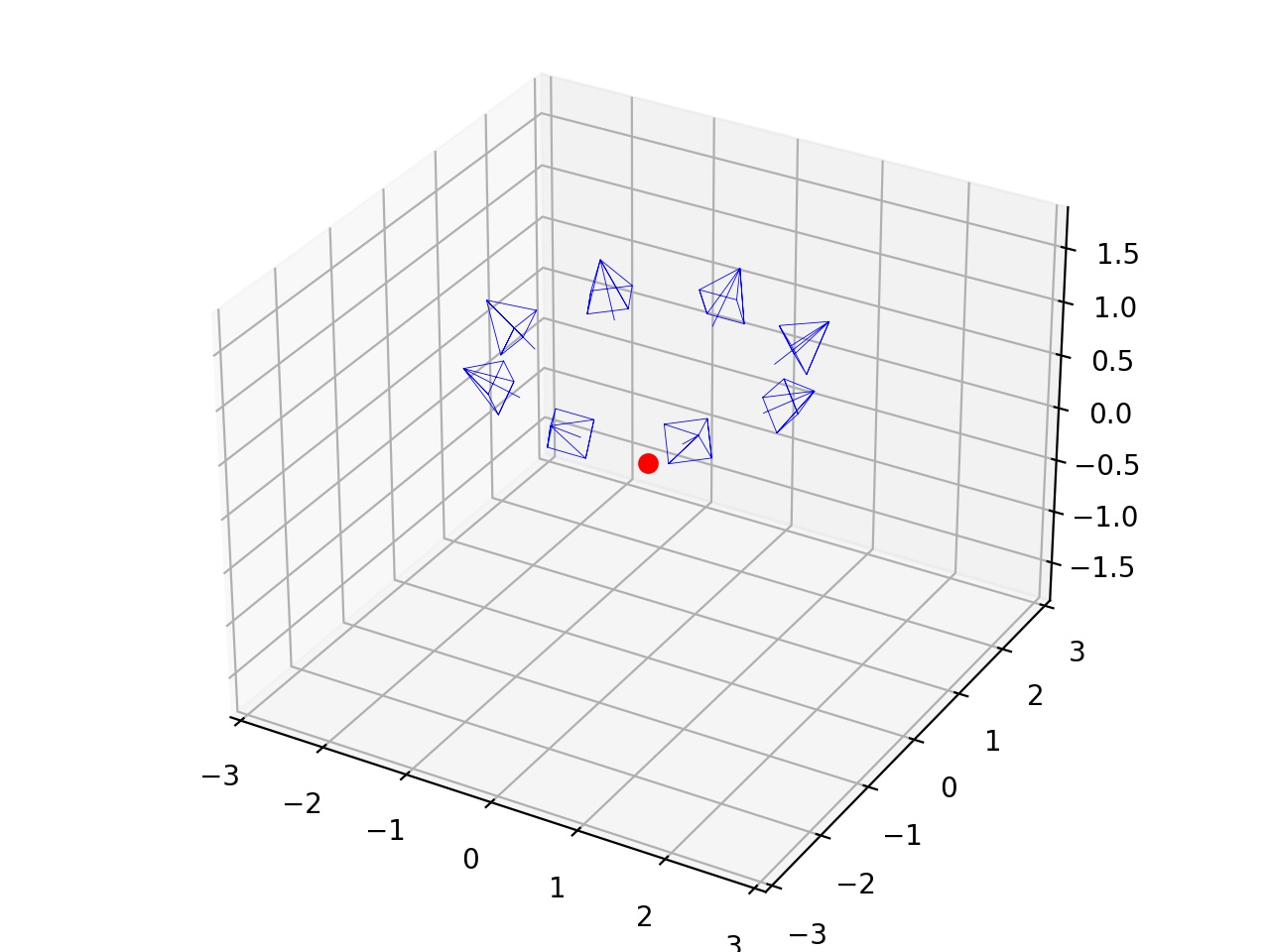} &
\includegraphics[trim= 4cm 2.7cm 4cm 2.2cm , clip, width = 0.135\linewidth]{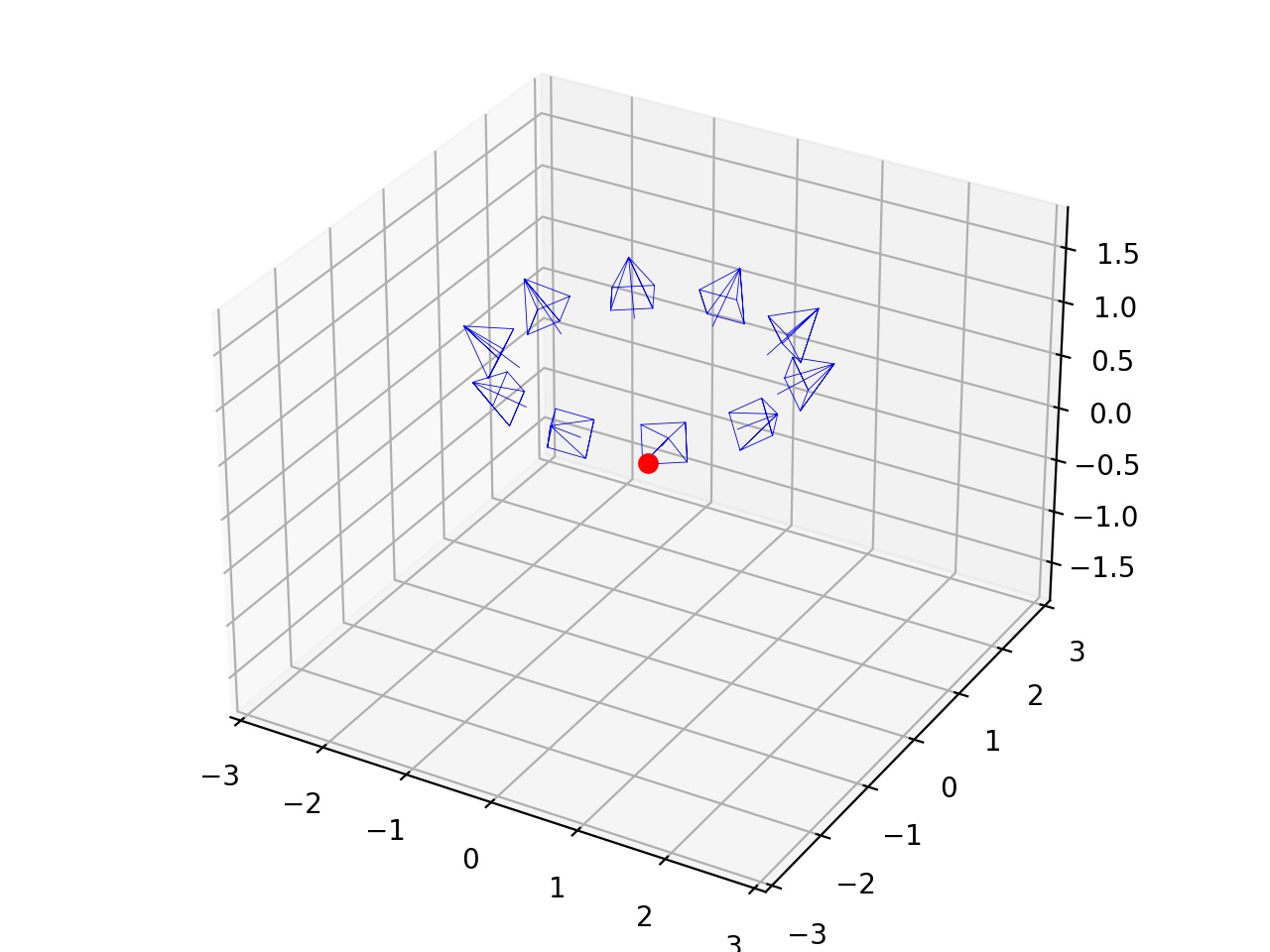} &
\includegraphics[trim= 4cm 2.7cm 4cm 2.2cm , clip, width = 0.135\linewidth]{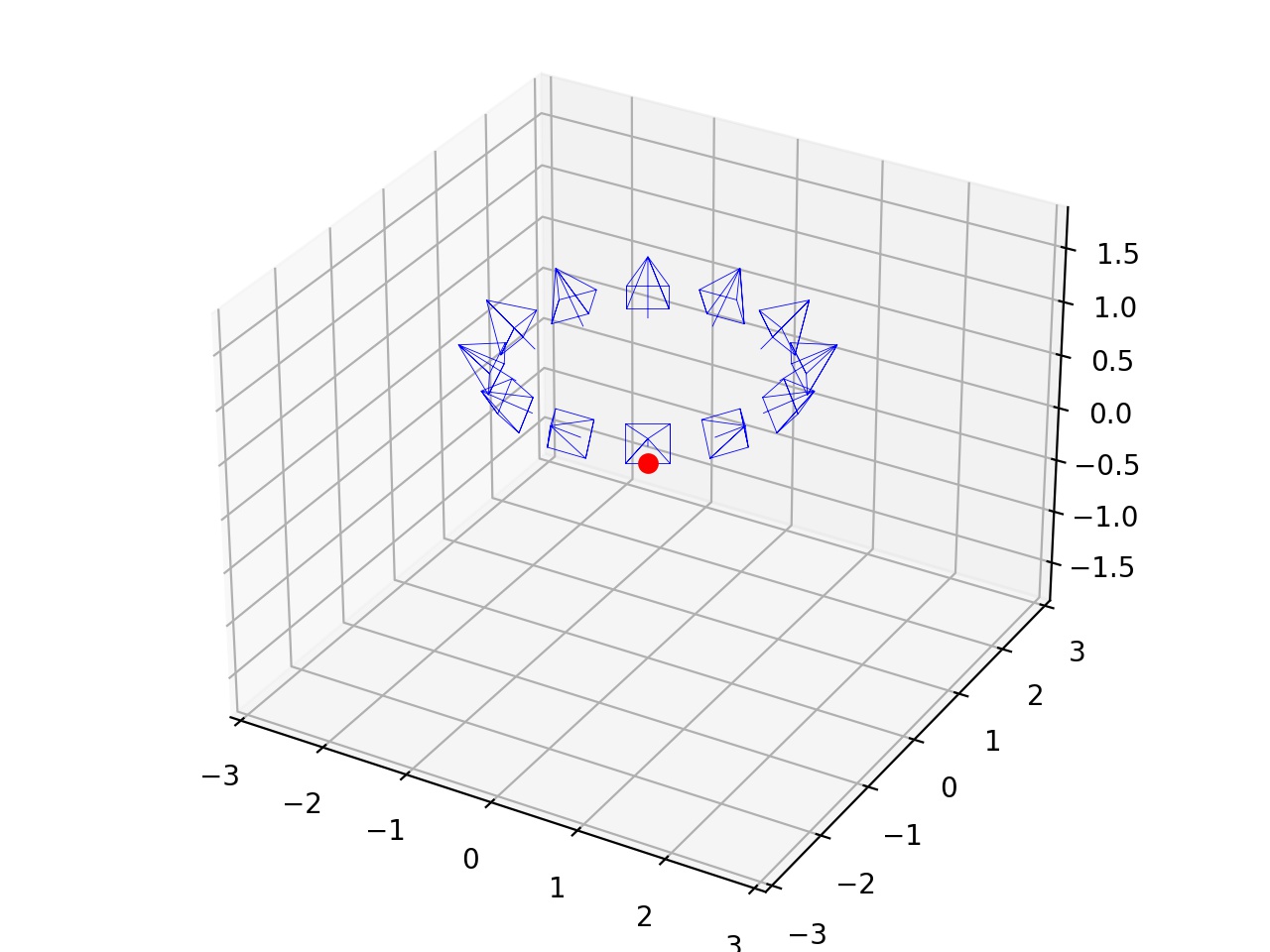} \\
\textbf{(b)} & 
\includegraphics[trim= 4cm 2.7cm 4cm 2.2cm , clip, width = 0.135\linewidth]{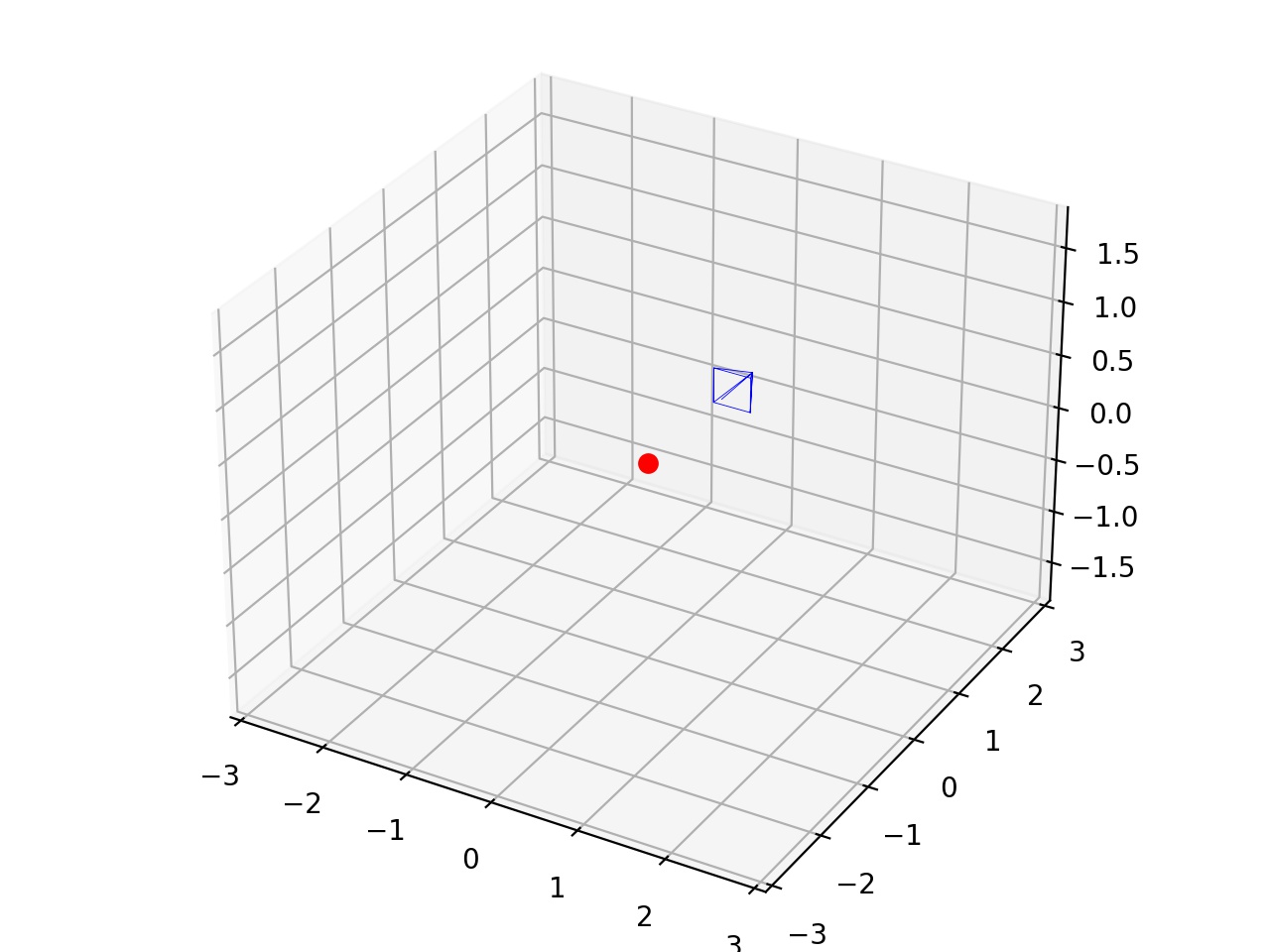} &
\includegraphics[trim= 4cm 2.7cm 4cm 2.2cm , clip, width = 0.135\linewidth]{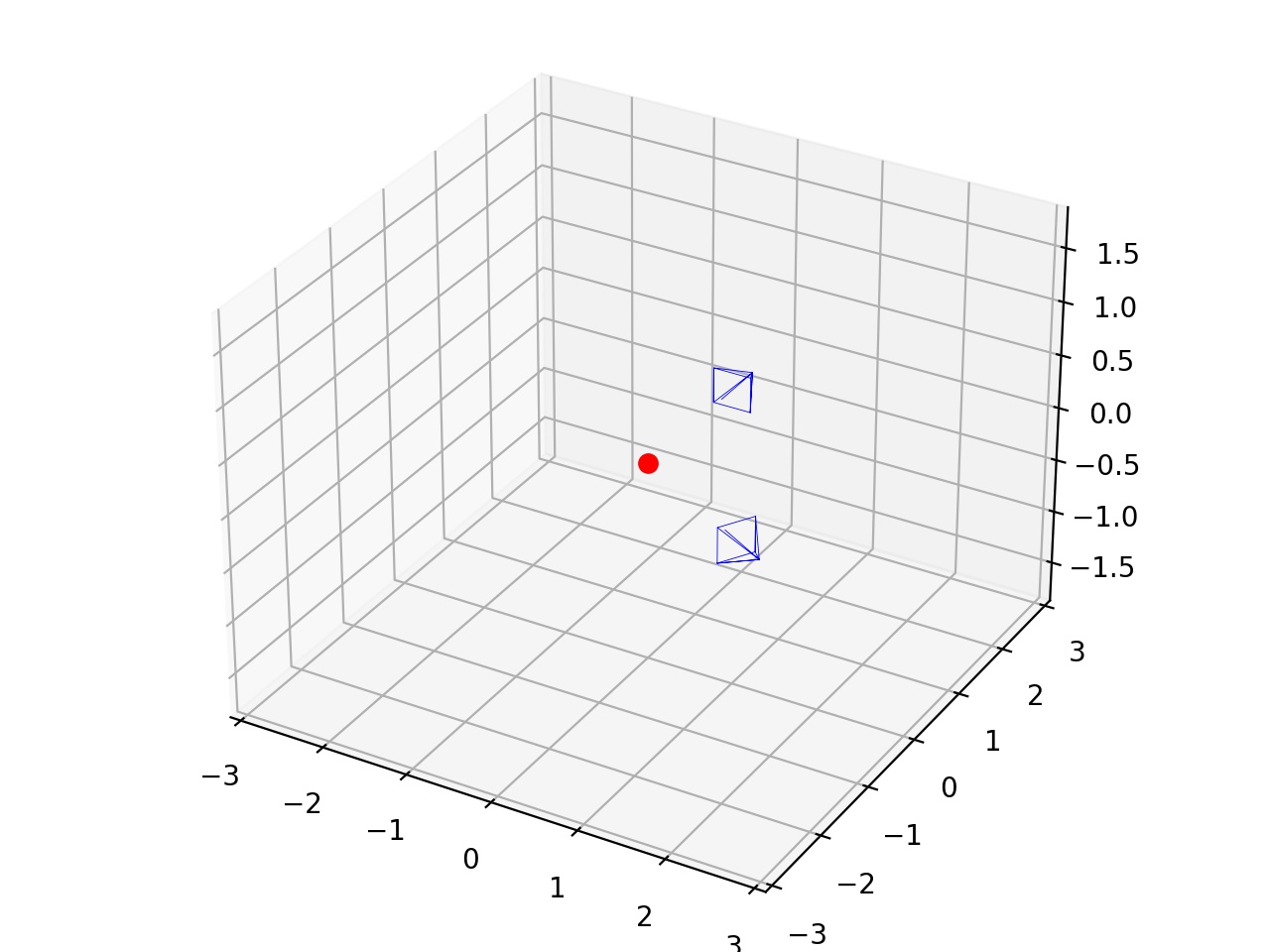} &
\includegraphics[trim= 4cm 2.7cm 4cm 2.2cm , clip, width = 0.135\linewidth]{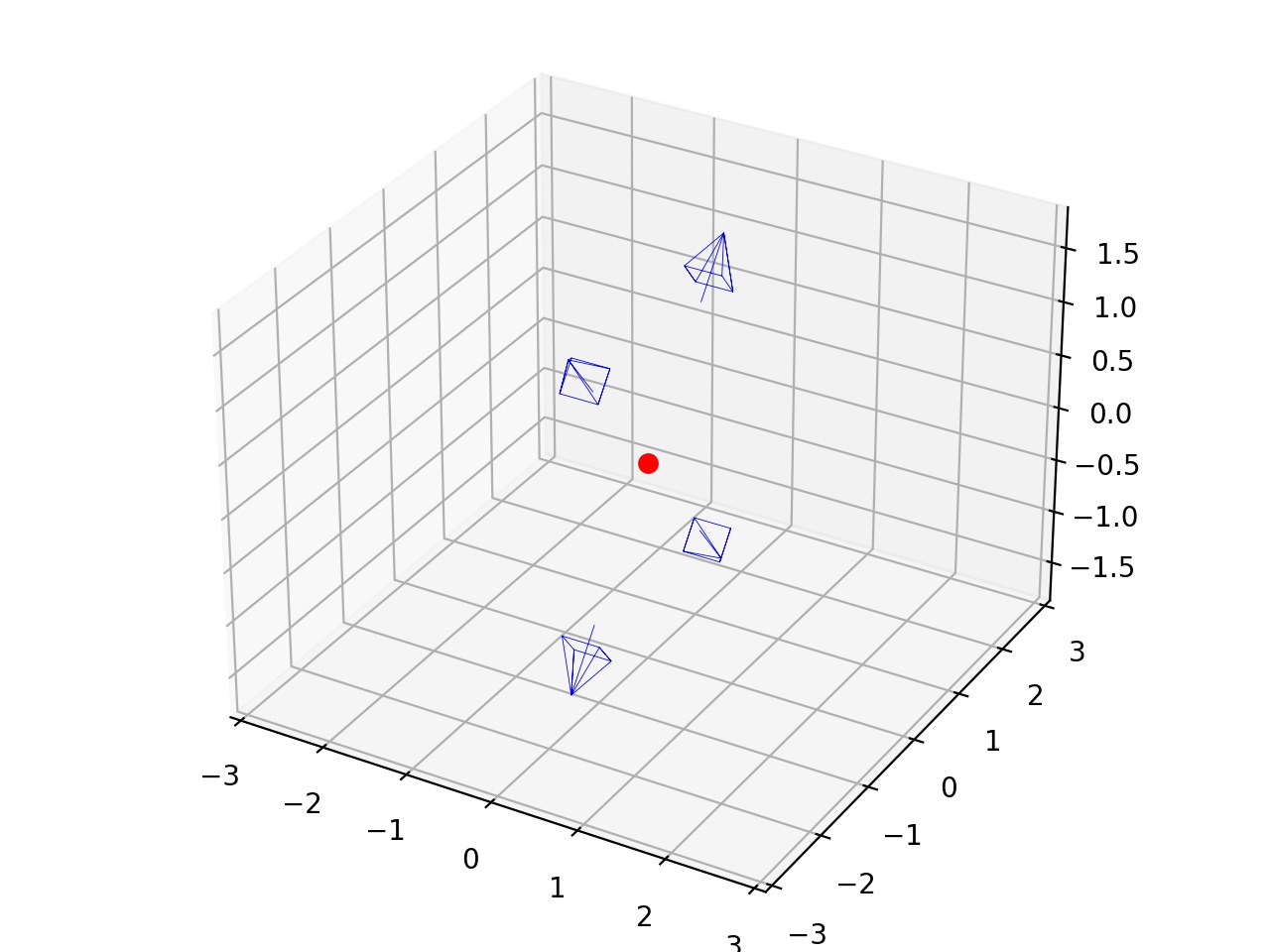} &
\includegraphics[trim= 4cm 2.7cm 4cm 2.2cm , clip, width = 0.135\linewidth]{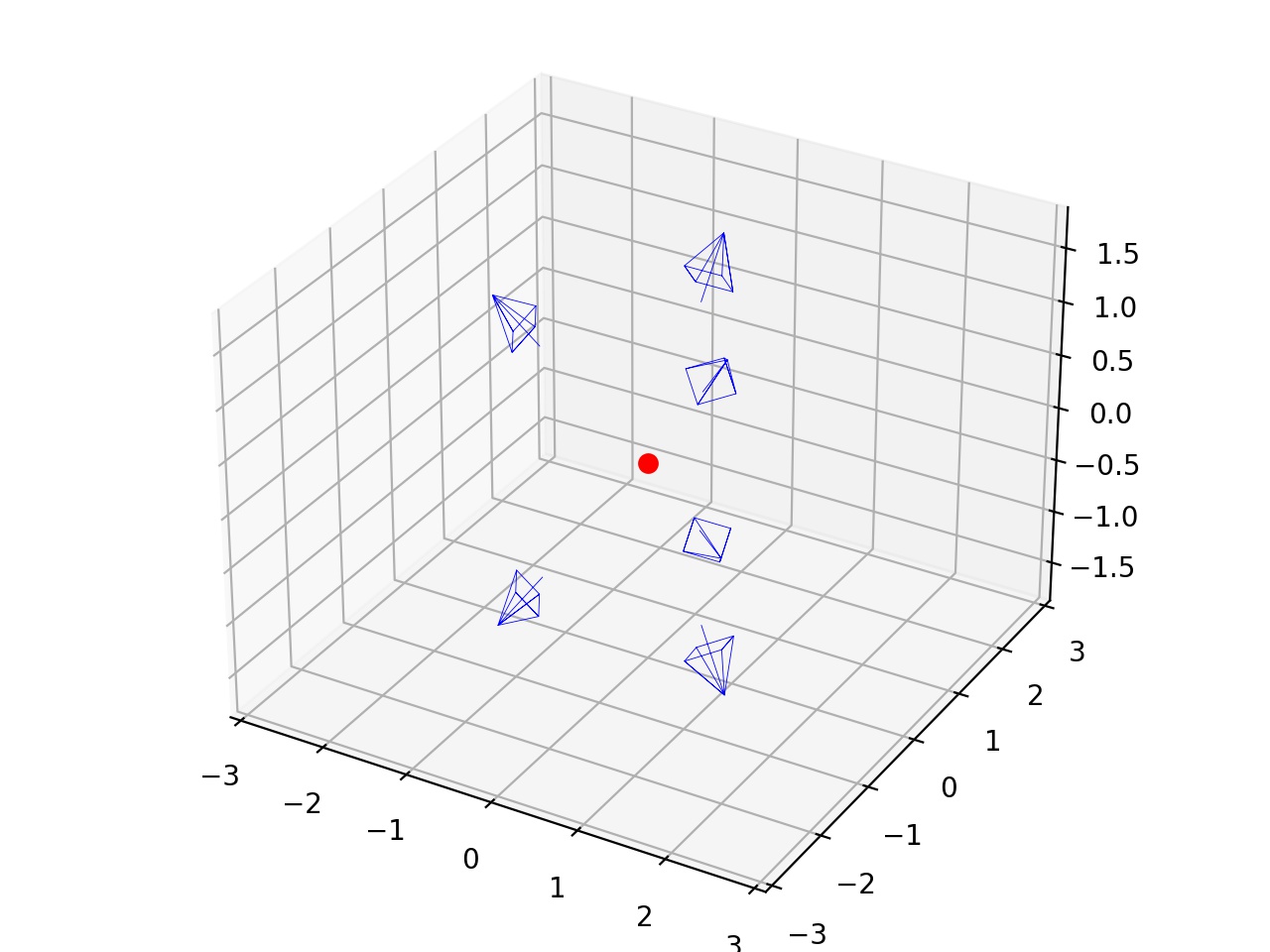} &
\includegraphics[trim= 4cm 2.7cm 4cm 2.2cm , clip, width = 0.135\linewidth]{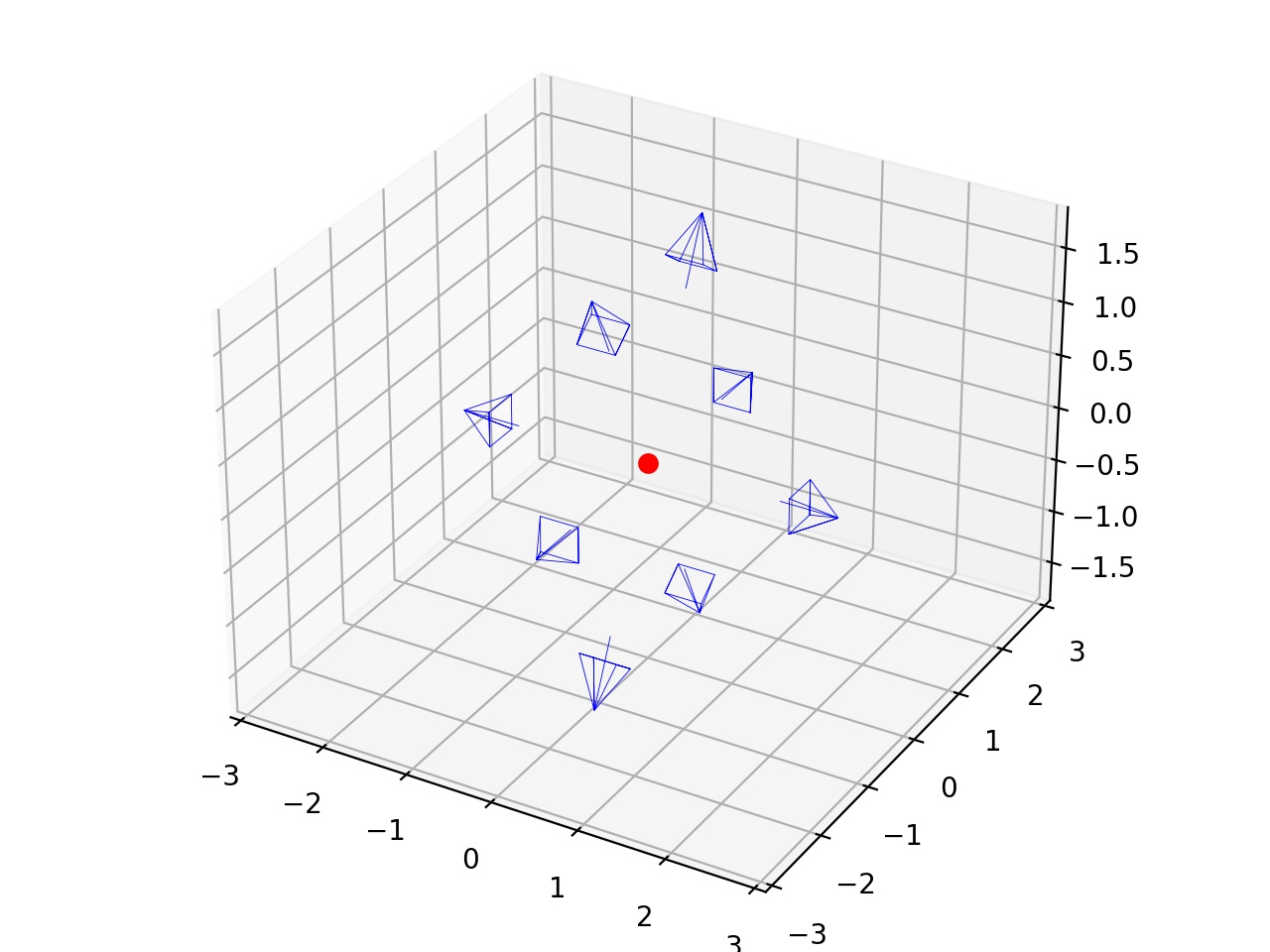} &
\includegraphics[trim= 4cm 2.7cm 4cm 2.2cm , clip, width = 0.135\linewidth]{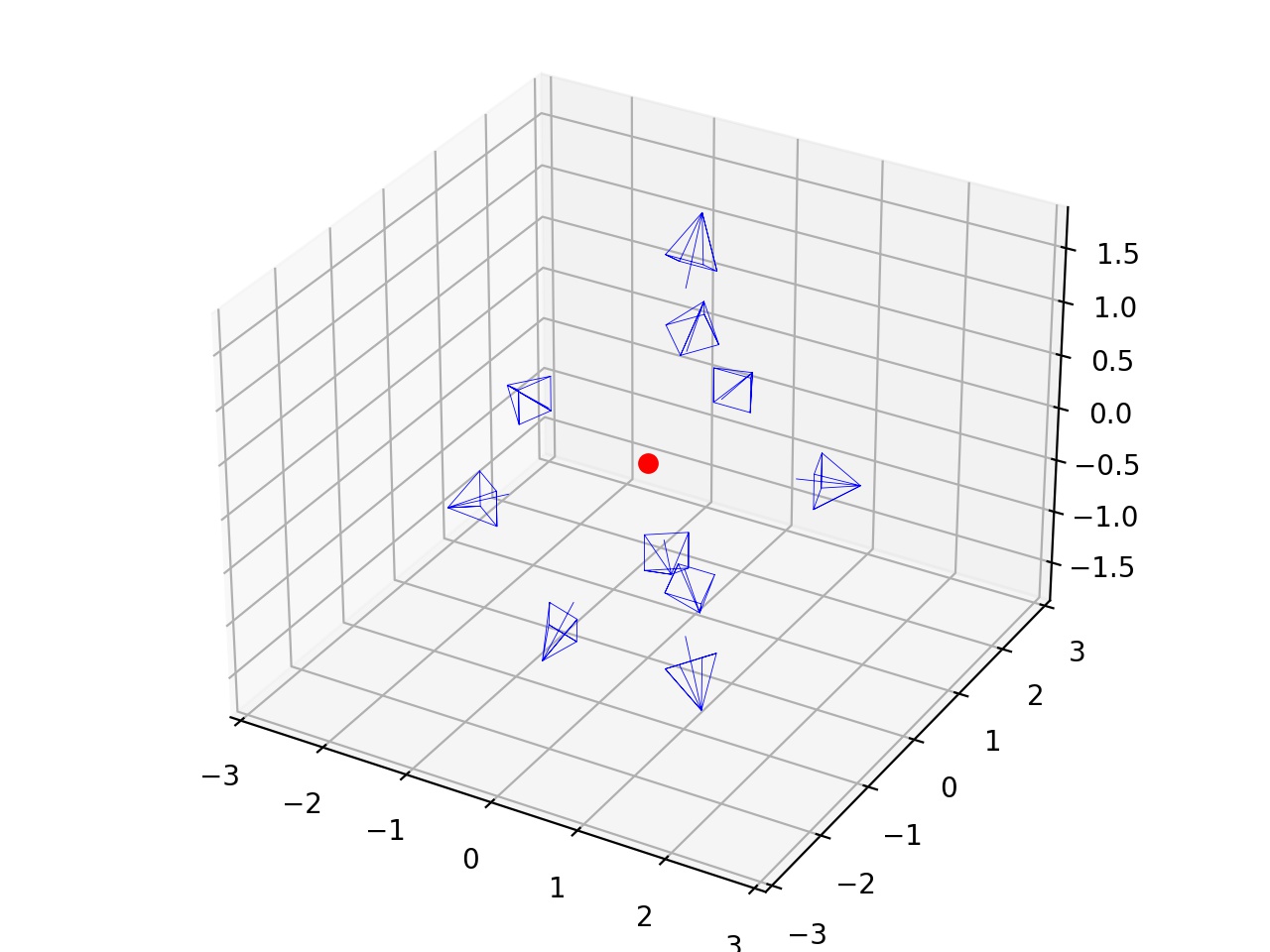} &
\includegraphics[trim= 4cm 2.7cm 4cm 2.2cm , clip, width = 0.135\linewidth]{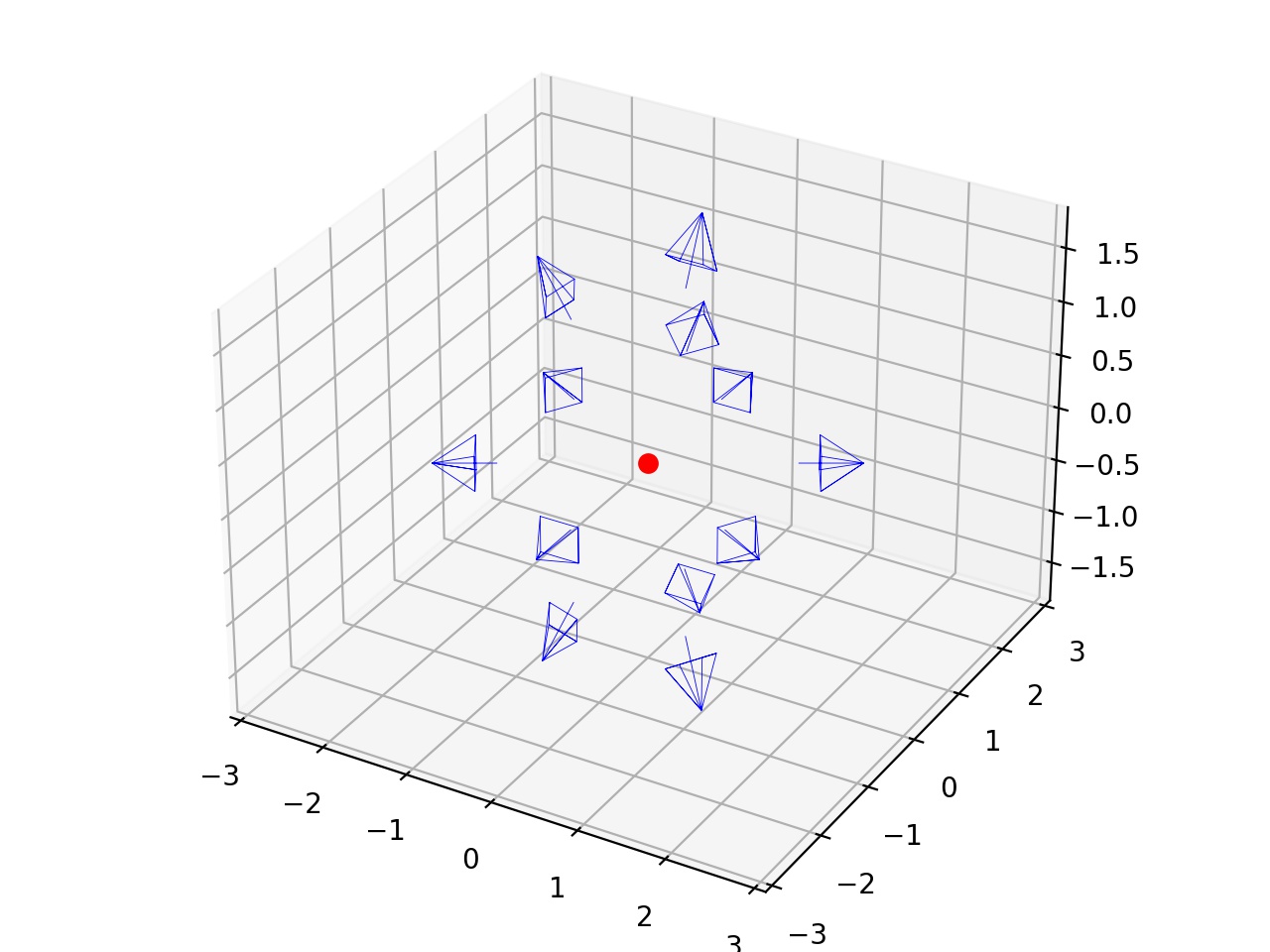} \\
\textbf{(c)} & 
\includegraphics[trim= 4cm 2.7cm 4cm 2.2cm , clip, width = 0.135\linewidth]{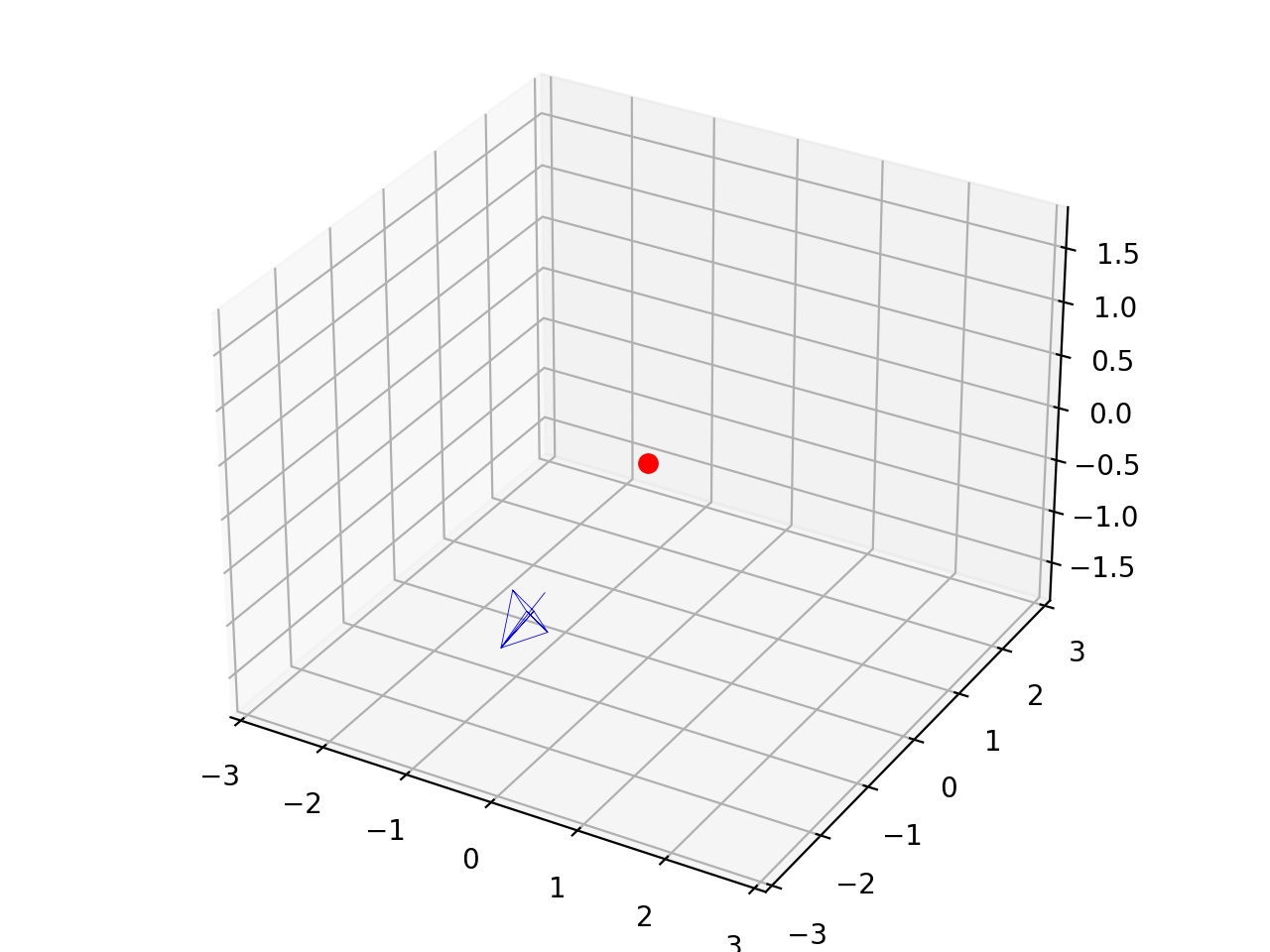} &
\includegraphics[trim= 4cm 2.7cm 4cm 2.2cm , clip, width = 0.135\linewidth]{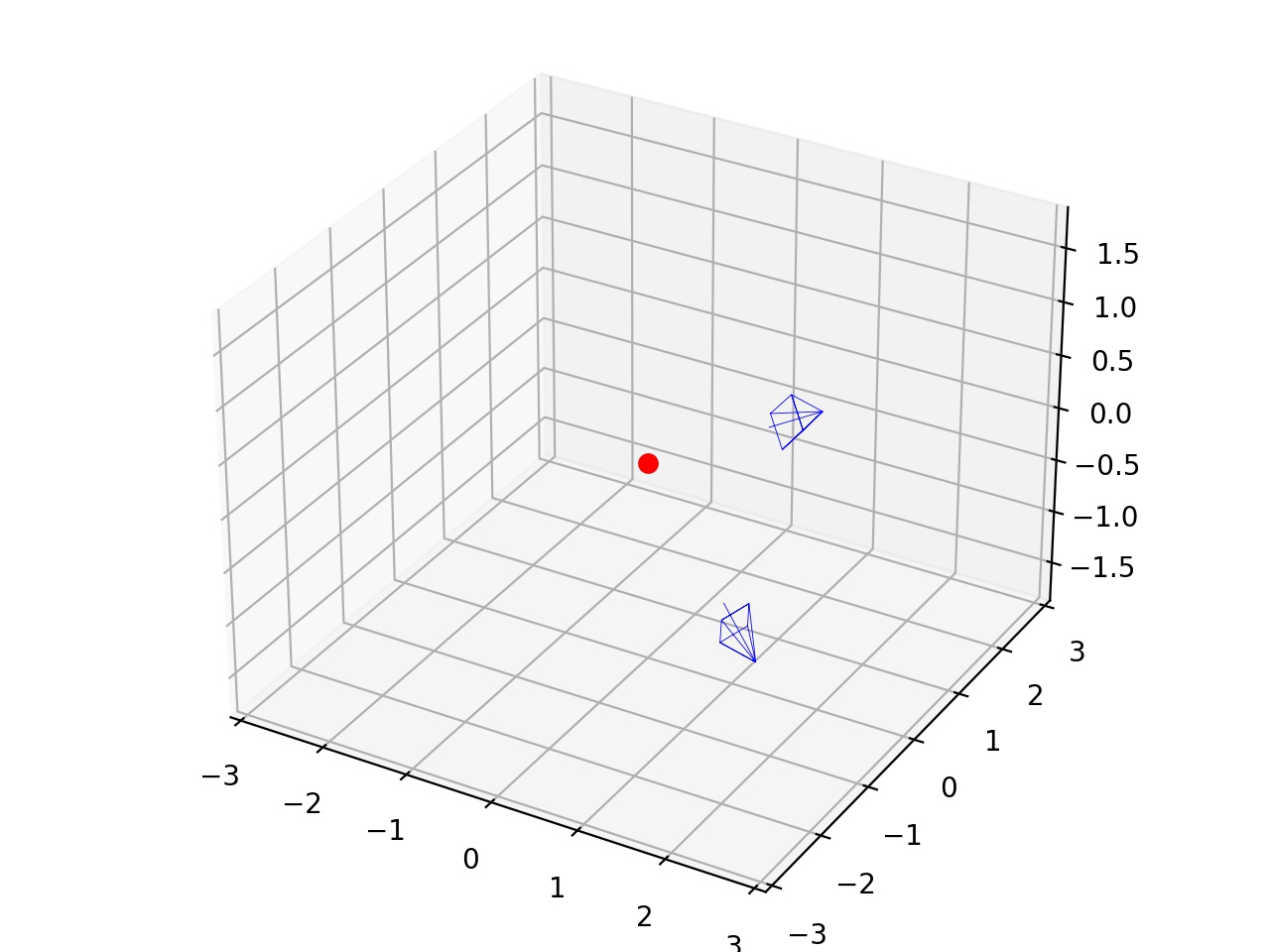} &
\includegraphics[trim= 4cm 2.7cm 4cm 2.2cm , clip, width = 0.135\linewidth]{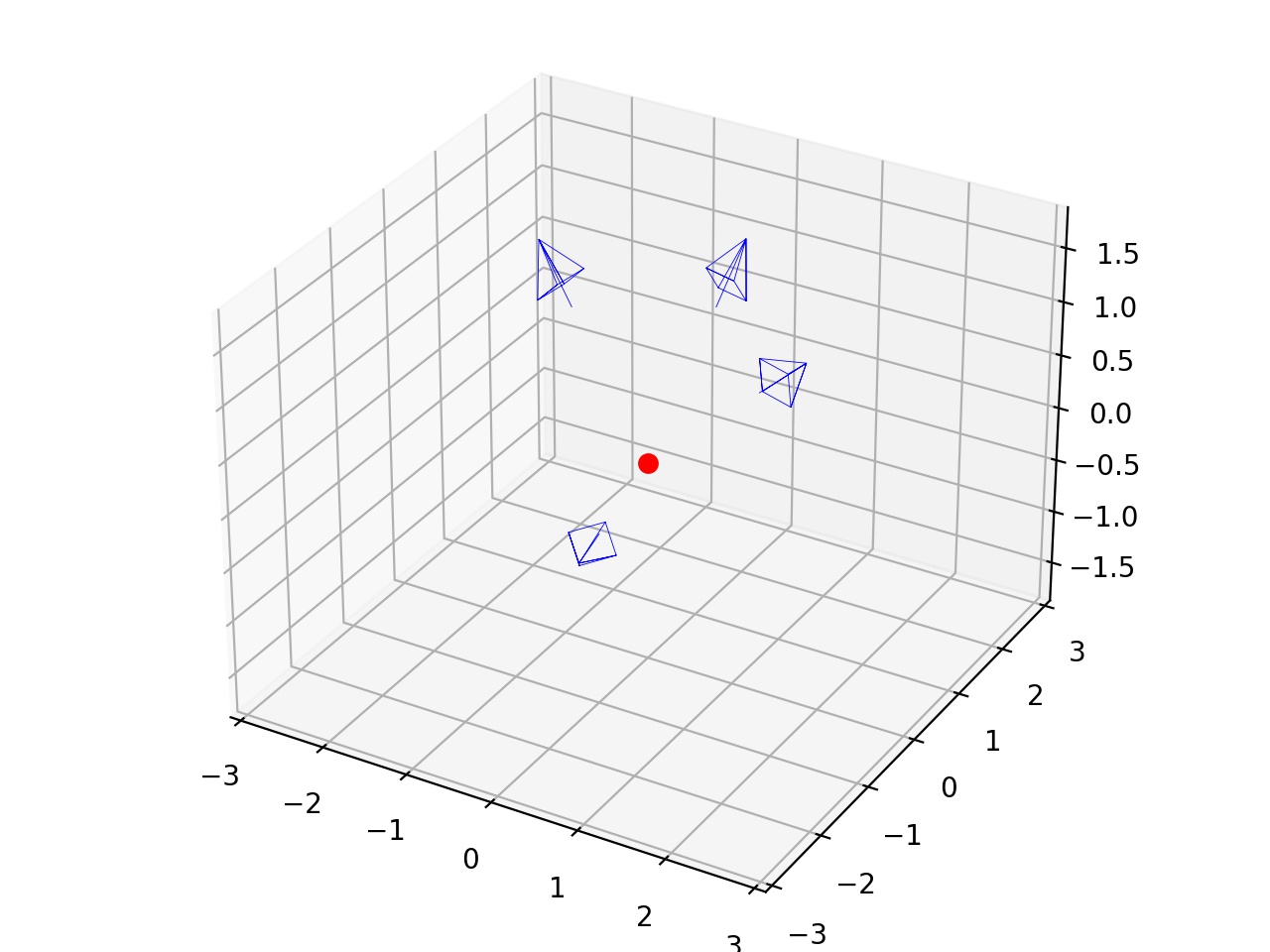} &
\includegraphics[trim= 4cm 2.7cm 4cm 2.2cm , clip, width = 0.135\linewidth]{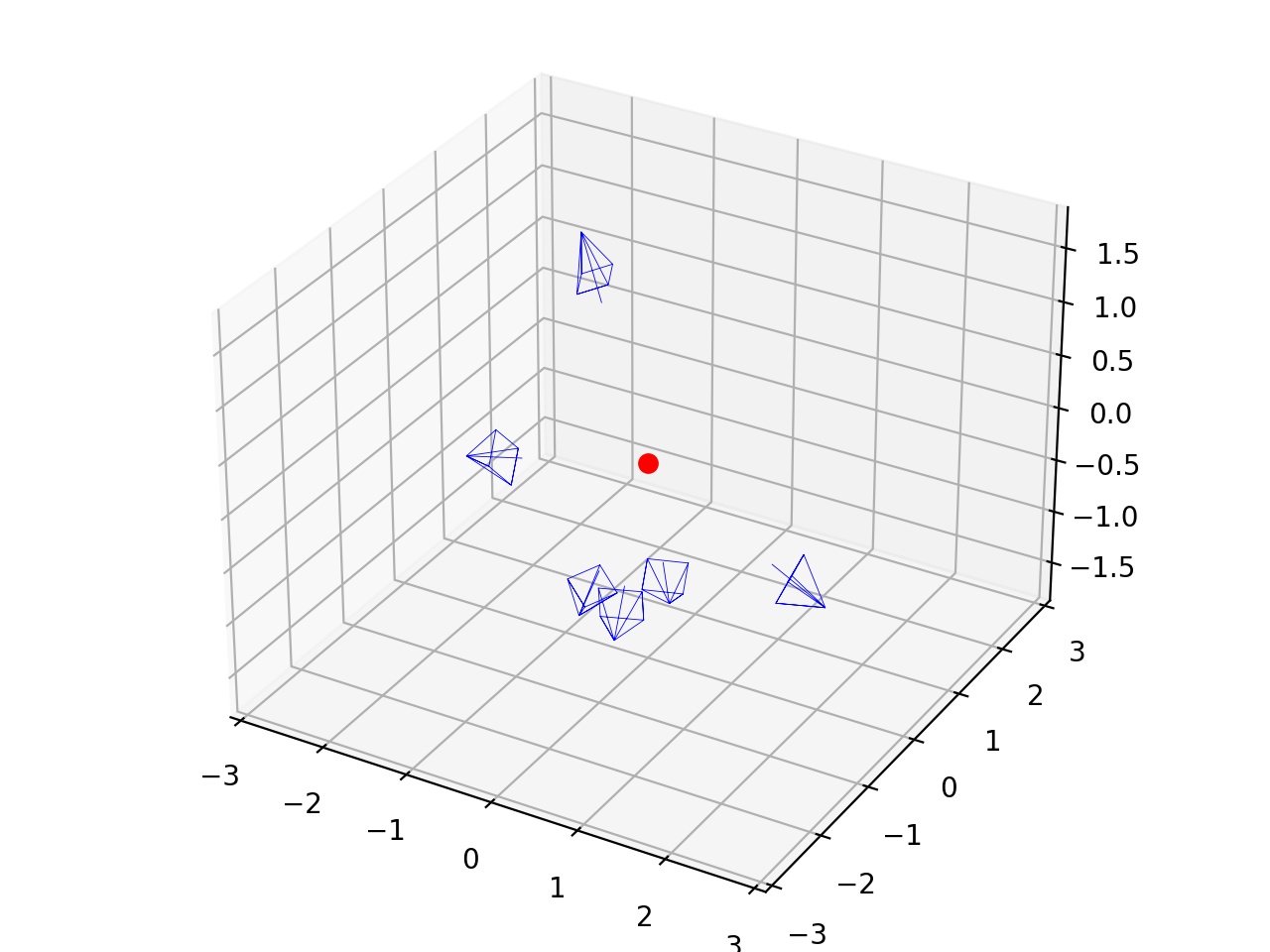} &
\includegraphics[trim= 4cm 2.7cm 4cm 2.2cm , clip, width = 0.135\linewidth]{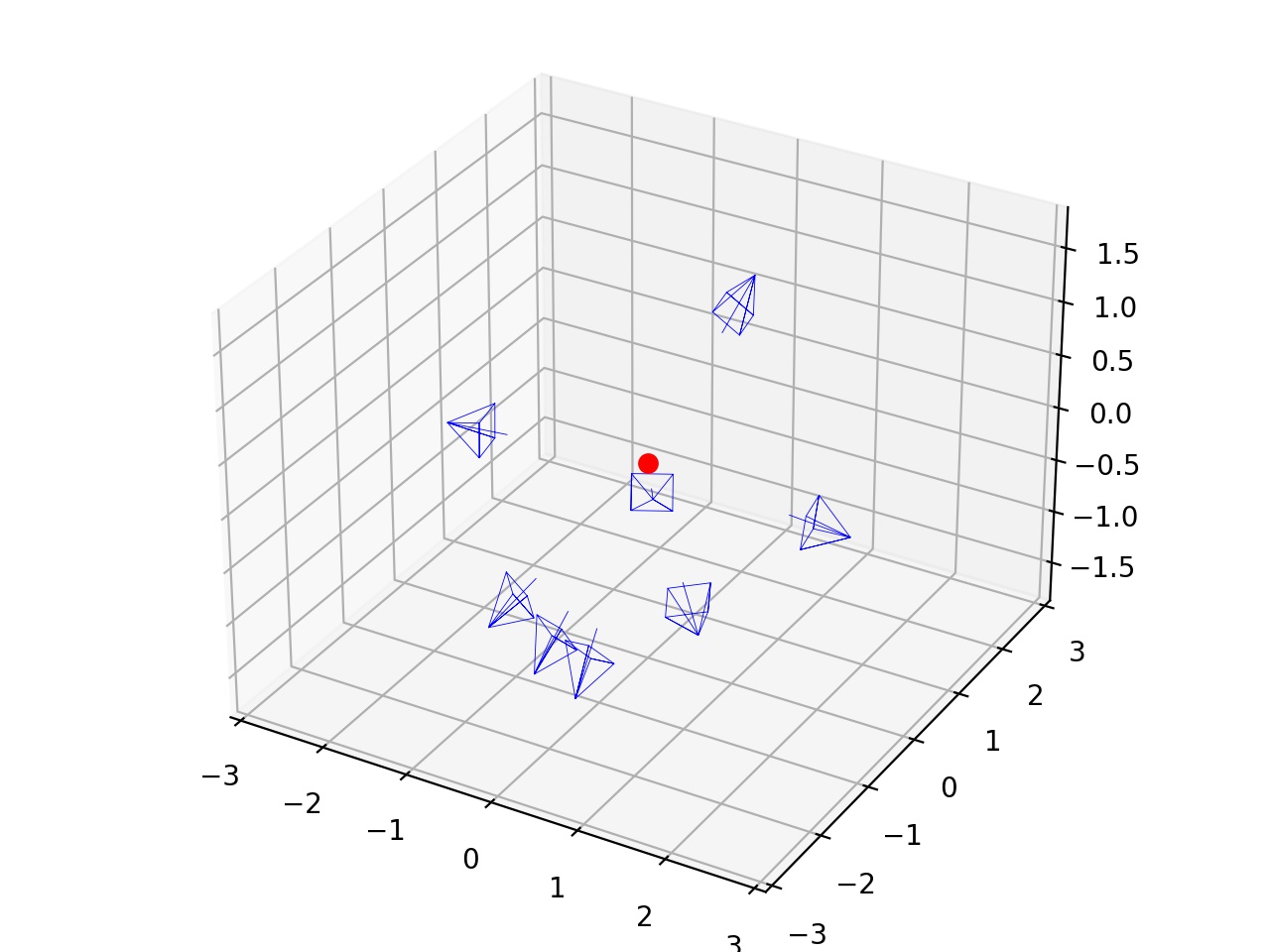} &
\includegraphics[trim= 4cm 2.7cm 4cm 2.2cm , clip, width = 0.135\linewidth]{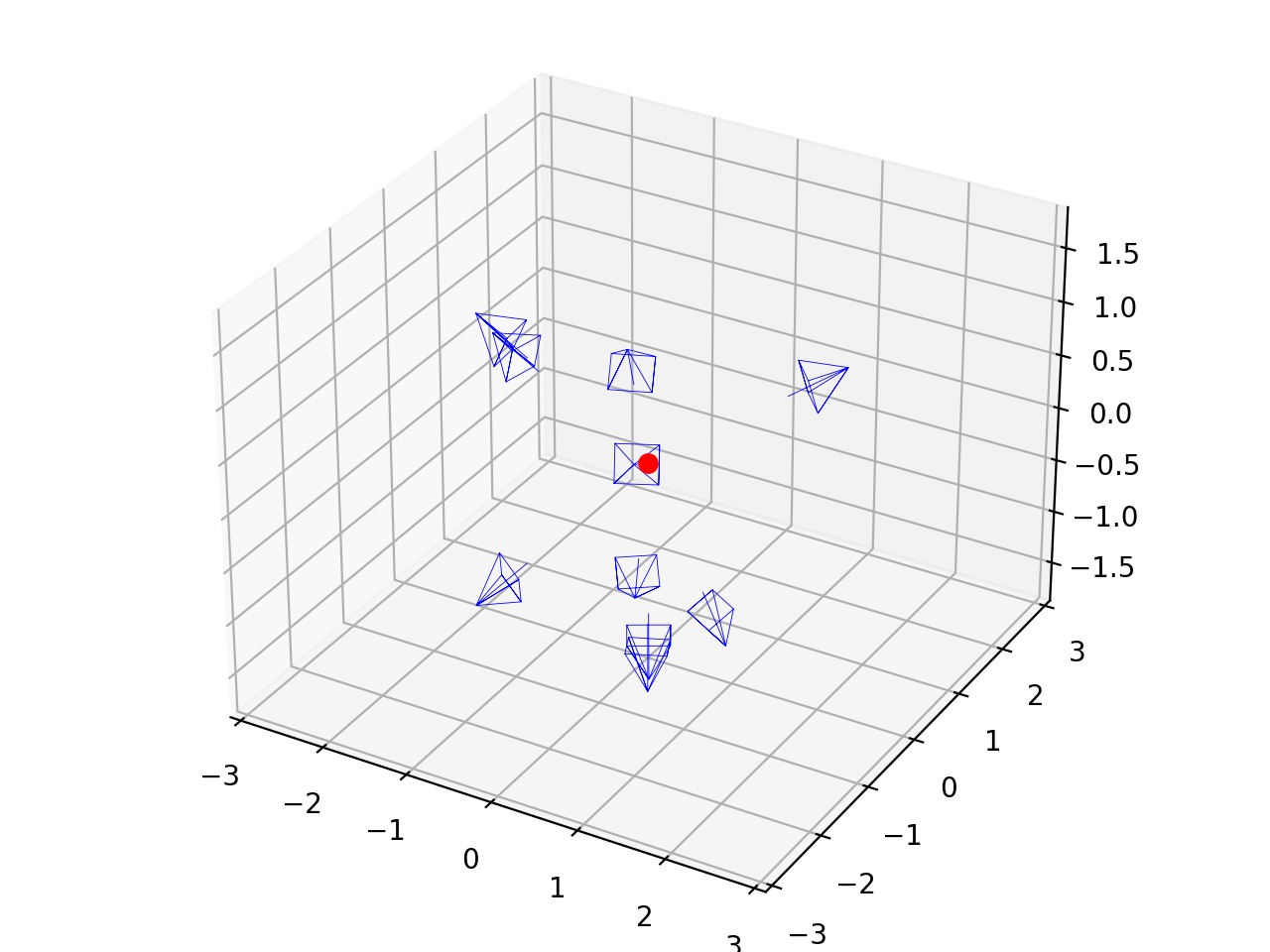} &
\includegraphics[trim= 4cm 2.7cm 4cm 2.2cm , clip, width = 0.135\linewidth]{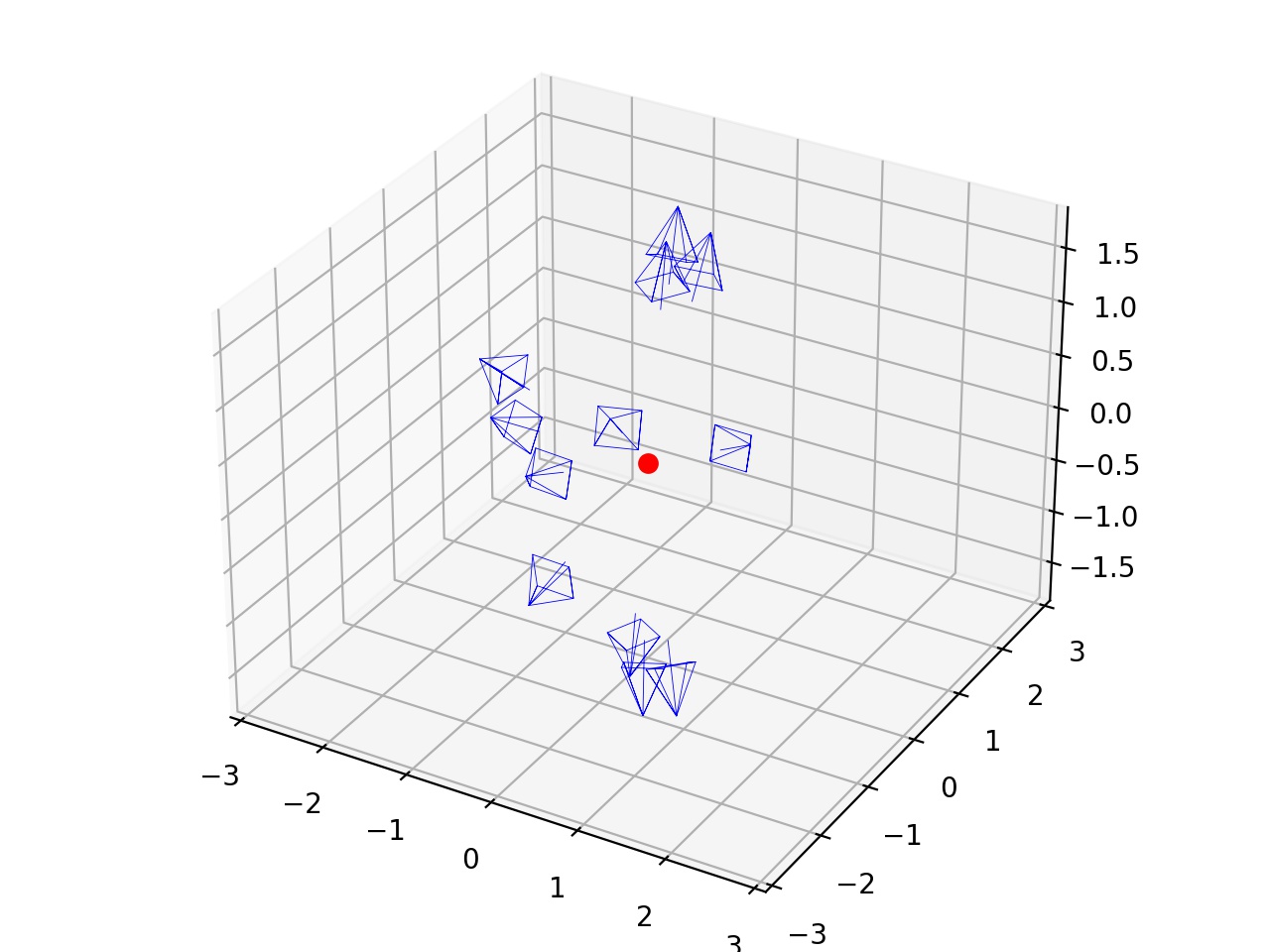} 

\end{tabular}
\vspace{2pt}
\caption{\small \textbf{Views Configurations.} We show some possible views configurations that can be used with a varying number of views. \textbf{(a):} circular, \textbf{(b):} spherical, \textbf{(c):} random 
}
    \label{fig:views-sup}
\end{figure*}

\mysection{View-Points}
In \figLabel{\ref{fig:views-sup}}, we show the basic views configurations for $M$ views previously used in the literature: circular, spherical, and random. MVTN's learned views are shown later in \specialcell{\ref{sec:ablation-supp}}
Since ViewGCN uses view sampling as a core operation, it requires the number of views to be at least 12, and hence, our MVTN with ViewGCN follows accordingly. 

\mysection{Training MVTN}
We use AdamW \cite{adamw} for our MVTN networks with a learning rate of 0.001. For other training details (\eg training epochs and optimization), we follow the previous works \cite{mvviewgcn,mvcnn} for a fair comparison. The training of MVTN with MVCNN is done in 100 epochs and a batch size of 20, while the MVTN with ViewGCN is performed in two stages as proposed in the official code of the paper \cite{mvviewgcn}. The first stage is 50 epochs of training the backbone CNN on the single view images, while the second stage is 35 epochs on the multi-view network on the $M$ views of the 3D object. We use learning rates of 0.0003 for MVCNN and 0.001 for ViewGCN, and a ResNet-18 \cite{resnet} as the backbone CNN for both baselines and our MVTN-based networks. A weight decay of 0.01 is applied for both the multi-view network and in the MVTN networks. Due to gradient instability from the renderer, we introduce gradient clipping in the MVTN to limit the $\ell_2$ norm of gradient updates to 30 for$\mathbf{G}$. The code is available at \url{https://github.com/ajhamdi/MVTN}.

\clearpage \clearpage
\section{Additional Results}

\subsection{Classification and Retrieval Benchmarks}
We provide in Tables \ref{tab:ModelNet40-cls-supp},\ref{tab:Scanobjectnn-supp}, and \ref{tab:retrieval-supp} comprehensive benchmarks of 3D classifications and 3D shape retrieval methods on ModelNet40 \cite{modelnet}, ScanObjectNN \cite{scanobjectnn}, and ShapeNet Core55 \cite{shapenet,shrek17}. These tables include methods that use points as representations as well as other modalities like multi-view and volumetric representations. Our reported results of four runs are presented in each table as ``max (avg $\pm$ std)''.  Note in Table \ref{tab:ModelNet40-cls-supp} how our MVTN improves the previous state-of-the-art in classification (ViewGCN \cite{mvviewgcn}) when tested on the same setup. Our implementations (highlighted using $*$) slightly differ from the reported results in their original paper. This can be attributed to the specific differentiable renderer of Pytorch3D \cite{pytorch3d} that we are using, which might not have the same quality of the non-differentiable OpenGL renderings \cite{opengl} used in their setups.

\subsection{Rotation Robustness}
\begin{figure}[ht]
    \centering
    \includegraphics[width=1.10\linewidth]{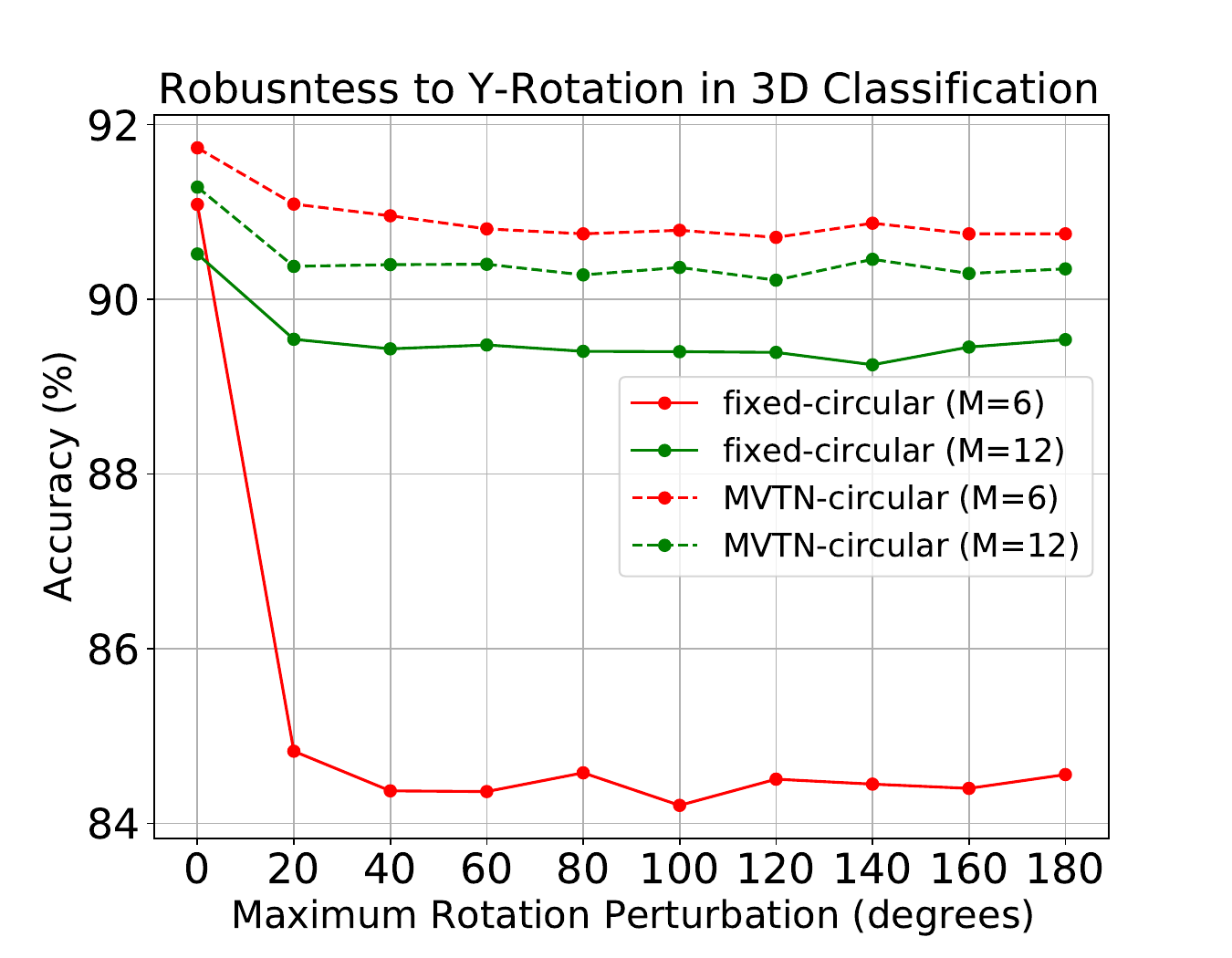}
    \caption{\textbf{Robustness on a Varying Y-Rotation}. We study the effect of varying the maximum rotation perturbation on the classification accuracies on ModelNet40. We compare the performance of circular MVCNN \cite{mvcnn} to our circular-MVTN when it equips MVCNN when the number of views is 6 and 12. Note how MVTN stabilizes the drop in performance for larger Y-rotation perturbations, and the improvement is more significant for the smaller number of views $M$.}
    \label{fig:y-robustness-ablate-sup}
\end{figure}

A common practice in the literature in 3D shape classification is to test the robustness of models trained on the aligned dataset by injecting perturbations during test time \cite{rspointcloud}. We follow the same setup as \cite{rspointcloud} by introducing random rotations during test time around the Y-axis (gravity-axis).
We also investigate the effect of varying rotation perturbations on the accuracy of circular MVCNN when $M=6$ and $M=12$. We note from \figLabel{\ref{fig:y-robustness-ablate-sup}} that using less views leads to higher sensitivity to rotations in general. Furthermore, we note that our MVTN helps in stabilizing the performance on increasing thresholds of rotation perturbations.

\subsection{Occlusion Robustness} \label{sec:occlusion-supp}
\begin{figure}[t]
    \centering
    \includegraphics[trim= {1cm 0 0 0},clip , width=1.10\linewidth]{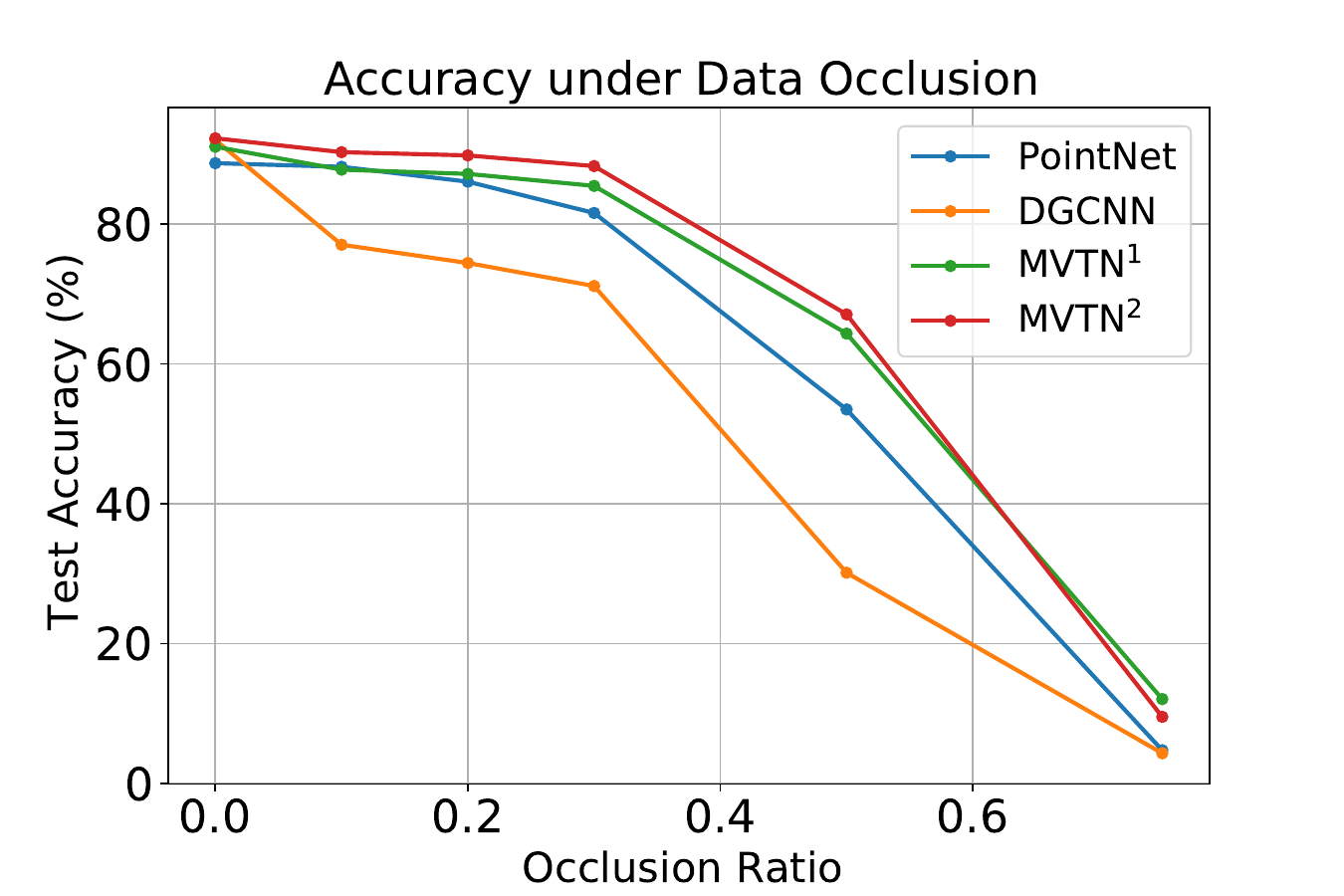}
    \caption{\textbf{Occlusion Robustness of 3D Methods.} We plot test accuracy \vs the Occlusion Ratio of the data to simulate the occlusion robustness of different 3D methods: PointNet \cite{pointnet}, DGCNN \cite{dgcn}, and MVTN. Our MVTN achieves close to 13\% better than PointNet when half of the object is occluded. MVTN$^1$ refers to MVTN with MVCNN as the multi-view network while MVTN$^2$ refers to MVTN with View-GCN as the multi-view network.}
    \label{fig:occlusion}
\end{figure}

\begin{figure*} [] 
\tabcolsep=0.03cm
\resizebox{0.99\linewidth}{!}{
\begin{tabular}{c|ccccc}

 & \multicolumn{5}{c}{Occlusion Ratio} \\ 
 \textbf{Direction} & 0.1 & 0.2 & 0.3 & 0.5 & 0.75 \\ \midrule

\textbf{+X} &
\includegraphics[trim= 0cm 2cm 0cm 0cm , clip, width = 0.19\linewidth]{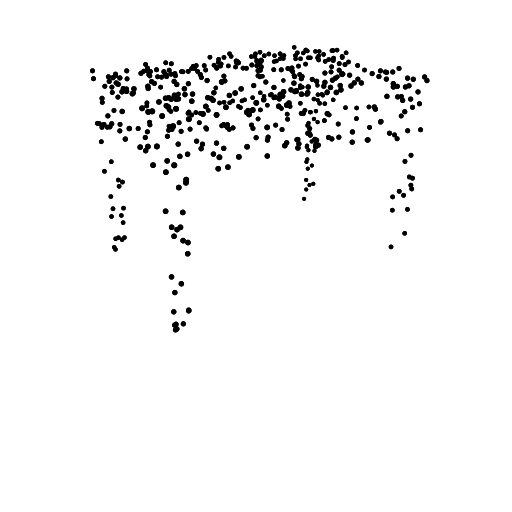} &
\includegraphics[trim= 0cm 2cm 0cm 0cm , clip, width = 0.19\linewidth]{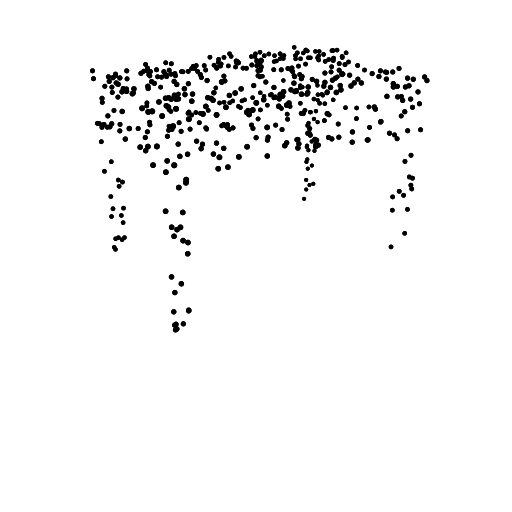} &
\includegraphics[trim= 0cm 2cm 0cm 0cm , clip, width = 0.19\linewidth]{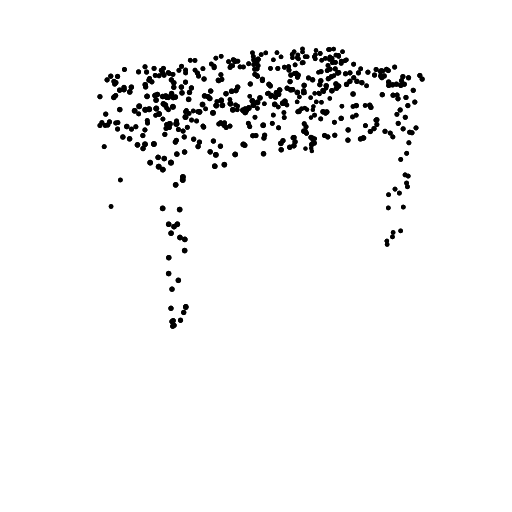} &
\includegraphics[trim= 0cm 2cm 0cm 0cm , clip, width = 0.19\linewidth]{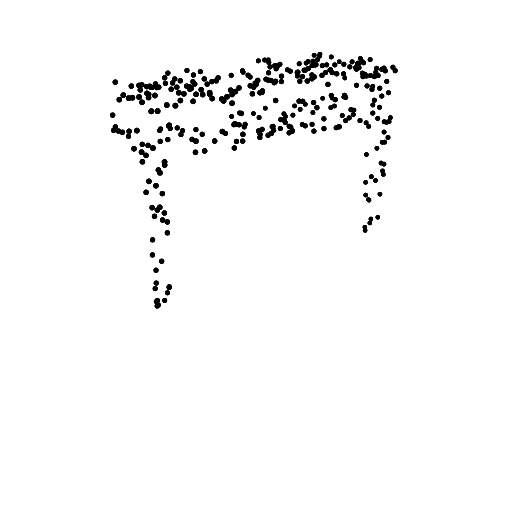} &
\includegraphics[trim= 0cm 2cm 0cm 0cm , clip, width = 0.19\linewidth]{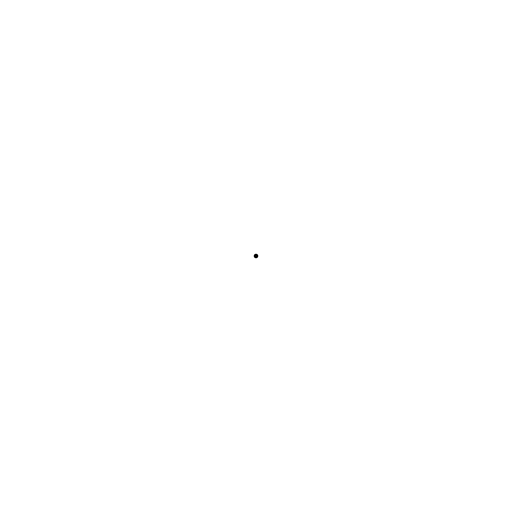} \\ \hline

\textbf{-X} &
\includegraphics[trim= 0cm 2cm 0cm 0cm , clip, width = 0.19\linewidth]{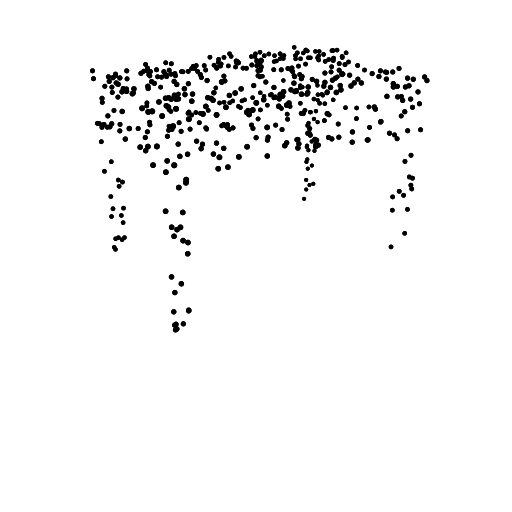} &
\includegraphics[trim= 0cm 2cm 0cm 0cm , clip, width = 0.19\linewidth]{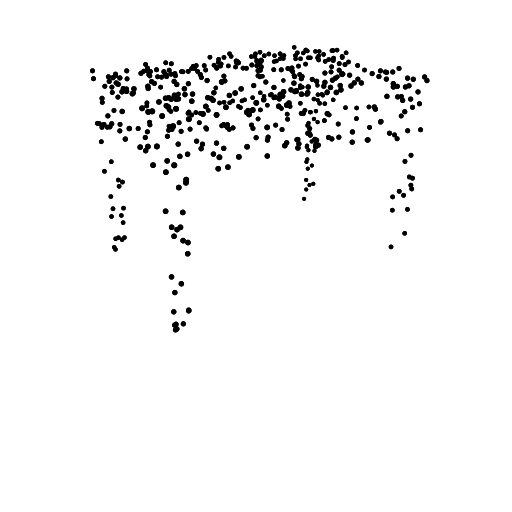} &
\includegraphics[trim= 0cm 2cm 0cm 0cm , clip, width = 0.19\linewidth]{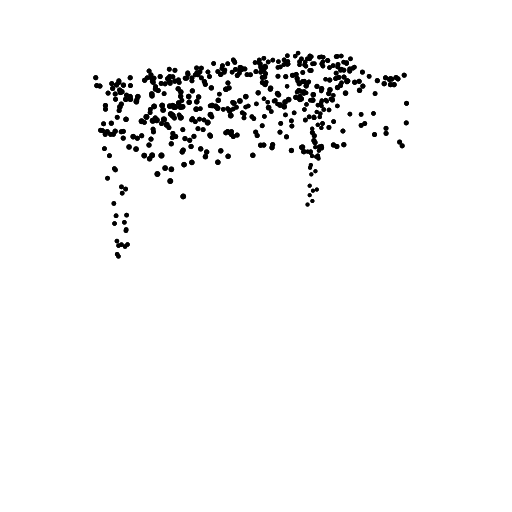} &
\includegraphics[trim= 0cm 2cm 0cm 0cm , clip, width = 0.19\linewidth]{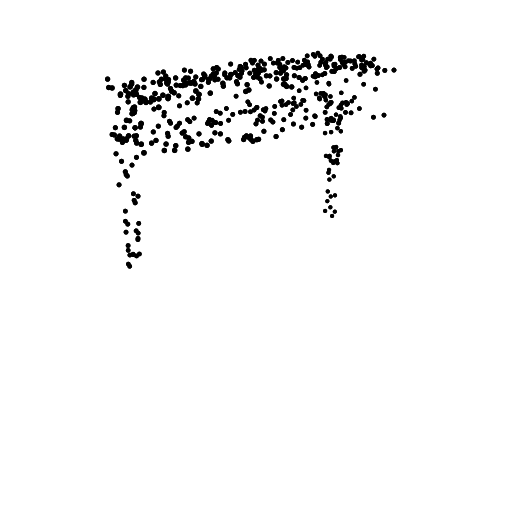} &
\includegraphics[trim= 0cm 2cm 0cm 0cm , clip, width = 0.19\linewidth]{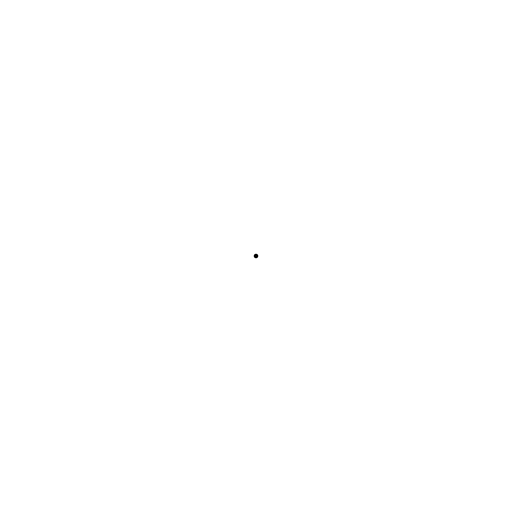} \\ \midrule

\textbf{+Z} &
\includegraphics[trim= 0cm 2cm 0cm 0cm , clip, width = 0.19\linewidth]{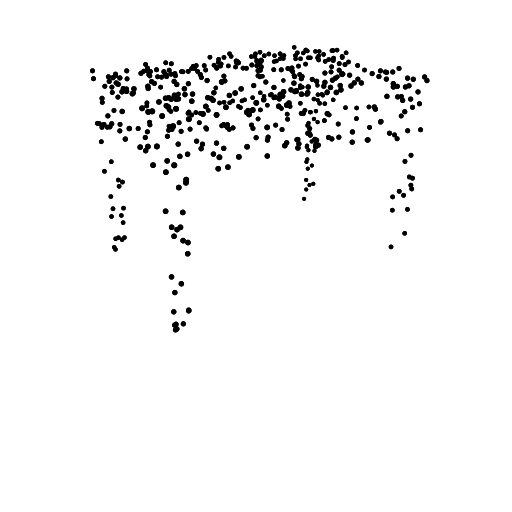} &
\includegraphics[trim= 0cm 2cm 0cm 0cm , clip, width = 0.19\linewidth]{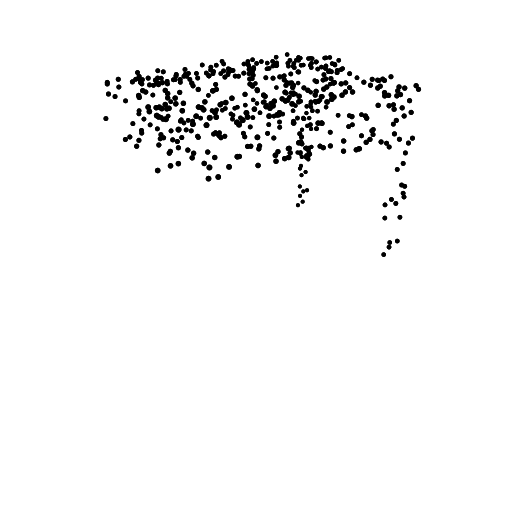} &
\includegraphics[trim= 0cm 2cm 0cm 0cm , clip, width = 0.19\linewidth]{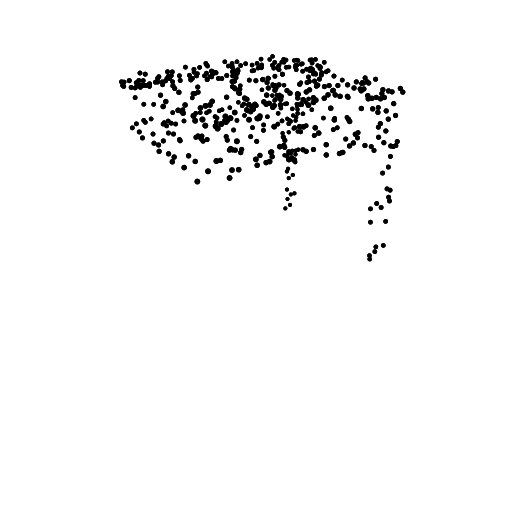} &
\includegraphics[trim= 0cm 2cm 0cm 0cm , clip, width = 0.19\linewidth]{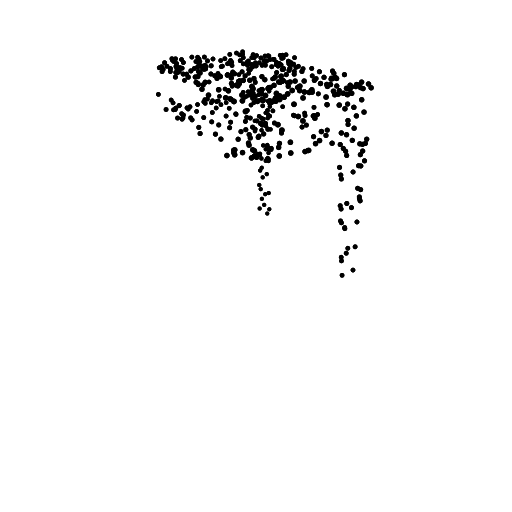} &
\includegraphics[trim= 0cm 2cm 0cm 0cm , clip, width = 0.19\linewidth]{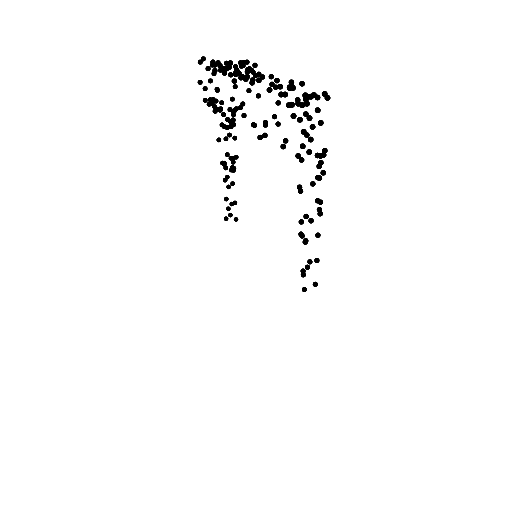} \\ \hline

\textbf{-Z} &
\includegraphics[trim= 0cm 2cm 0cm 0cm , clip, width = 0.19\linewidth]{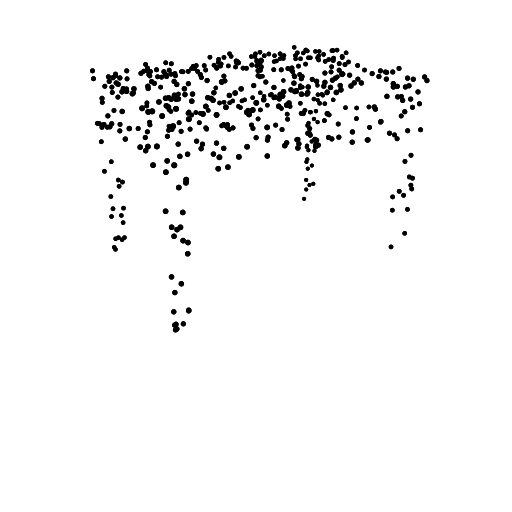} &
\includegraphics[trim= 0cm 2cm 0cm 0cm , clip, width = 0.19\linewidth]{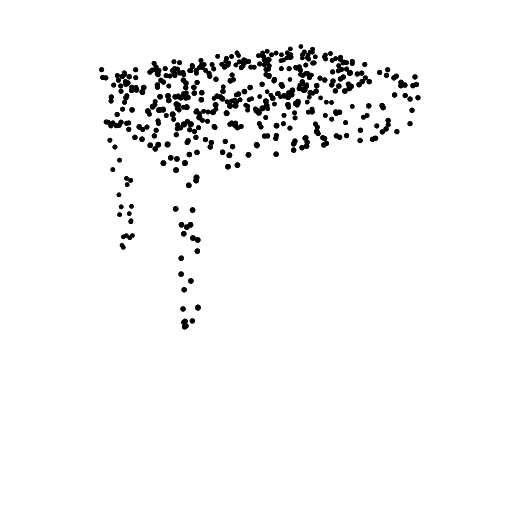} &
\includegraphics[trim= 0cm 2cm 0cm 0cm , clip, width = 0.19\linewidth]{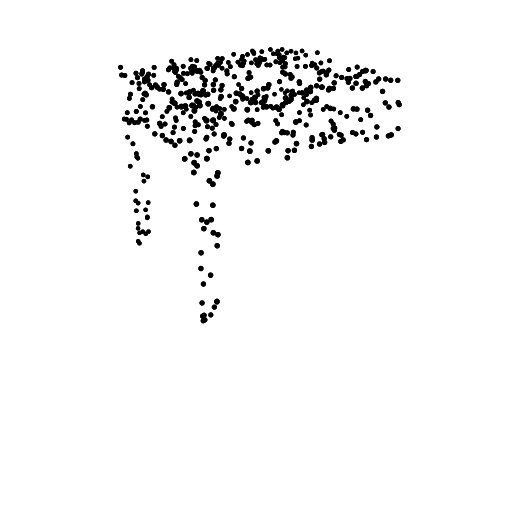} &
\includegraphics[trim= 0cm 2cm 0cm 0cm , clip, width = 0.19\linewidth]{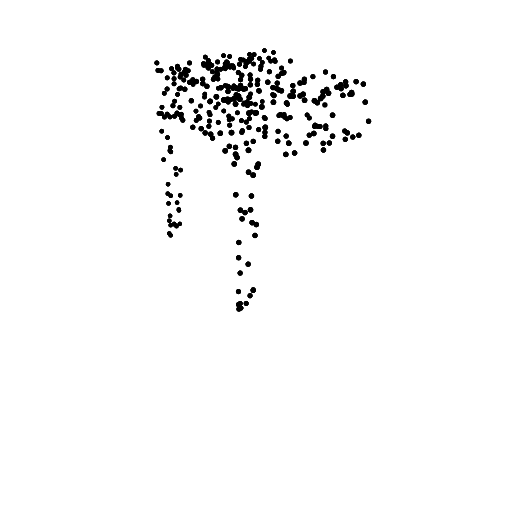} &
\includegraphics[trim= 0cm 2cm 0cm 0cm , clip, width = 0.19\linewidth]{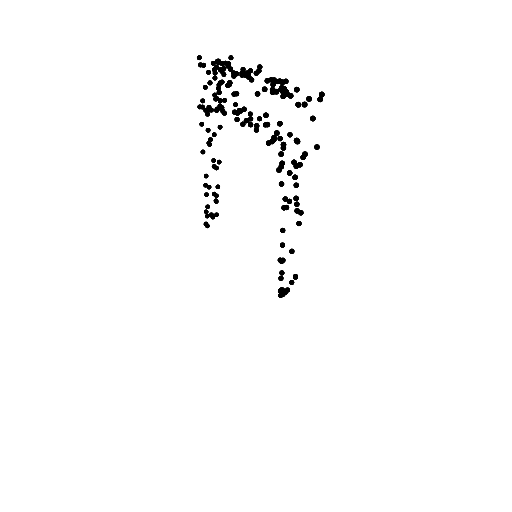} \\ \midrule

\textbf{+Y} &
\includegraphics[trim= 0cm 2cm 0cm 0cm , clip, width = 0.19\linewidth]{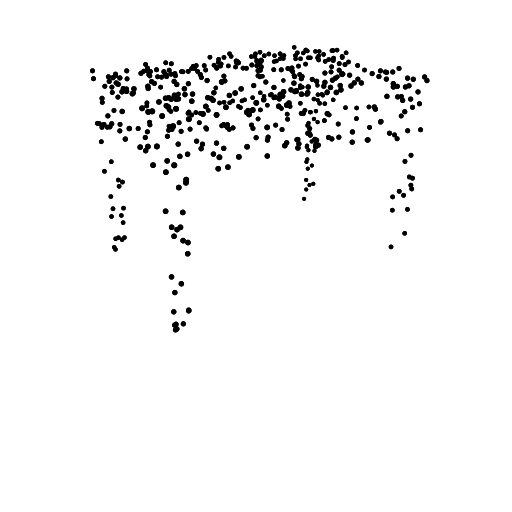} &
\includegraphics[trim= 0cm 2cm 0cm 0cm , clip, width = 0.19\linewidth]{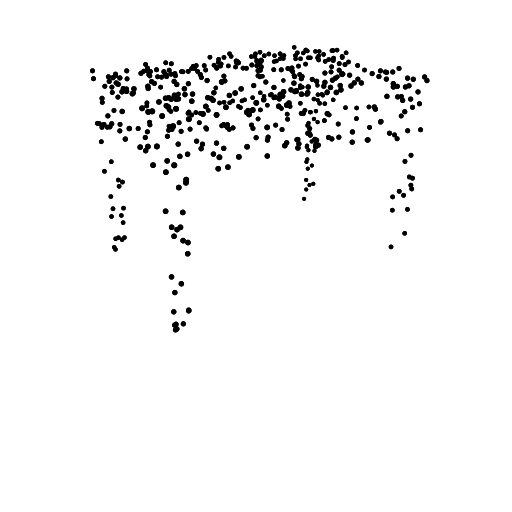} &
\includegraphics[trim= 0cm 2cm 0cm 0cm , clip, width = 0.19\linewidth]{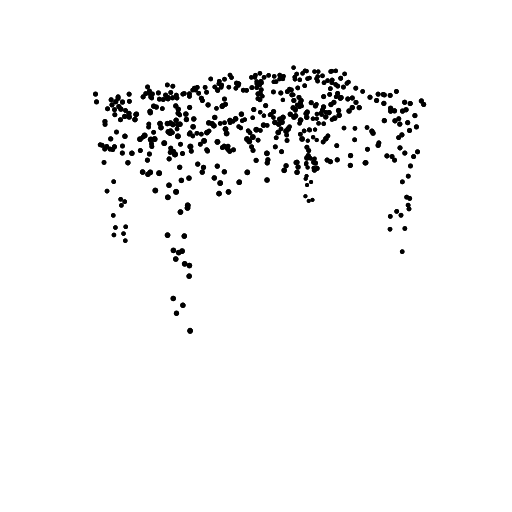} &
\includegraphics[trim= 0cm 2cm 0cm 0cm , clip, width = 0.19\linewidth]{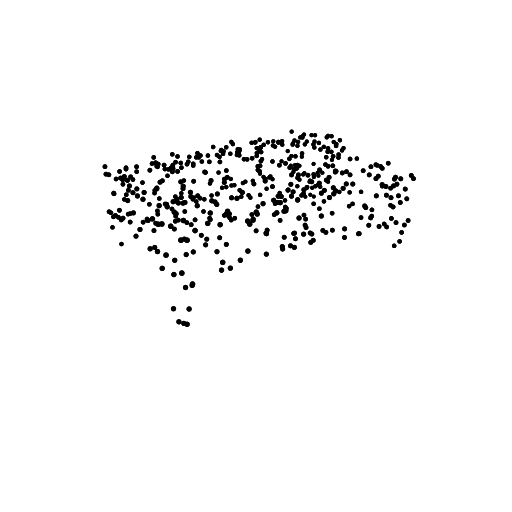} &
\includegraphics[trim= 0cm 2cm 0cm 0cm , clip, width = 0.19\linewidth]{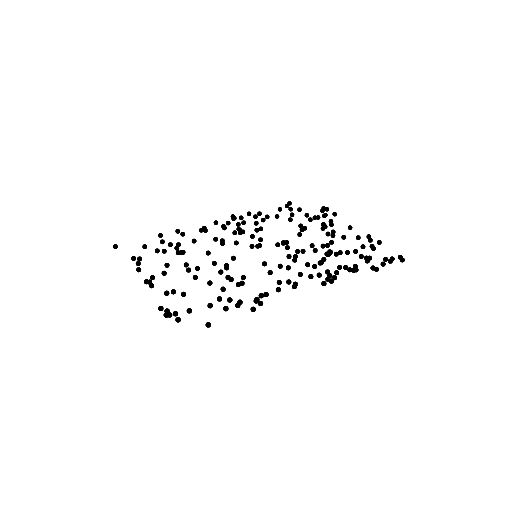} \\ \hline

\textbf{-Y} &
\includegraphics[trim= 0cm 2cm 0cm 0cm , clip, width = 0.19\linewidth]{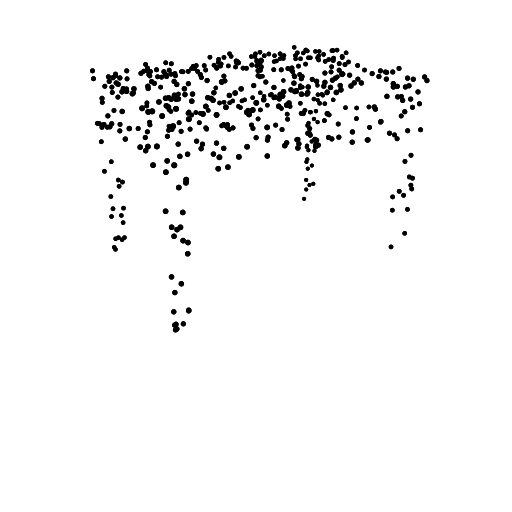} &
\includegraphics[trim= 0cm 2cm 0cm 0cm , clip, width = 0.19\linewidth]{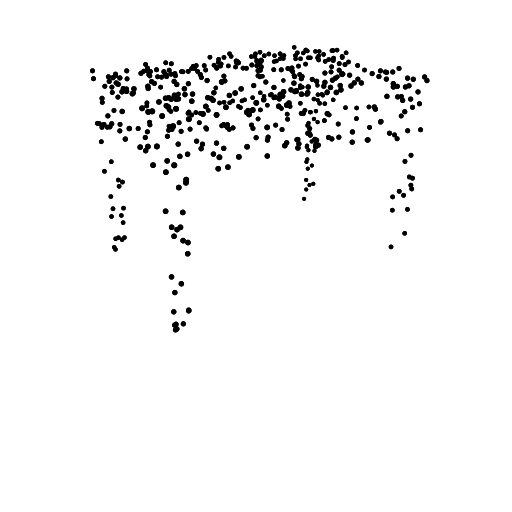} &
\includegraphics[trim= 0cm 2cm 0cm 0cm , clip, width = 0.19\linewidth]{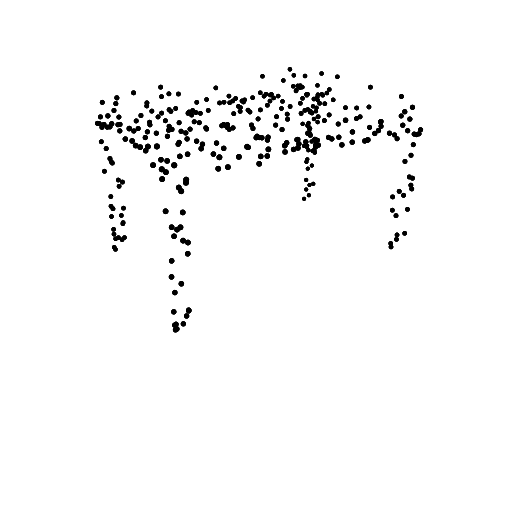} &
\includegraphics[trim= 0cm 2cm 0cm 0cm , clip, width = 0.19\linewidth]{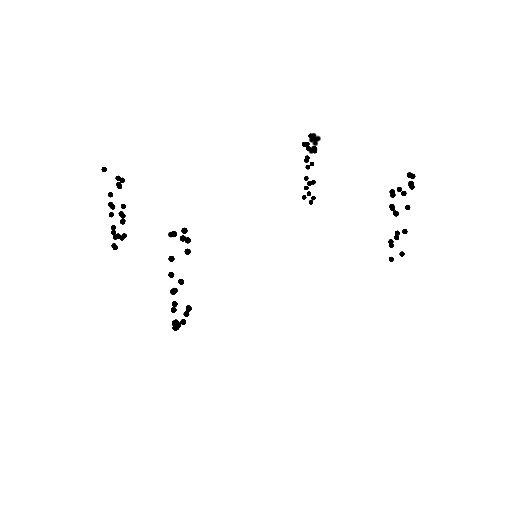} &
\includegraphics[trim= 0cm 2cm 0cm 0cm , clip, width = 0.19\linewidth]{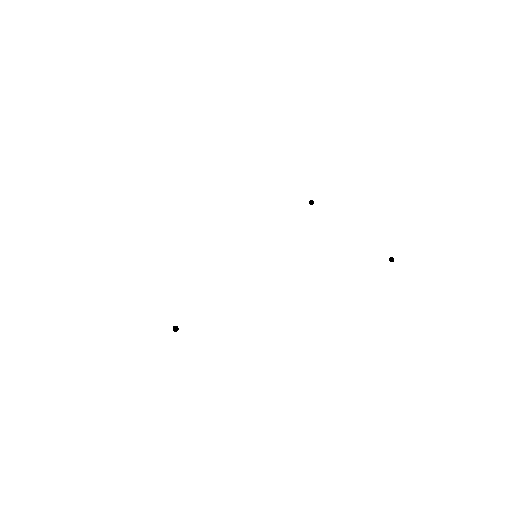} \\ \hline

 \bottomrule
\end{tabular}
}
\vspace{2pt}
\caption{\small \textbf{Occlusion of 3D Objects}: We simulate realistic occlusion scenarios in 3D point clouds by cropping a percentage of the object along canonical directions. Here, we show an object occluded with different ratios and from different directions. 
}
    \label{fig:occlusion-supp}
\end{figure*}

To quantify the occlusion effect due to the viewing angle of the 3D sensor in our setup of 3D classification, we simulate realistic occlusion by cropping the object from canonical directions. We train PointNet \cite{pointnet}, DGCNN \cite{dgcn}, and MVTN on the ModelNet40 point cloud dataset. Then, at test time, we crop a portion of the object (from 0\% occlusion ratio to 75\%) along the $\pm$X, $\pm$Y, and $\pm$Z directions independently. \figLabel{\ref{fig:occlusion-supp}} shows examples of this occlusion effect with different occlusion ratios. We report the average test accuracy (on all the test set) of the six cropping directions for the baselines and MVTN in \figLabel{\ref{fig:occlusion}}. Note how MVTN achieves high test accuracy even when large portions of the object are cropped. Interestingly, MVTN outperforms PointNet \cite{pointnet} by 13\% in test accuracy when half of the object is occluded. This result is significant, given that PointNet is well-known for its robustness \cite{pointnet,advpc}. 

\begin{table*}[t]
\tabcolsep=0.25cm
    \centering
\resizebox{0.7\linewidth}{!}{\begin{tabular}{rccc}
\toprule
 &   & \multicolumn{2}{c}{Classification Accuracy} \\
\multicolumn{1}{c}{Method}       & Data Type & (\textbf{Per-Class})   & (\textbf{Overall}) \\ \midrule
SPH  \cite{sph}                   &  Voxels                 & 68.2 & - \\
LFD     \cite{lfd}                &  Voxels                 & 75.5 & - \\ 
3D ShapeNets \cite{modelnet} &  Voxels                 & 77.3 & - \\
VoxNet \cite{voxnet}     & Voxels                 & 83.0 & 85.9      \\
VRN  \cite{vrn}                      & Voxels                 & - & 91.3      \\
MVCNN-MS \cite{multimvcnn} &        Voxels                 & - & 91.4      \\ 
FusionNet \cite{fusionnet}                & Voxels+MV  & - & 90.8      \\ 
PointNet \cite{pointnet} &  Points                  &       86.2 & 89.2      \\
PointNet++ \cite{pointnet++}   & Points                & - & 91.9      \\
KD-Network \cite{kdnet}&    Points                &     88.5 &     91.8   \\
PointCNN \cite{pc_li2018pointcnn}  & Points            &   88.1   & 91.8      \\
DGCNN \cite{dgcn}             & Points                &  90.2     & 92.2      \\
KPConv\cite{kpconv}  &  Points & -  & 92.9 \\ 
PVNet\cite{pvnet}  &  Points & -  & 93.2 \\ 
PTransformer\cite{pointtransformer}  &  Points & \textbf{90.6} & \textbf{93.7}  \\ \midrule
MVCNN  \cite{mvcnn}         & 12 Views                   & 90.1  & 90.1 \\
GVCNN \cite{mvgvcnn}         & 12 Views                   & 90.7 & 93.1 \\
ViewGCN \cite{mvviewgcn}  & 20 Views   & \textbf{96.5} & \textbf{97.6} \\ 

\midrule
ViewGCN \cite{mvviewgcn}$^*$& 12 views &    90.7 (90.5 $\pm$ 0.2)   &93.0 (92.8 $\pm$ 0.1) \\
ViewGCN \cite{mvviewgcn}$^*$& 20 views &    91.3 (91.0 $\pm$ 0.2)   &93.3 (93.1 $\pm$ 0.2) \\
MVTN (ours)$^*$  & 12 Views       & 92.0 (91.2 $\pm$ 0.6) & \textbf{93.8} (93.4 $\pm$ 0.3) \\
MVTN (ours)$^*$  & 20 Views       & \textbf{92.2} (91.8 $\pm$ 0.3) & 93.5 (93.1 $\pm$ 0.5) \\
\bottomrule
\end{tabular}
}
\vspace{2pt}
    \caption{\textbf{3D Shape Classification on ModelNet40}. We compare MVTN against other methods in 3D classification on ModelNet40 \cite{modelnet}. $^*$ indicates results from our rendering setup (differentiable pipeline), while other multi-view results are reported from pre-rendered views. \textbf{Bold} denotes the best result in its setup. In brackets, we report the average and standard deviation of four runs}
    \label{tab:ModelNet40-cls-supp}
\end{table*}
\begin{table*}[t]
\tabcolsep=0.2cm
    \centering
\resizebox{0.85\linewidth}{!}{\begin{tabular}{rccc}
\toprule
 &  \multicolumn{3}{c}{Classification Overall Accuracy } \\
\multicolumn{1}{c}{Method}& \textbf{Object with Background}  & \textbf{Object Only} & \textbf{PB\_T50\_RS (Hardest)}  \\ \midrule
3DMFV \cite{3Dmfv} &  68.2                  &  73.8  &  63.0  \\
PointNet \cite{pointnet}   & 73.3                &   79.2  & 68.0 \\
SpiderCNN \cite{pc_xu2018spidercnn}&    77.1                &    79.5   &  73.7    \\
PointNet ++ \cite{pointnet++}            & 82.3 & 84.3  &   77.9   \\
PointCNN \cite{pc_li2018pointcnn}  & 86.1 & 85.5  & 78.5 \\
DGCNN \cite{dgcn}  & 82.8 & 86.2  & 78.1 \\ 
SimpleView \cite{simpleview}& - & - & 79.5 \\
BGA-DGCNN \cite{scanobjectnn}   & - & - & 79.7 \\
BGA-PN++ \cite{scanobjectnn}   & - & - & 80.2 \\
\midrule
ViewGCN $^*$  & 91.9 (91.12 $\pm$ 0.5)  & 90.4 (89.7 $\pm$ 0.5) & 80.5 (80.2 $\pm$ 0.4) \\
MVTN (ours)  & \textbf{92.6} (92.5 $\pm$ 0.2)  & \textbf{92.3} (91.7 $\pm$ 0.7) & \textbf{82.8} (81.8 $\pm$ 0.7) \\
\bottomrule
\end{tabular}
}
\vspace{2pt}
    \caption{\textbf{3D Point Cloud Classification on ScanObjectNN}. We compare the performance of MVTN in 3D point cloud classification on three different variants of ScanObjectNN \cite{scanobjectnn}. The variants include object with background, object only, and the hardest variant. $^*$ indicates results from our rendering setup (differentiable pipeline), and we report the average and standard deviation of four runs in brackets.}
    \label{tab:Scanobjectnn-supp}
\end{table*}

\begin{table*}[t]
\tabcolsep=0.25cm
    \centering
\resizebox{0.7\linewidth}{!}{\begin{tabular}{rccc}
\toprule
 &   & \multicolumn{2}{c}{ Shape Retrieval (mAP)} \\
\multicolumn{1}{c}{Method} & Data Type  & \textbf{ModelNet40}  & \textbf{ShapeNet Core} \\ \midrule
ZDFR \cite{zfdrretr} &  Voxels                  &      - & 19.9      \\
DLAN \cite{dlanretr}   & Voxels                &   - & 66.3      \\
SPH  \cite{sph}                   &  Voxels                 & 33.3 & - \\
LFD     \cite{lfd}                &  Voxels                 & 40.9 & - \\ 
3D ShapeNets \cite{modelnet} &  Voxels                 & 49.2 & - \\
PVNet\cite{pvnet}  &  Points & 89.5 & - \\ 
MVCNN  \cite{mvcnn}         & 12 Views                   & 80.2 & 73.5 \\
GIFT \cite{giftretr}&    20 Views                &     - & 64.0      \\
MVFusionNet \cite{mvfusionnet}            & 12 Views                &      - & 62.2      \\
ReVGG \cite{shrek17}  & 20 Views            &     - & 74.9      \\
RotNet \cite{mvrotationnet}  & 20 Views   & - & 77.2 \\ 
ViewGCN \cite{mvviewgcn}  & 20 Views   & - &  78.4 \\ 
MLVCNN \cite{mlvcnn} &   24 Views               & 92.2      & -\\
\midrule
MVTN (ours)  & 12 Views         & \textbf{92.9} (92.4 $\pm$ 0.6) &  \textbf{82.9} (82.4 $\pm$ 0.6) \\
\bottomrule
\end{tabular}
}
\vspace{2pt}
    \caption{\textbf{3D Shape Retrieval}. We benchmark the shape retrieval capability of MVTN on ModelNet40 \cite{modelnet} and ShapeNet Core55 \cite{shapenet,shrek17}. MVTN achieves the best retrieval performance among recent state-of-the-art methods on both datasets with only 12 views. In brackets, we report the average and standard deviation of four runs.}
     \label{tab:retrieval-supp}
\end{table*}

\clearpage \clearpage
\section{Analysis and Insights}

\subsection{Ablation Study} \label{sec:ablation-supp}
This section introduces a comprehensive ablation study on the different components of MVTN, and their effect on test accuracy on the standard ModelNet40 \cite{modelnet}.

\begin{figure}[t]
    \centering
    \includegraphics[,trim=0.5cm 0 1.5cm 0, clip,width=0.95\linewidth]{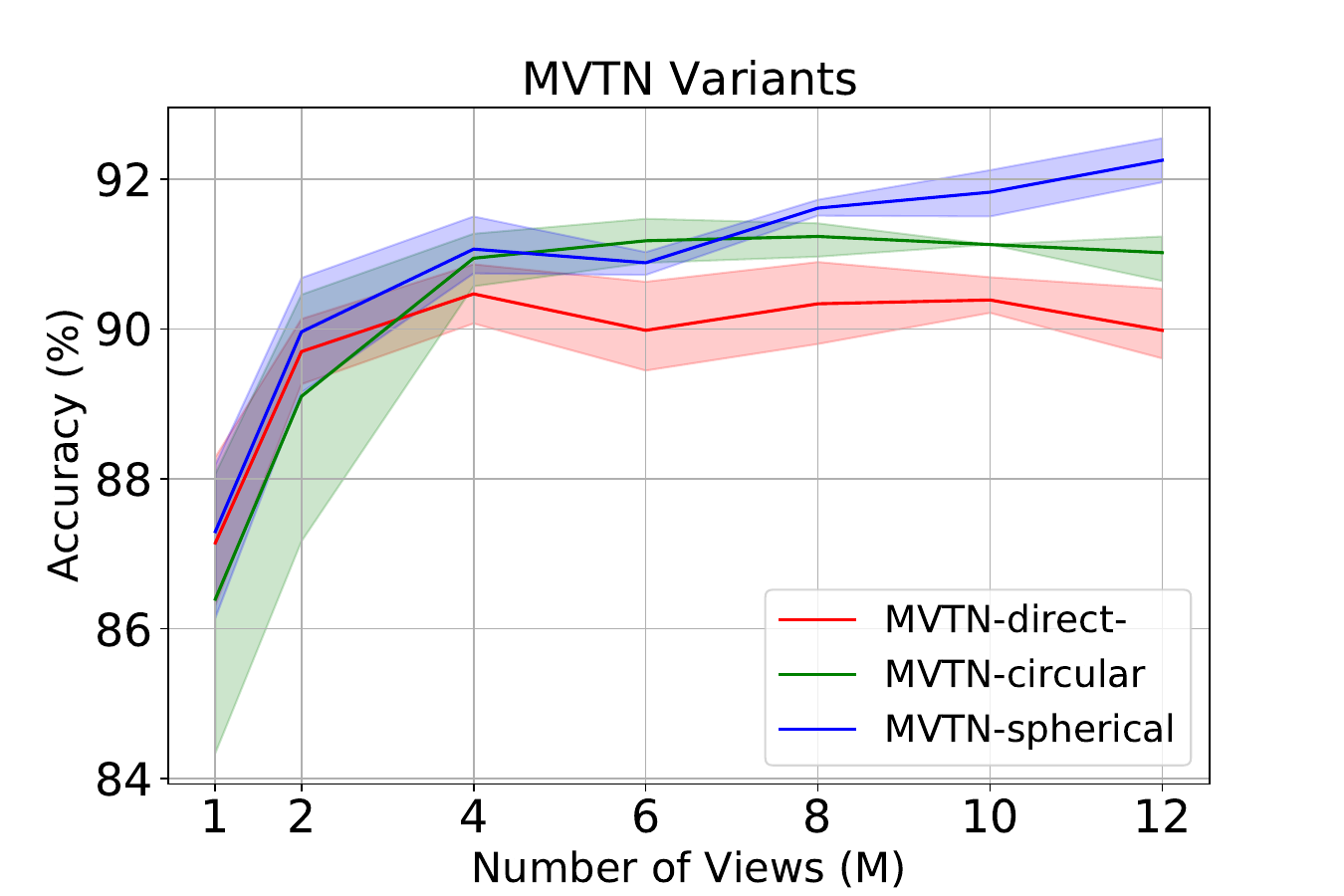}
    \caption{\textbf{Variants of MVTN.} We plot test accuracy vs. the number of views used in training different variants of our MVTN. Note how MVTN-spherical is generally more stable in achieving better performance on ModelNet40. 95\% confidence interval is also plotted on each setup (repeated four times).}
    \label{fig:mvtn-variants}
\end{figure}

\mysection{MVTN Variants} \label{sec:views-supp}
We study the effect of the number of views $M$ on the performance of different MVTN variants (direct, circular, spherical). 
The experiments are repeated four times, and the average test accuracies with confidence intervals are shown in  \figLabel{\ref{fig:mvtn-variants}}. 
 The plots show how learned MVTN-spherical achieves consistently superior performance across a different number of views. Also, note that MVTN-direct suffers from over-fitting when the number of views is larger than four (\ie it gets perfect training accuracy but deteriorates in test accuracy). This can be explained by observing that the predicted view-points tend to be similar to each other for MVTN-direct when the number of views is large. The similarity in views leads the multi-view network to memorize the training but to suffer in testing.

\mysection{Backbone}
In the main manuscript (Table 6), we study MVTN with ViewGCN as the multi-view network. Here, we study the backbone effect on MVTN with MVCNN as the multi-view network and report all results in Table \ref{tbl:ablation-sup}. The study includes the backbone choice, and the point encoder choice. Note that including more sophisticated backbones does not improve the accuracy

\mysection{Late Fusion}
In the MVTN pipeline, we use a point encoder and a multi-view network. One can argue that an easy way to combine them would be to fuse them later in the architecture. For example, PointNet \cite{pointnet} and MVCNN \cite{mvcnn} can be max pooled together at the last layers and trained jointly. We train such a setup and compare it to MVTN. We observe that  MVTN achieves $91.8\%$ compared to $88.4\%$ by late fusion. More results are reported in Table \ref{tbl:ablation-sup}

\begin{figure}[t]
    \centering
    \includegraphics[width=1.10\linewidth]{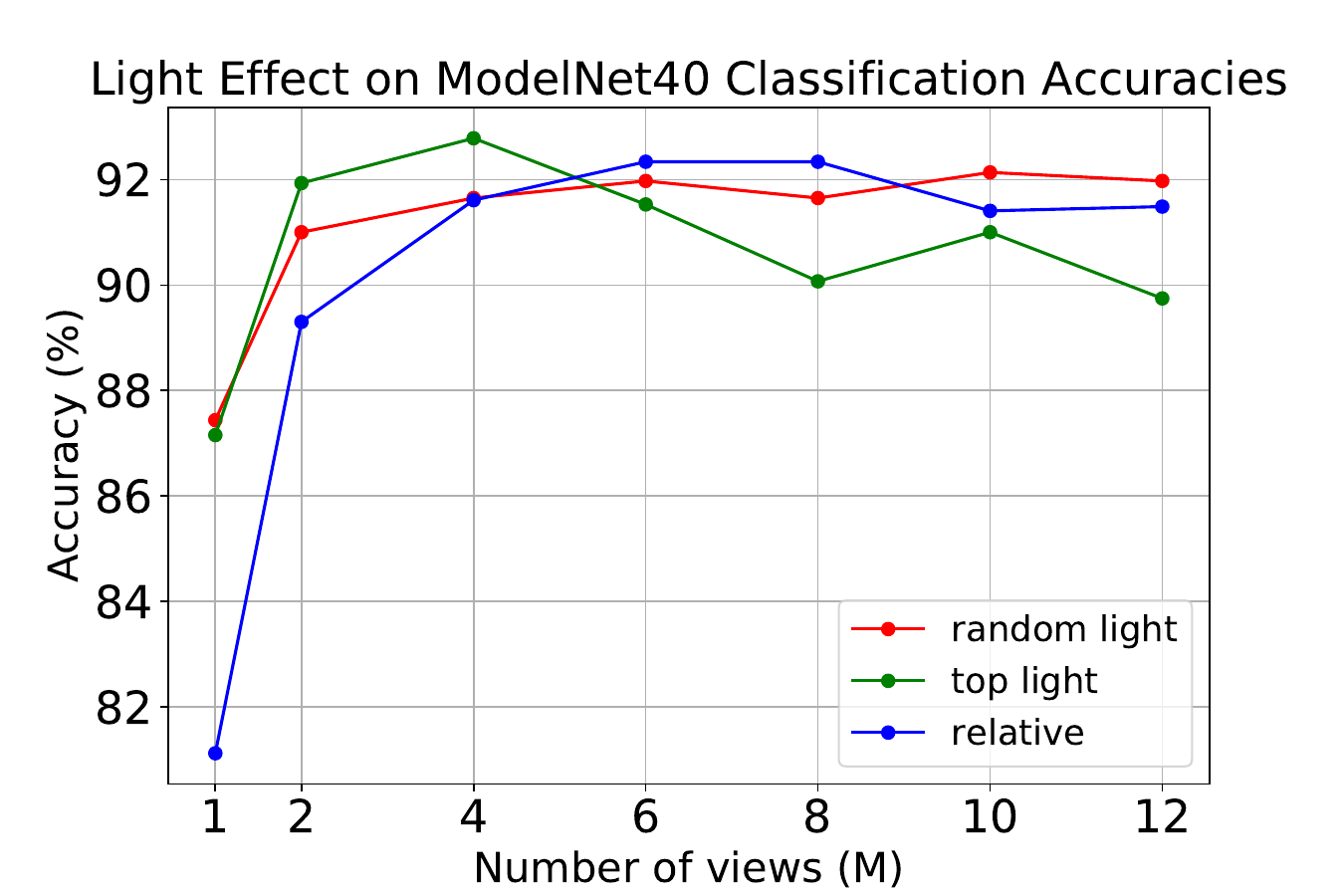}
    \caption{\textbf{Light Direction Effect}. We study the effect of light direction in the performance of the MVTN-circular. We note that randomizing the light direction in training reduce overfitting for larger number of views and leads to better generalization.}
    \label{fig:ablation-light-sup}
\end{figure}

\mysection{Light Direction Effect}
We study the effect of light's direction on the performance of multi-view networks. We note that picking a random light in training helps the network generalize to the test set. Please see \figLabel{\ref{fig:ablation-light-sup}} for the results on circular MVTN with MVCNN when comparing this strategy to fixed light from the top or from camera (\textit{relative}). Note that we use relative light in test time to stabilize the performance.

\mysection{Effect of Object Color}
Our main experiments used random colors for the objects during training and fixed them to white in testing. We tried different coloring approaches, like using a fixed color during training and test. The results are illustrated in Table \ref{tbl:color}. 

\mysection{Image size and number of points}
We study the effect of rendered image size and the number of points sampled in a 4-view MVTN trained on ModelNet40 and report the overall accuracies (averaged over four runs) as follows. For image sizes 160$\times$160, 224$\times$224, and 280$\times$280, the results are 91.0\%, 91.6\%, and 91.9\% respectively. For the number of randomly sampled points $P$ = 512, 1024, and 2048, the results are 91.2\% 91.6\% , and 91.6\% respectively.

\begin{figure}[h]
    \centering
  \quad ~~ Effect of Distance
    \includegraphics[trim= 0 0 1cm 1.7cm , clip, width=0.95\linewidth]{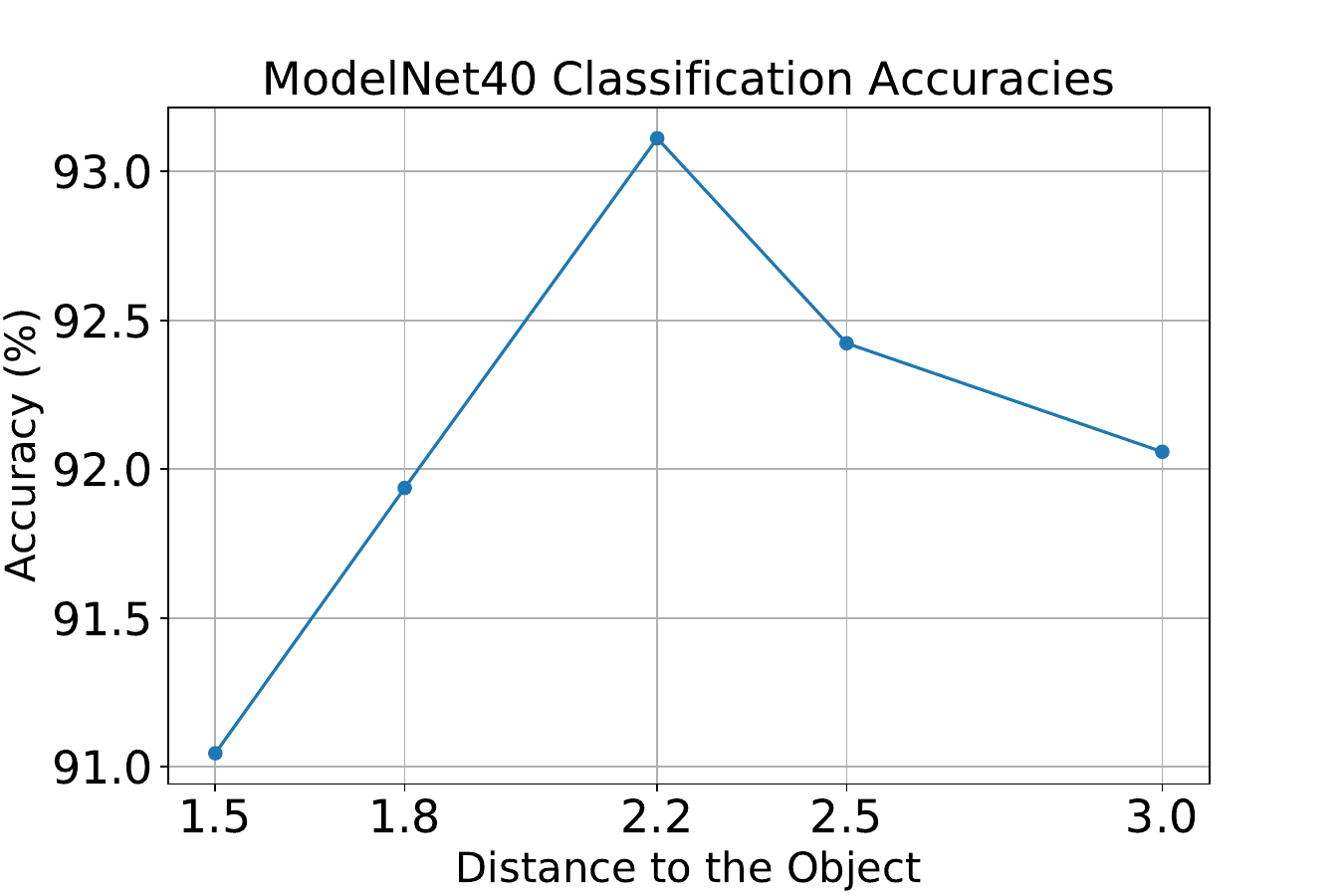}
    \caption{\textbf{Effect of Distance to 3D Object}. We study the effect of changing the distance on training a spherical ViewGCN. We show that the distance of 2.2 units to the center is in between far and close it and gives the best accuracy.}
    \label{fig:ablation-distance-sup}
\end{figure}

\mysection{Learning Distance to the Object}
One possible ablation to the MVTN is to learn the distance to the object. This feature should allow the cameras to get closer to details that might be important to the classifier to understand the object properly. However, we observe that MVTN generally performs worse or does not improve with this setup, and hence, we refrain from learning it. In all of our main experiments, we fixed the distance to $2.2$ units, which is a good middle ground providing best accuracy. Please see \figLabel{\ref{fig:ablation-distance-sup}} for the effect of picking a fixed distance in training spherical ViewGCN.

\subsection{Time and Memory of MVTN}
We compare the time and memory requirements of different parts of our pipeline to assess the MVTN module's contribution. We record FLOPs and MACs to count each module's operations and record the time of a forward pass for a single input sample and the number of parameters for each module. We find in Table \ref{tbl:speed-supp} that MVTN  contributes negligibly to the time and memory requirements of the multi-view networks and the 3D point encoders. 


\begin{table}[h]
\tabcolsep=0.25cm
\centering
\resizebox{0.85\linewidth}{!}{
\begin{tabular}{c|cc} 
\toprule
 & \multicolumn{2}{c}{Object Color}  \\ 
Method & White & Random \\
\midrule
Fixed views & 92.8 $\pm$  0.1  & 92.8 $\pm$  0.1     \\
MVTN (learned)   &93.3 $\pm$  0.1 & \textbf{93.4} $\pm$  0.1    \\
\bottomrule
\end{tabular}
}
\vspace{2pt}
\caption{\small \textbf{Effect of Color Selection}. We ablate selecting the color of the object in training our MVTN and when views are fixed in the spherical configuration. Fixed white color is compared to random colors in training. Note how randomizing the color helps in improving the test accuracy on ModelNet40 a little bit. }
\label{tbl:color}
\end{table}

\begin{table}[h]
\tabcolsep=0.2cm
\centering
\resizebox{0.98\linewidth}{!}{
\begin{tabular}{l|cccc} 
\toprule
Network & FLOPs & MACs & Params. \# & Time\\
\midrule
PointNet &  1.78 G   & 0.89 G    & 3.49 M   &3.34 ms    \\
DGCNN &  10.42 G  &  5.21 G & 0.95 M  & 16.35 ms\\
MVCNN &  43.72 G  & 21.86 G  &  11.20 M &  39.89 ms\\
ViewGCN &  44.19 G &  22.09 G & 23.56 M  & 26.06 ms \\ \midrule
MVTN$^*$ &  \textbf{18.52 K}  & \textbf{9.26 K}  & \textbf{9.09 K}  & \textbf{0.9 ms}\\
MVTN$^\circ$ & 1.78 G & 0.89 G & 4.24 M   & 3.50 ms \\
\bottomrule
\end{tabular}
}
\vspace{2pt}
\caption{\small \textbf{Time and Memory Requirements}. We assess the contribution of the MVTN module to time and memory requirements in the multi-view pipeline. MVTN$^*$ refers to MVTN's regressor excluding the point encoder, while MVTN$^\circ$ refers to the full MVTN module including PointNet as a point encoder.}
\label{tbl:speed-supp}
\end{table}

\subsection{Transferability of MVTN View-Points}
We hypothesize that the views learned by MVTN are transferable across multi-view classifiers. Looking at results in \figLabel{\ref{fig:views-mvt-sup-1}, \ref{fig:views-mvt-sup-2}}, we believe MVTN picks the best views based on the actual shape and is less influenced by the multi-view network.
This means that MVTN learns views that are more representative of the object, making it easier for \textit{any} multi-view network to recognize it.
As such, we ask the following:
\textit{can we transfer the views MVTN learns under one setting to a different multi-view network?}

To test our hypothesis, we take a 12-view MVTN-spherical module trained with MVCNN as a multi-view network and transfer the predicted views to a ViewGCN multi-view network. In this case, we freeze the MVTN module and only train ViewGCN on these learned but fixed views. ViewGCN with transferred MVTN views reaches $93.1\%$ accuracy in classification. It corresponds to a boost of $0.7\%$ from the $92.4\%$ of the original ViewGCN. Although this result is lower than fully trained MVTN($-0.3\%$), we observe a decent transferability between both multi-view architectures.

\begin{table*}[h]
\footnotesize
\setlength{\tabcolsep}{4pt} %
\renewcommand{\arraystretch}{1.1} %
\centering
\resizebox{\hsize}{!}{
\begin{tabular}{c|cc|cc|cc|cc||c} 
\toprule
\textbf{Views} &  \multicolumn{2}{c|}{\textbf{Backbone}}& \multicolumn{2}{c|}{\textbf{Point Encoder}}& \multicolumn{2}{c|}{\textbf{Setup}}&  \multicolumn{2}{c||}{\textbf{Fusion}} & \multicolumn{1}{c}{\textbf{Results}}\\
 \textbf{number} & \textbf{ResNet18} & \textbf{ResNet50} & \textbf{PointNet\cite{pointnet}} &  \textbf{DGCNN\cite{dgcn}} & \textbf{circular} & \textbf{spherical} &  \textbf{late} & \textbf{MVTN}& \textbf{accuracy}     \\
\midrule
6 & \checkmark & - & \checkmark &   - &  \checkmark  & - &  \checkmark & - & 90.48  \% \\ \hline
6 & \checkmark & - & \checkmark &   - &  \checkmark  & - & - &  \checkmark &  91.13 \% \\ \hline
6 & \checkmark & - & \checkmark &   - &  -  & \checkmark  &  \checkmark & - & 89.51  \% \\ \hline
6 & \checkmark & - & \checkmark &   - &  -  & \checkmark  & - &  \checkmark &  91.94 \% \\ \hline
6 & \checkmark & - & - &  \checkmark &  \checkmark   & - &  \checkmark & - & 87.80  \% \\ \hline
6 & \checkmark & - & - &  \checkmark & \checkmark   & - & - &  \checkmark &  91.49 \% \\ \hline
6 & \checkmark & - & - &  \checkmark &  -  & \checkmark  &  \checkmark & - & 89.82  \% \\ \hline
6 & \checkmark & - & - &  \checkmark &  -  & \checkmark  & - &  \checkmark &  91.29 \% \\  \hline

 6 &   - & \checkmark & \checkmark &   - &  \checkmark  & - &  \checkmark & - & 89.10  \% \\ \hline
 6 &   - & \checkmark & \checkmark &   - &  \checkmark  & - & - &  \checkmark &  90.40 \% \\ \hline
 6 &   - & \checkmark & \checkmark &   - &  -  & \checkmark  &  \checkmark & - & 89.22  \% \\ \hline
 6 &   - & \checkmark & \checkmark &   - &  -  & \checkmark  & - &  \checkmark &  90.76 \% \\ \hline
 6 &   - & \checkmark & - &  \checkmark &  \checkmark   & - &  \checkmark & - & 89.99  \% \\ \hline
 6 &   - & \checkmark & - &  \checkmark & \checkmark   & - & - &  \checkmark &  89.91 \% \\ \hline
 6 &   - & \checkmark & - &  \checkmark &  -  & \checkmark  &  \checkmark & - & 89.95  \% \\ \hline
 6 &   - & \checkmark & - &  \checkmark &  -  & \checkmark  & - &  \checkmark &  90.43 \% \\ \midrule
 
  12 & \checkmark & - & \checkmark &   - &  \checkmark  & - &  \checkmark & - & 87.35\% \\ \hline
12 & \checkmark & - & \checkmark &   - &  \checkmark  & - & - &  \checkmark &  90.68\% \\ \hline
12 & \checkmark & - & \checkmark &   - &  -  & \checkmark  &  \checkmark & - & 88.41\% \\ \hline
12 & \checkmark & - & \checkmark &   - &  -  & \checkmark  & - &  \checkmark &  91.82 \\ \hline
12 & \checkmark & - & - &  \checkmark &  \checkmark   & - &  \checkmark & - & 90.24\% \\ \hline
12 & \checkmark & - & - &  \checkmark & \checkmark   & - & - &  \checkmark &  90.28\% \\ \hline
12 & \checkmark & - & - &  \checkmark &  -  & \checkmark  &  \checkmark & - & 89.83\% \\ \hline
12 & \checkmark & - & - &  \checkmark &  -  & \checkmark  & - &  \checkmark &  91.98\% \\  \hline

 12 &   - & \checkmark & \checkmark &   - &  \checkmark  & - &  \checkmark & - & 86.87\% \\ \hline
 12 &   - & \checkmark & \checkmark &   - &  \checkmark  & - & - &  \checkmark &  88.86\% \\ \hline
 12 &   - & \checkmark & \checkmark &   - &  -  & \checkmark  &  \checkmark & - & 87.16\% \\ \hline
 12 &   - & \checkmark & \checkmark &   - &  -  & \checkmark  & - &  \checkmark &  88.41\% \\ \hline
 12 &   - & \checkmark & - &  \checkmark &  \checkmark   & - &  \checkmark & - & 90.15\% \\ \hline
 12 &   - & \checkmark & - &  \checkmark & \checkmark   & - & - &  \checkmark &  88.37\% \\ \hline
 12 &   - & \checkmark & - &  \checkmark &  -  & \checkmark  &  \checkmark & - & 90.48\% \\ \hline
 12 &   - & \checkmark & - &  \checkmark &  -  & \checkmark  & - &  \checkmark &  89.63\% \\ 
 \bottomrule
\end{tabular}
}
\vspace{2pt}
\caption{\small \textbf{Ablation Study}. We study the effect of ablating different components of MVTN on the test accuracy on ModelNet40. Namely, we observe that using more complex backbone CNNs (like ResNet50 \cite{resnet}) or a more complex features extractor (like DGCNN \cite{dgcn}) does not increase the performance significantly compared to ResNet18 and PointNet \cite{pointnet} respectively. Furthermore, combining the shape features extractor with the MVCNN \cite{mvcnn} in \textit{late fusion} does not work as well as MVTN with the same architectures. All the reported results are using MVCNN \cite{mvcnn} as multi-view network.}
\label{tbl:ablation-sup}
\end{table*}

\subsection{MVTN Predicted Views}
\begin{figure}[t]
    \centering
        \includegraphics[width=0.98\linewidth]{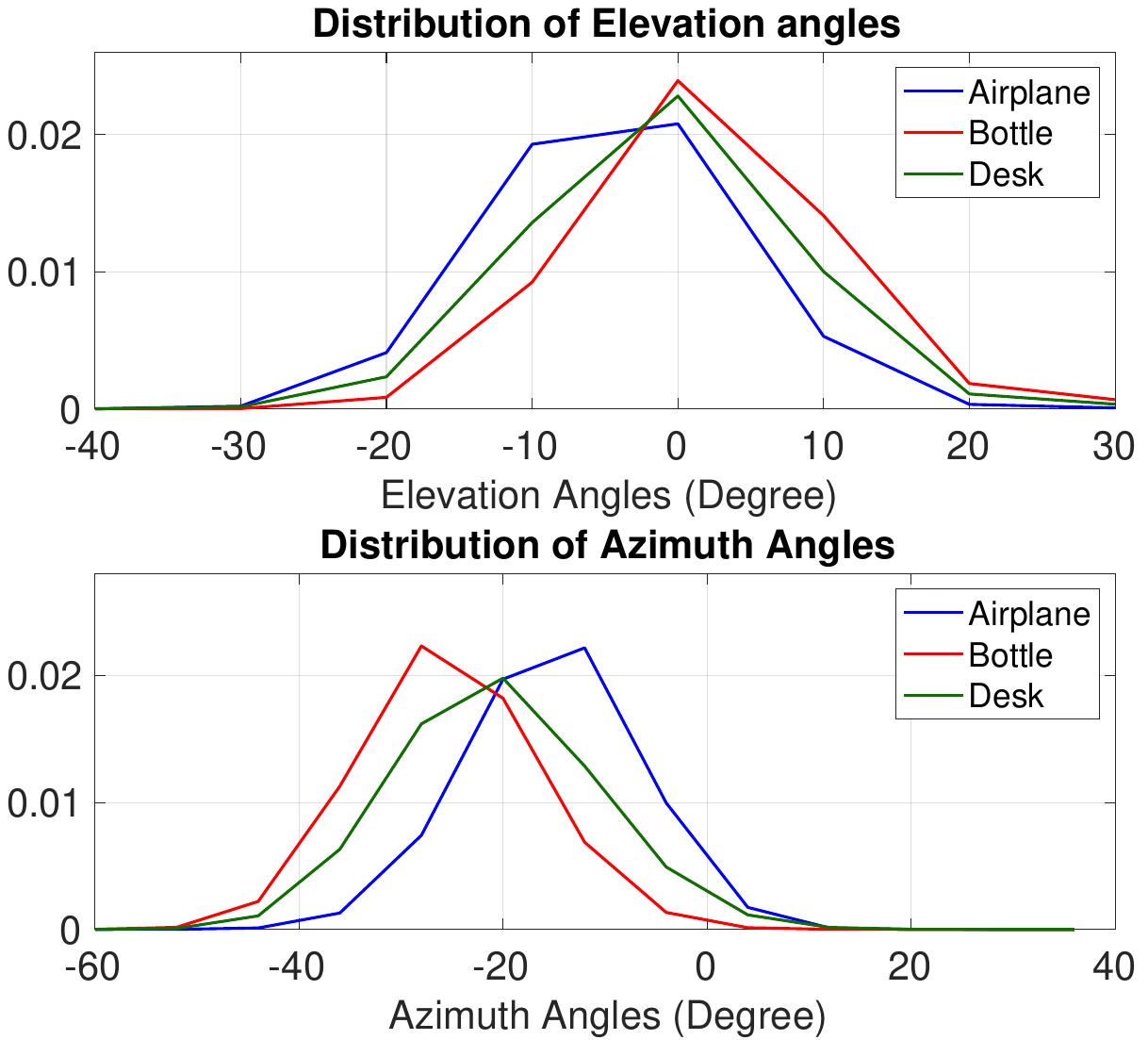}
    \caption{\textbf{Visualizing MVTN learned Views.} We visualize the distribution of azimuth and elevation angles predicted by the MVTN for three different classes. Note that MVTN learns inter-class variations (between different classes) and intra-class variations (on the same class).}
    \label{fig:distribution-sup}
\end{figure}

\begin{figure*} [h] 
\tabcolsep=0.03cm
\begin{tabular}{lc|c}  
\toprule
\textbf{circular:} \hspace{10pt}  & \includegraphics[trim= 4cm 2.7cm 4cm 2.2cm , clip, width = 0.135\linewidth]{images/qualitative_views/circular_cam.jpg} &
\includegraphics[width = 0.7\linewidth ]{images/qualitative_views/circular_rend.jpg} \\ \hline
\textbf{MVTN-circular:}  &\includegraphics[trim= 4cm 2.7cm 4cm 2.2cm , clip, width = 0.135\linewidth]{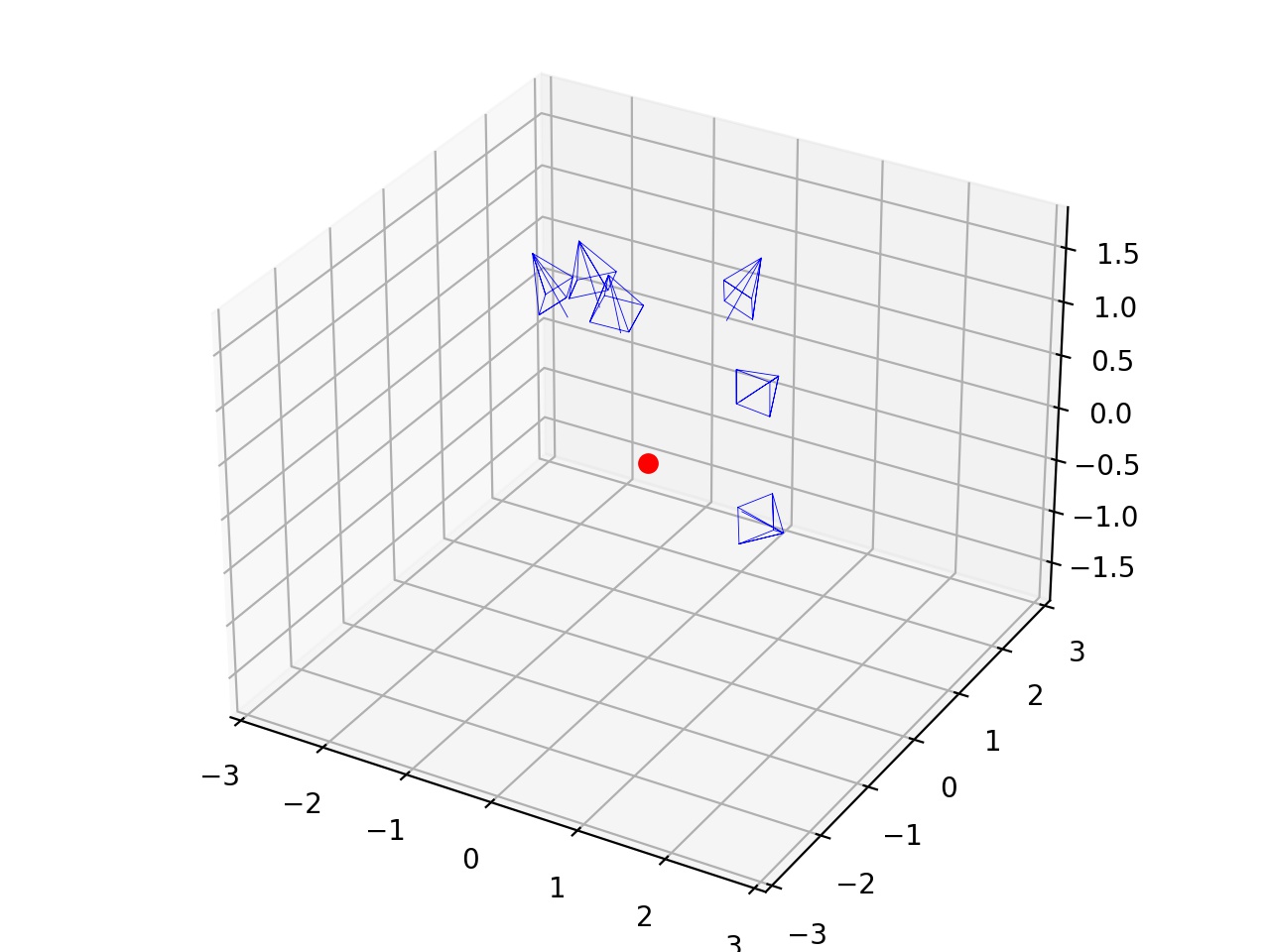} &
\includegraphics[width = 0.7\linewidth]{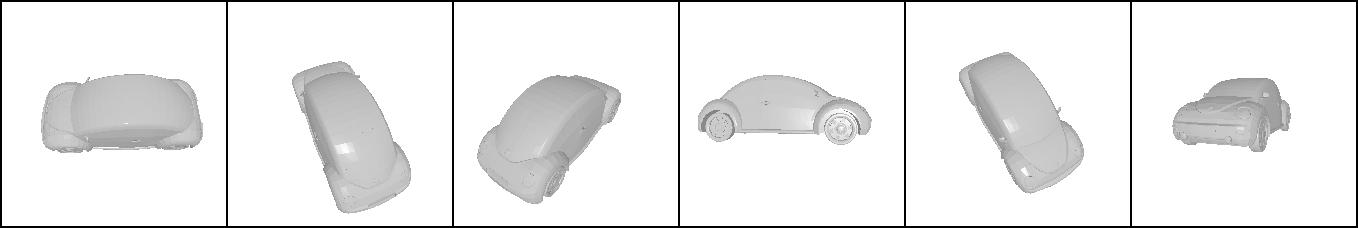} \\ \hline
 \textbf{spherical:} \hspace{10pt} &\includegraphics[trim= 4cm 2.7cm 4cm 2.2cm , clip, width = 0.135\linewidth]{images/qualitative_views/spherical_cam.jpg} &
\includegraphics[width = 0.7\linewidth]{images/qualitative_views/spherical_rend.jpg} \\ \hline
 \textbf{MVTN-spherical:}  & \includegraphics[trim= 4cm 2.7cm 4cm 2.2cm , clip, width = 0.135\linewidth]{images/qualitative_views/mvt_spherical_cam.jpg} &
\includegraphics[width = 0.7\linewidth]{images/qualitative_views/mvt_spherical_rend.jpg} \\ \midrule \midrule

\textbf{circular:} \hspace{10pt}  & \includegraphics[trim= 4cm 2.7cm 4cm 2.2cm , clip, width = 0.135\linewidth]{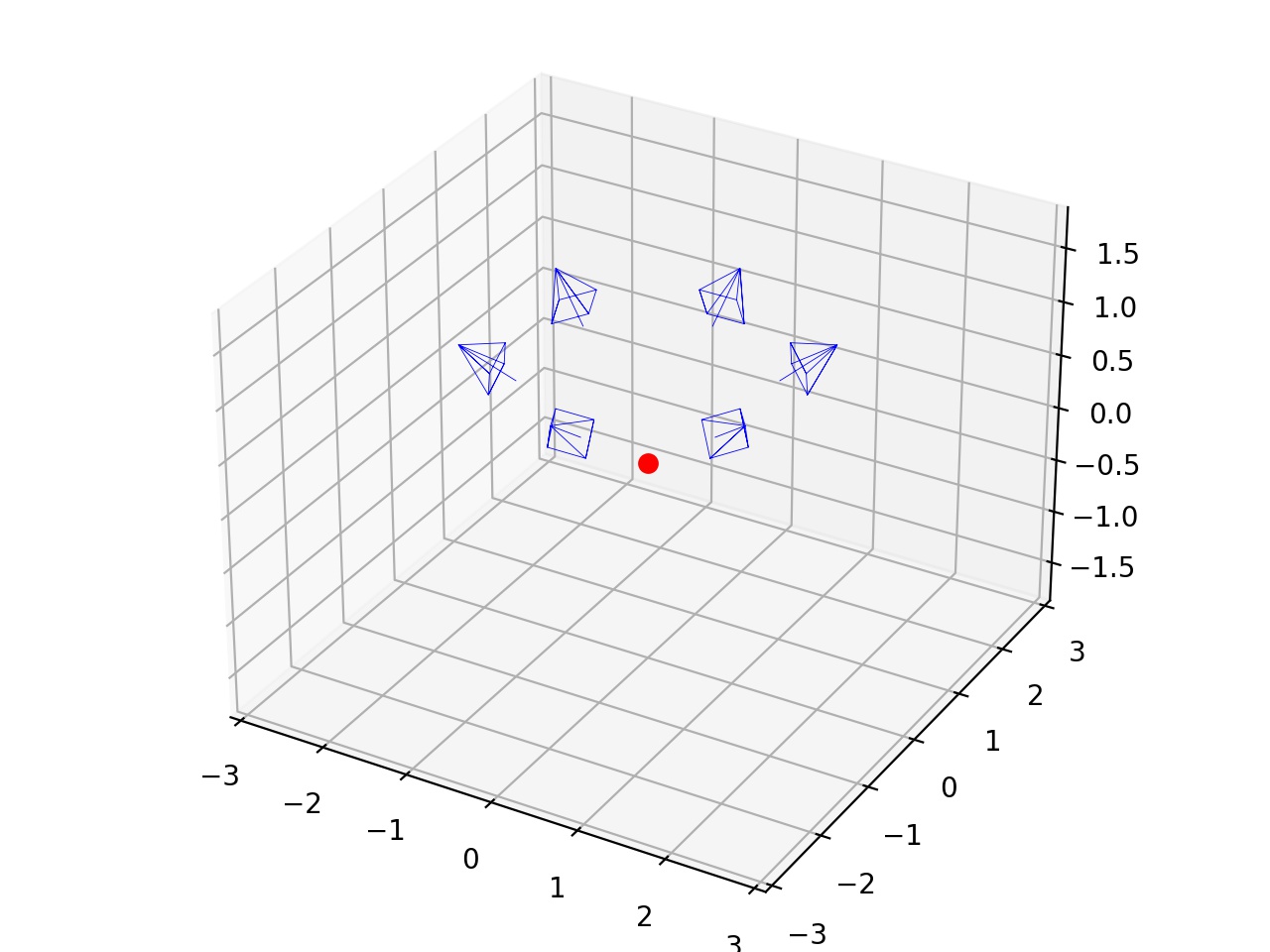} &
\includegraphics[width = 0.7\linewidth ]{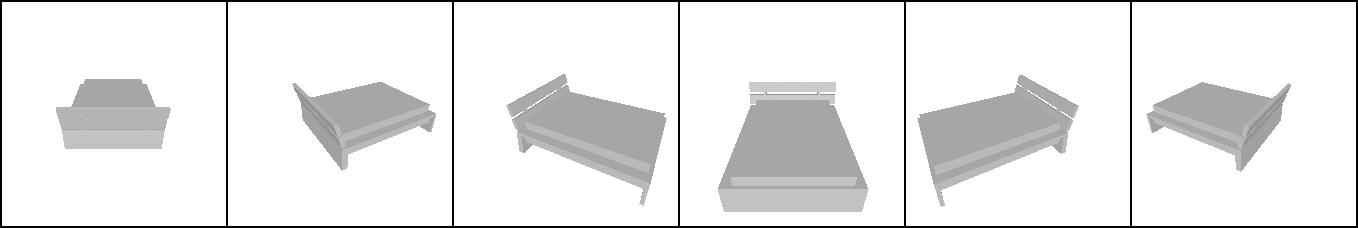} \\ \hline
\textbf{MVTN-circular:}  &\includegraphics[trim= 4cm 2.7cm 4cm 2.2cm , clip, width = 0.135\linewidth]{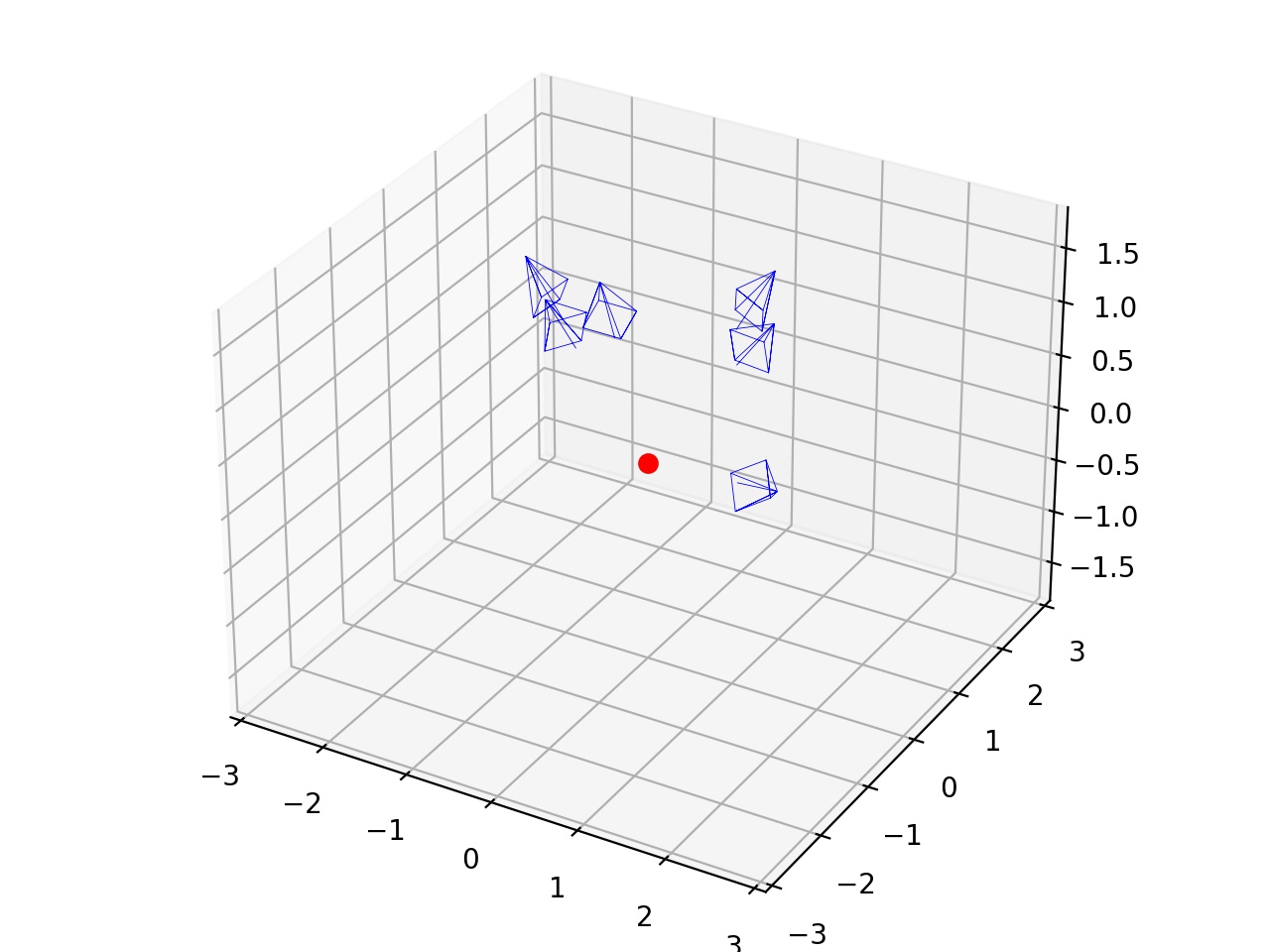} &
\includegraphics[width = 0.7\linewidth]{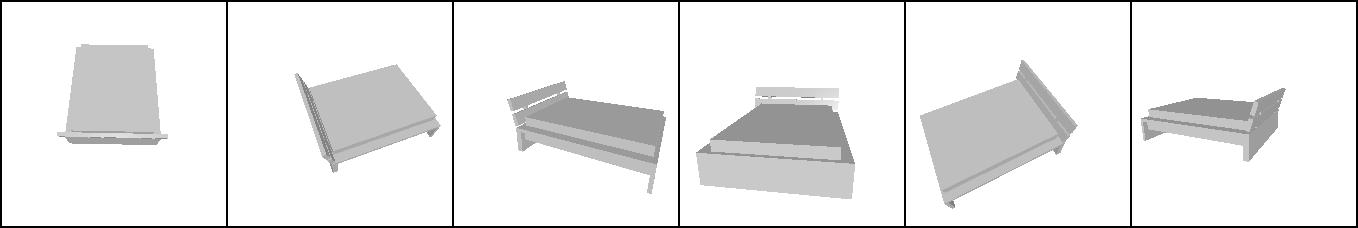} \\ \hline
 \textbf{spherical:} \hspace{10pt} &\includegraphics[trim= 4cm 2.7cm 4cm 2.2cm , clip, width = 0.135\linewidth]{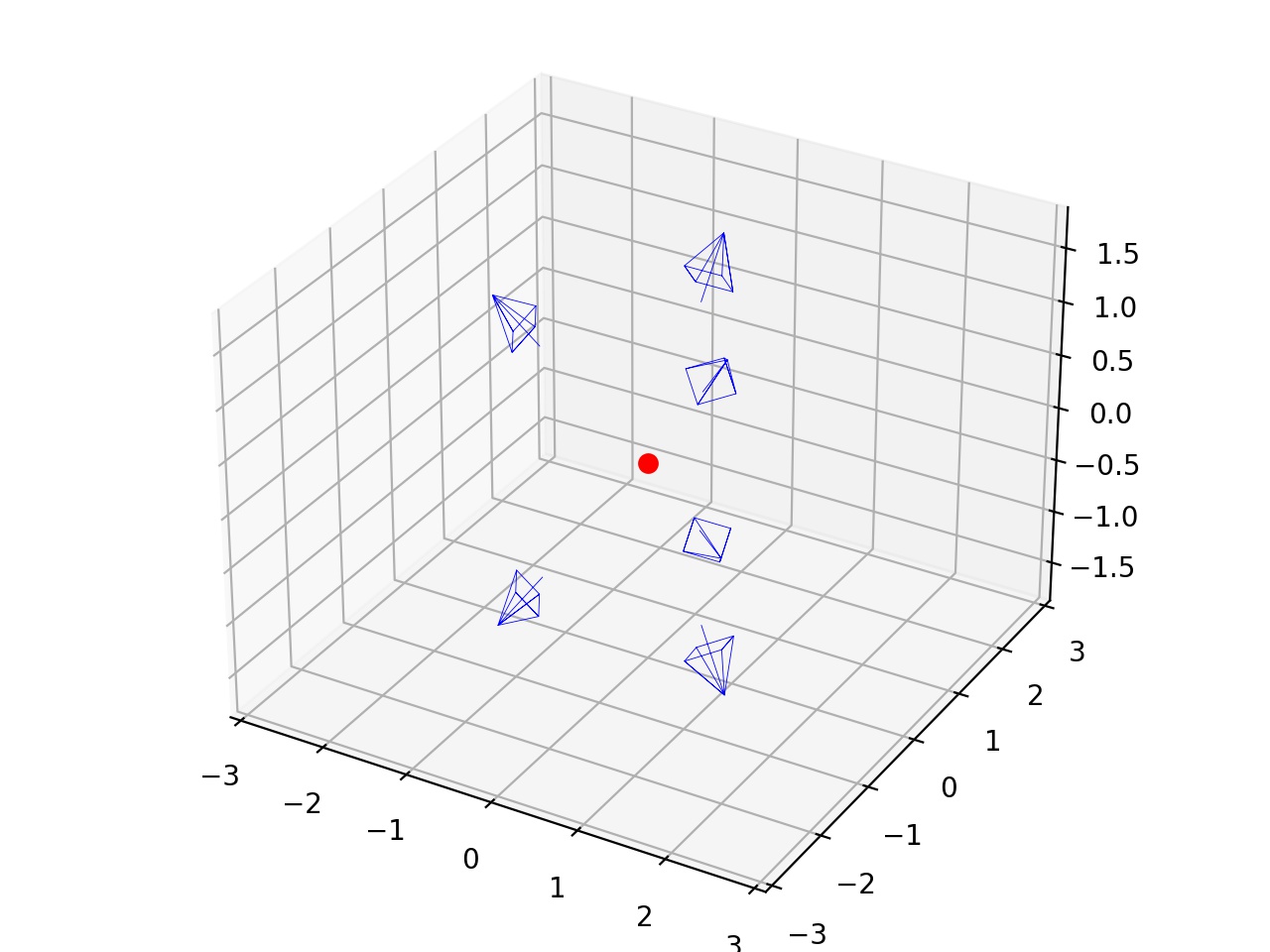} &
\includegraphics[width = 0.7\linewidth]{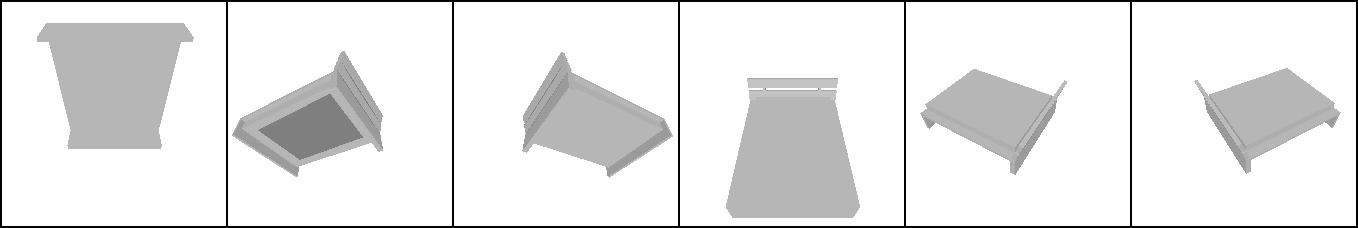} \\ \hline
 \textbf{MVTN-spherical:}  & \includegraphics[trim= 4cm 2.7cm 4cm 2.2cm , clip, width = 0.135\linewidth]{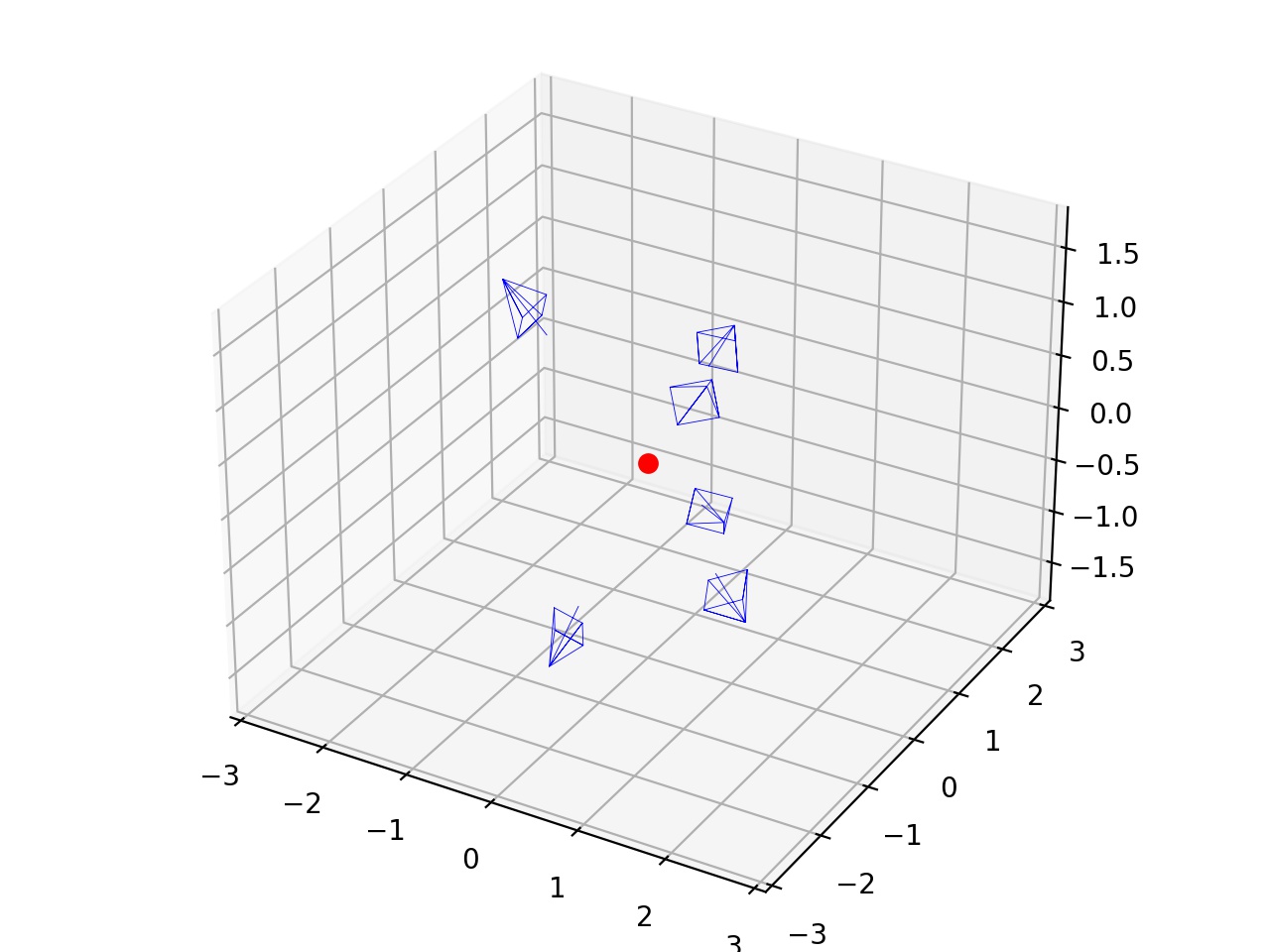} &
\includegraphics[width = 0.7\linewidth]{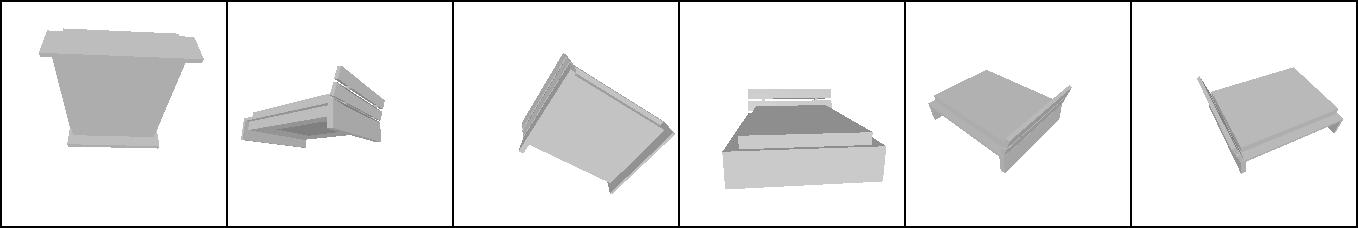} \\ 
\bottomrule
\end{tabular}
\vspace{4pt}
\caption{\small \textbf{Qualitative Examples for MVTN predicted views (I)}: The view setups commonly followed  in the multi-view literature are circular \cite{mvcnn} or spherical \cite{mvviewgcn,mvrotationnet}. The red dot is the center of the object. MVTN-circular/MVTN-spherical are trained to predict the views as offsets to these common configurations. Note that MVTN adjust the original views to make the 3D object better represented by the multi-view images.
}
    \label{fig:views-mvt-sup-1}
\end{figure*}

\begin{figure*} [h] 
\tabcolsep=0.03cm
\begin{tabular}{lc|c}  
\toprule

\textbf{circular:} \hspace{10pt}  & \includegraphics[trim= 4cm 2.7cm 4cm 2.2cm , clip, width = 0.135\linewidth]{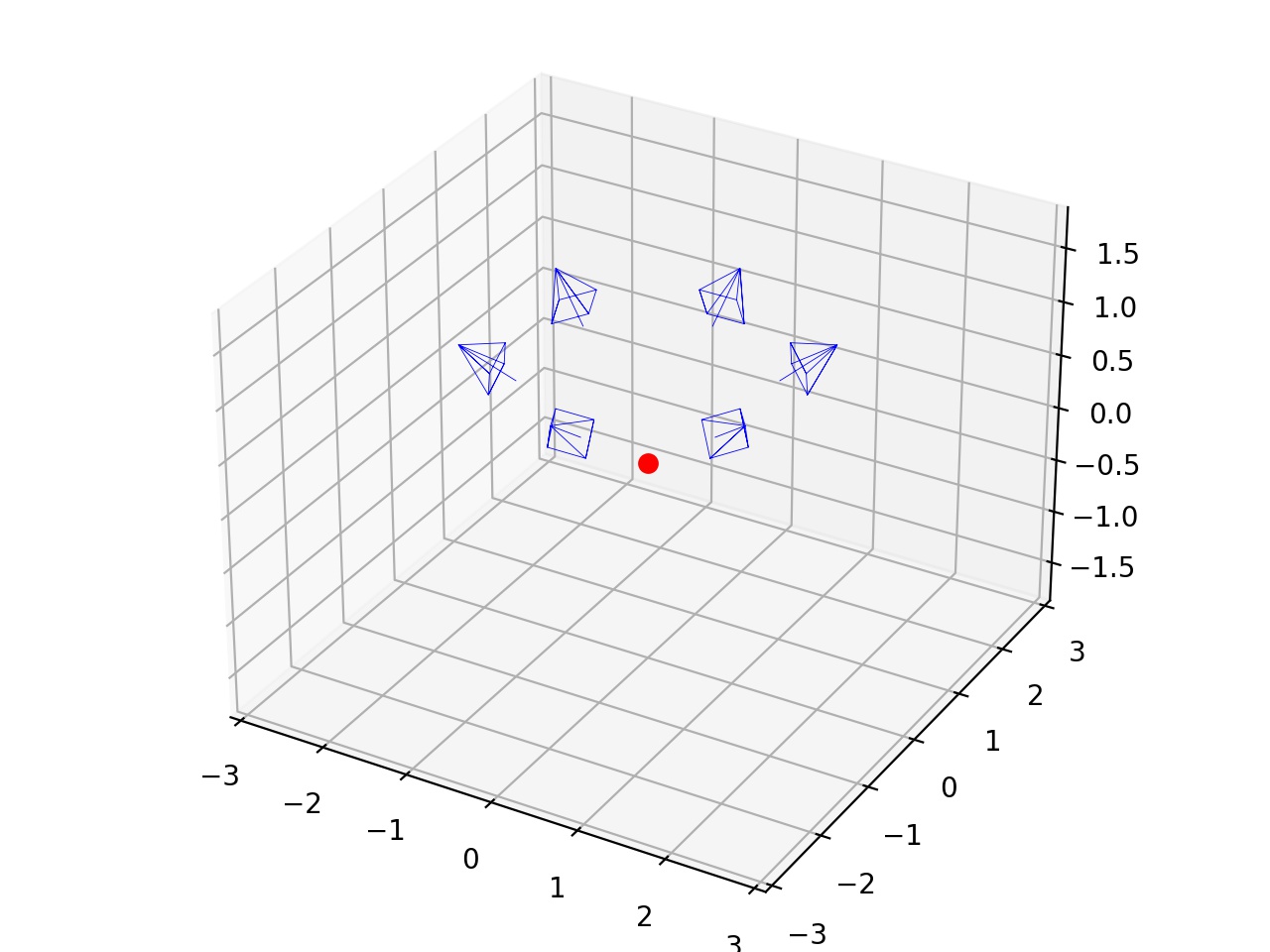} &
\includegraphics[width = 0.7\linewidth ]{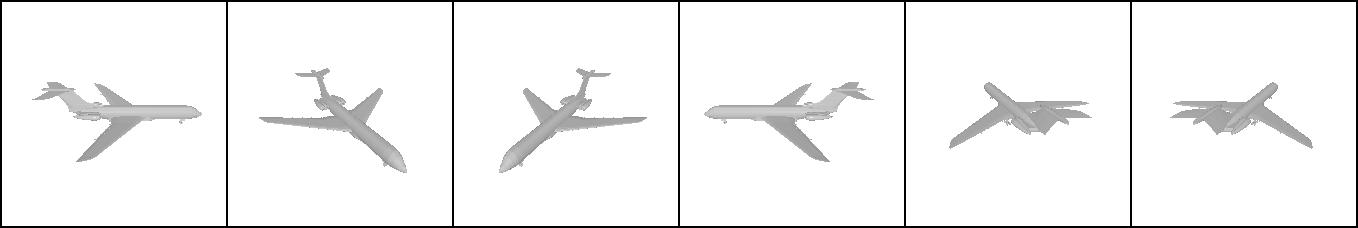} \\ \hline
\textbf{MVTN-circular:}  &\includegraphics[trim= 4cm 2.7cm 4cm 2.2cm , clip, width = 0.135\linewidth]{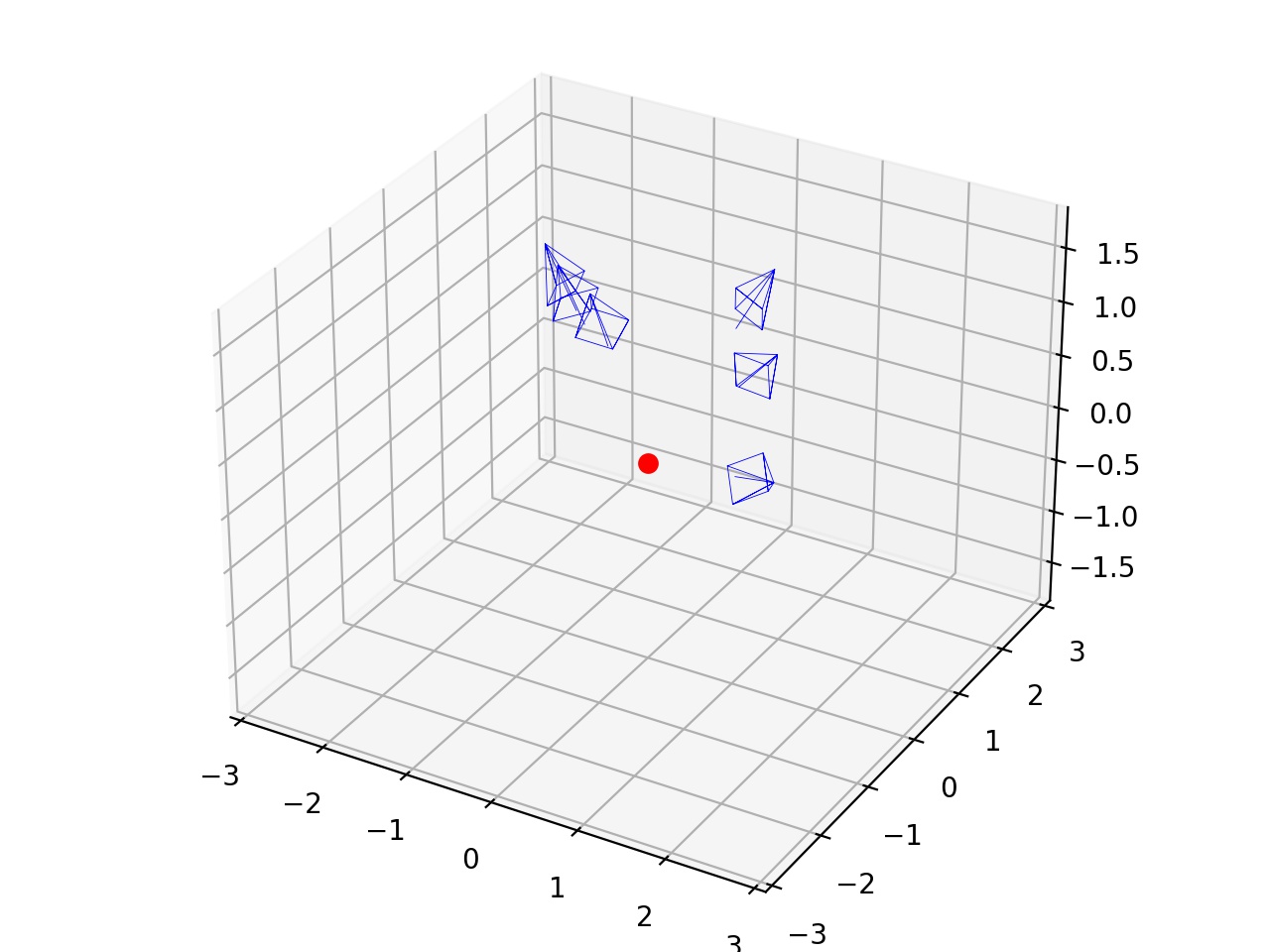} &
\includegraphics[width = 0.7\linewidth]{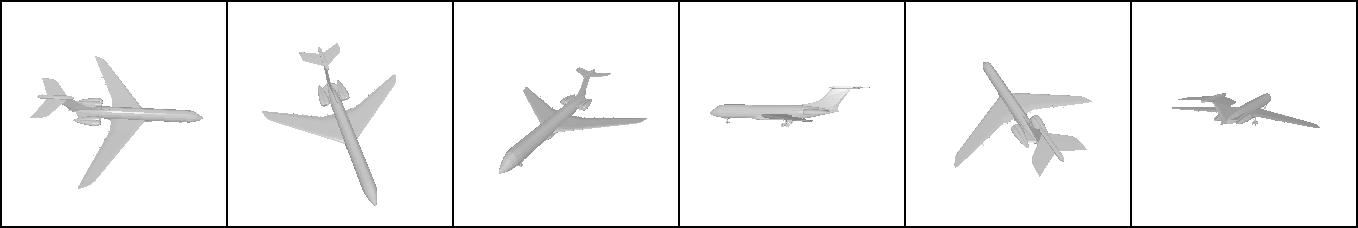} \\ \hline
 \textbf{spherical:} \hspace{10pt} &\includegraphics[trim= 4cm 2.7cm 4cm 2.2cm , clip, width = 0.135\linewidth]{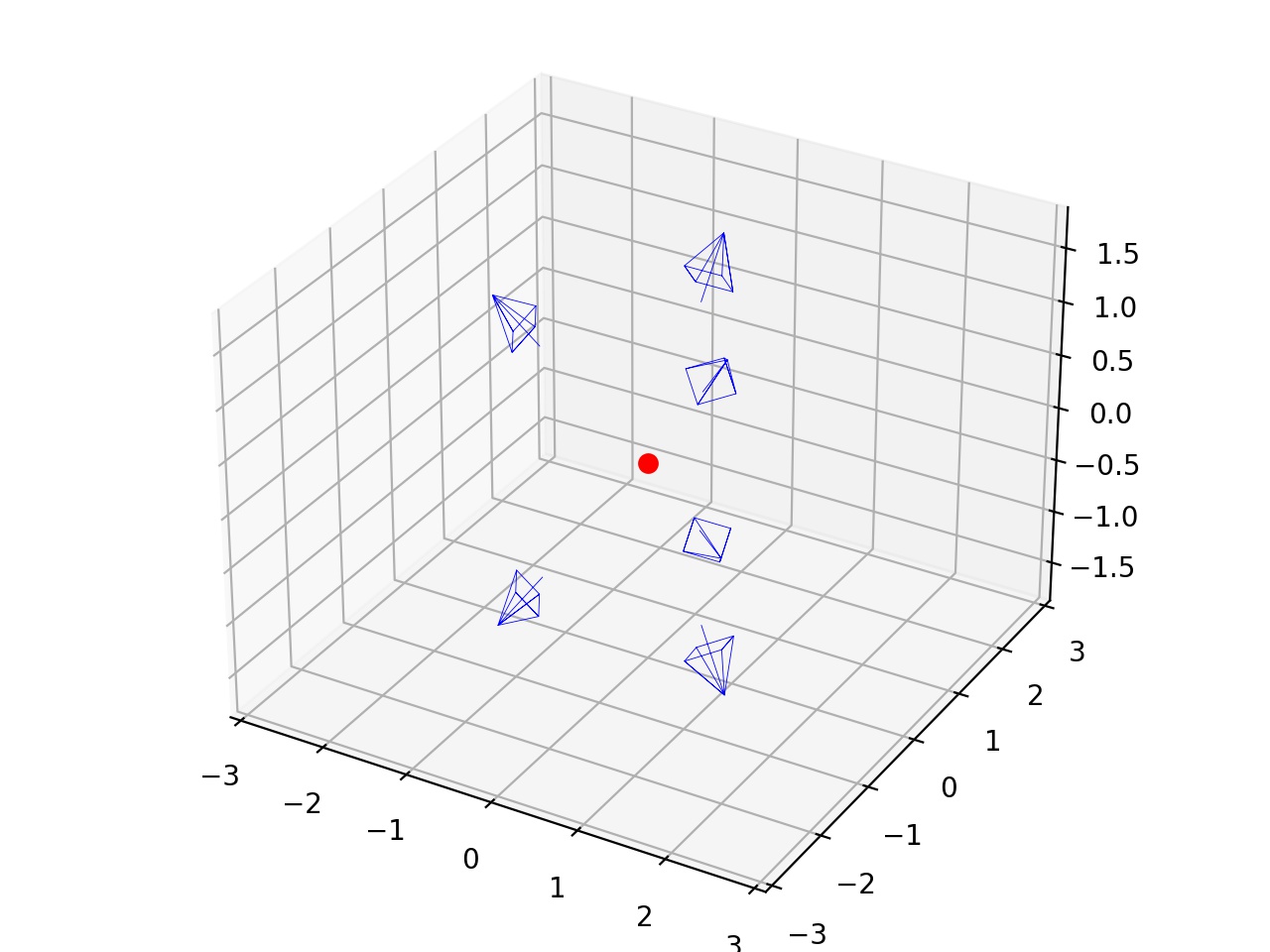} &
\includegraphics[width = 0.7\linewidth]{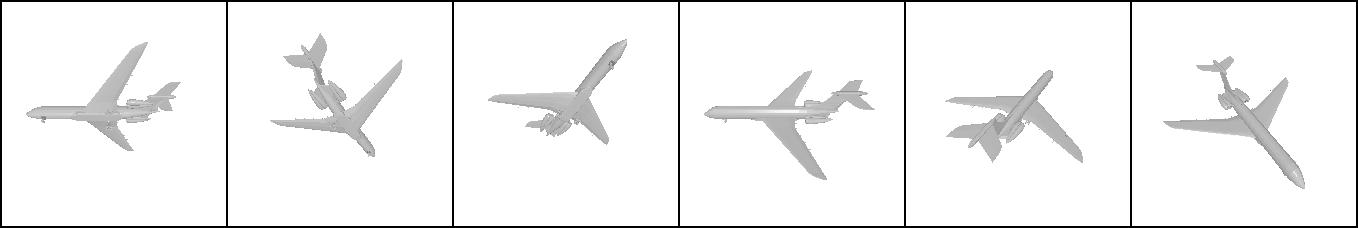} \\ \hline
 \textbf{MVTN-spherical:}  & \includegraphics[trim= 4cm 2.7cm 4cm 2.2cm , clip, width = 0.135\linewidth]{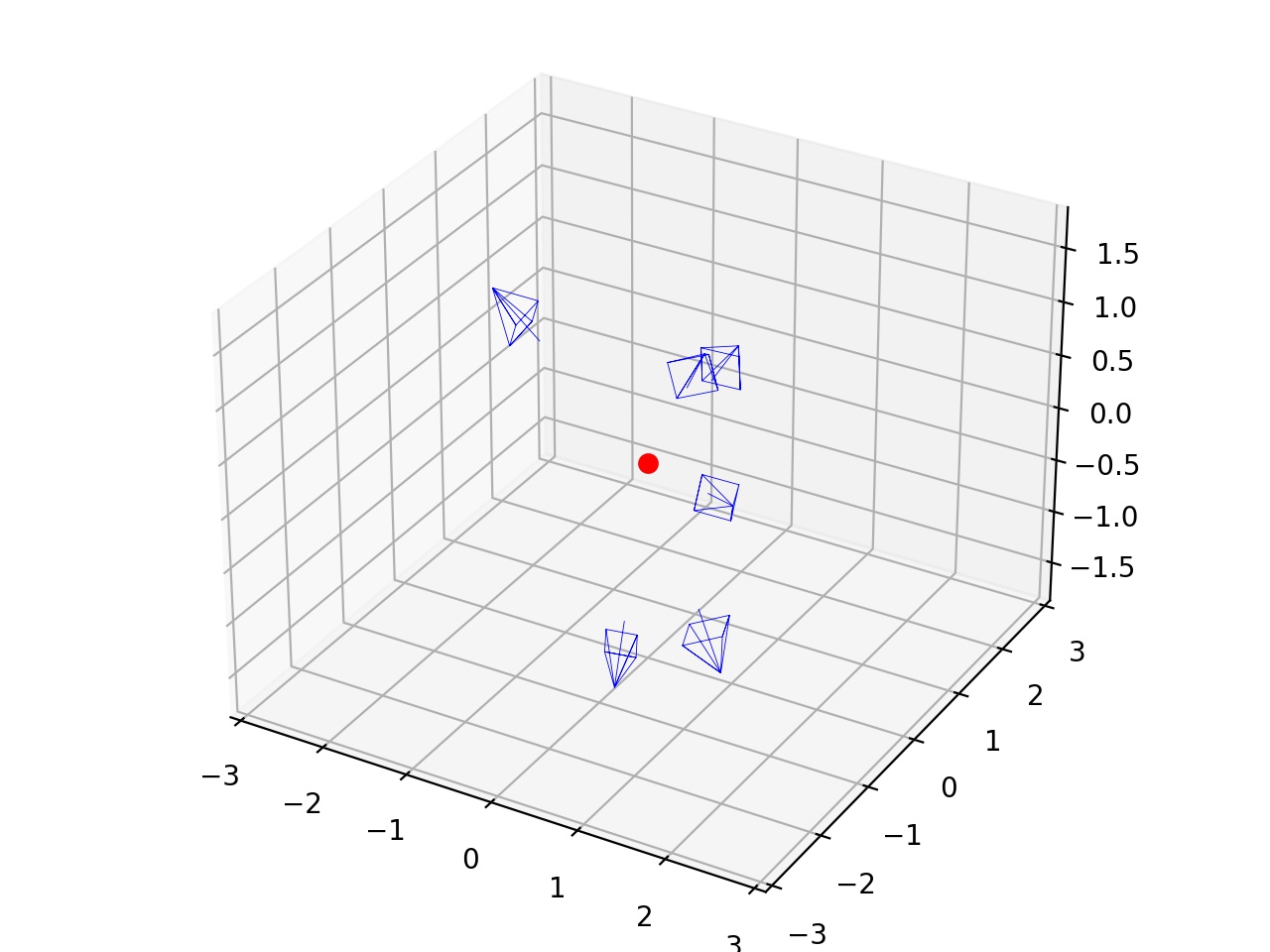} &
\includegraphics[width = 0.7\linewidth]{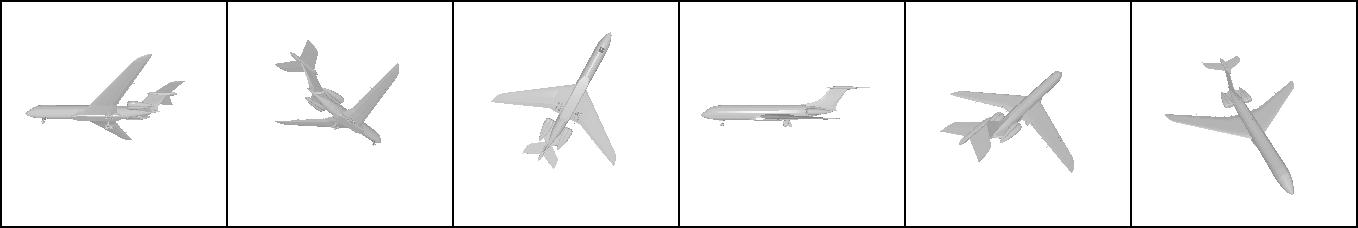} \\ \midrule \midrule

\textbf{circular:} \hspace{10pt}  & \includegraphics[trim= 4cm 2.7cm 4cm 2.2cm , clip, width = 0.135\linewidth]{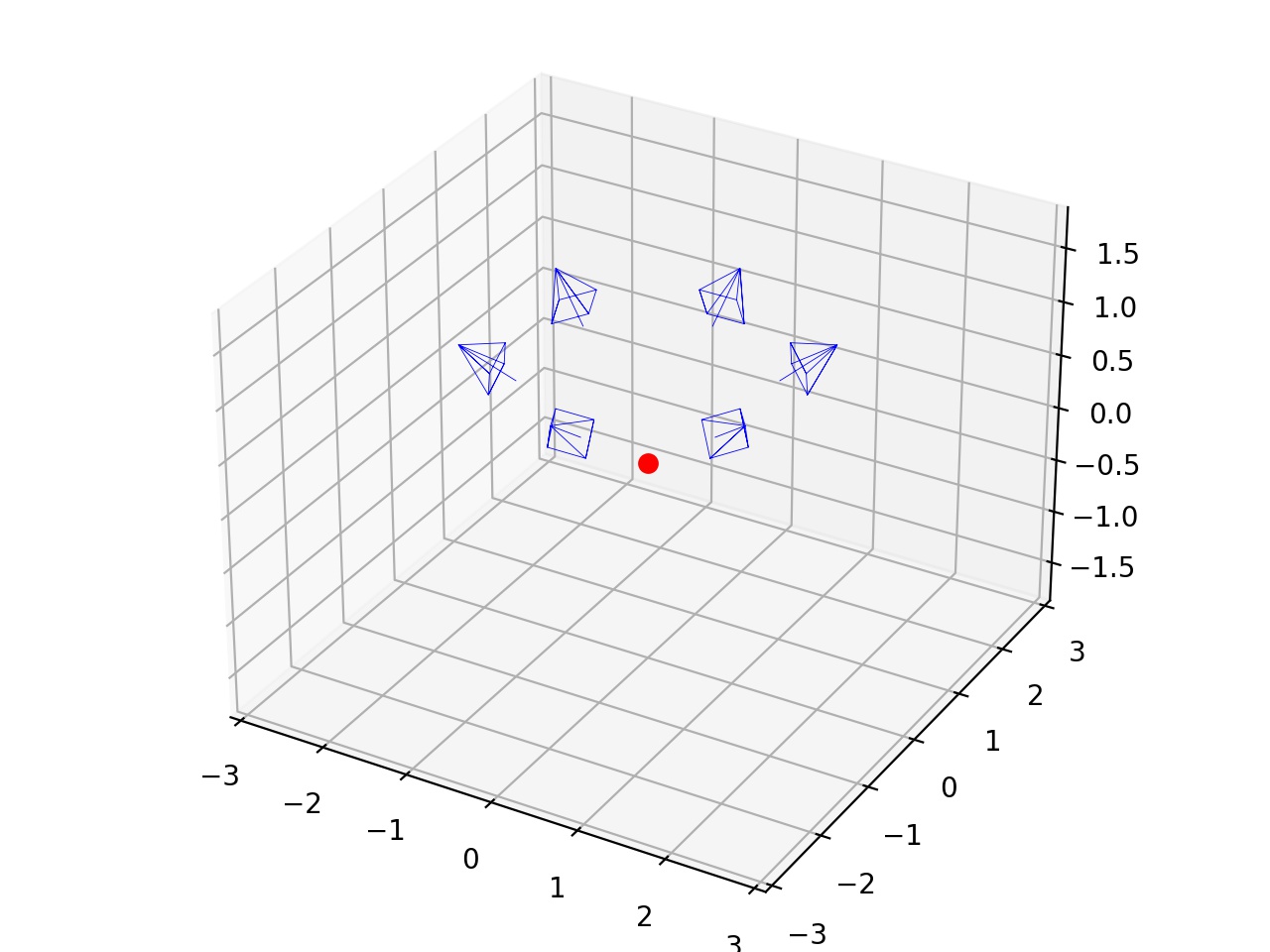} &
\includegraphics[width = 0.7\linewidth ]{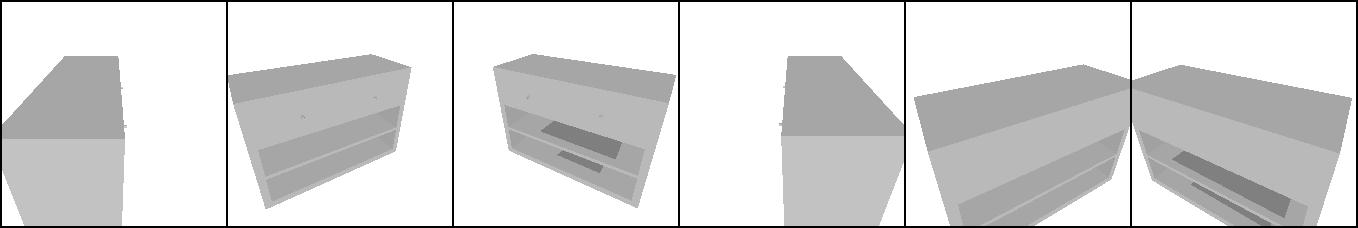} \\ \hline
\textbf{MVTN-circular:}  &\includegraphics[trim= 4cm 2.7cm 4cm 2.2cm , clip, width = 0.135\linewidth]{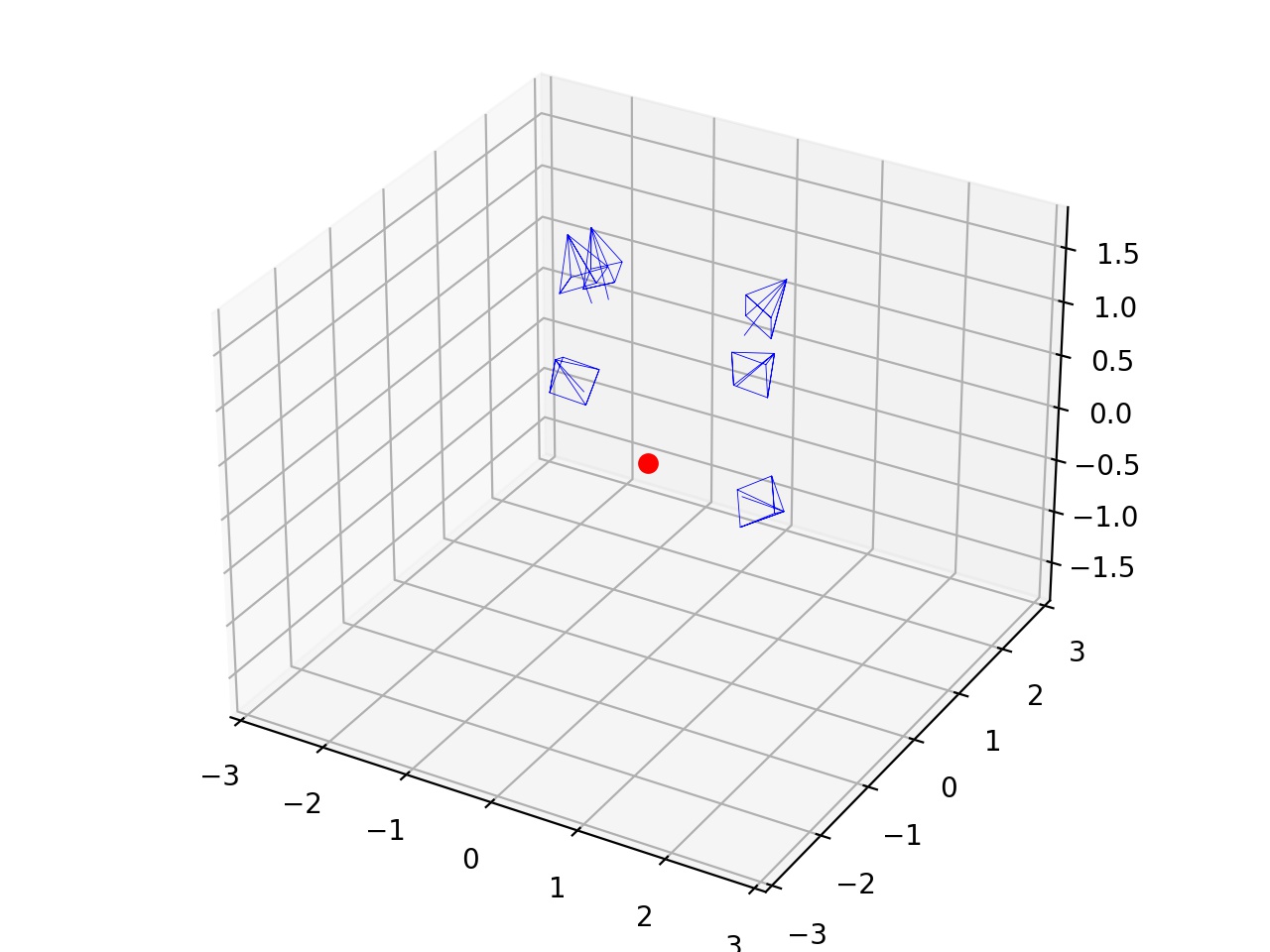} &
\includegraphics[width = 0.7\linewidth]{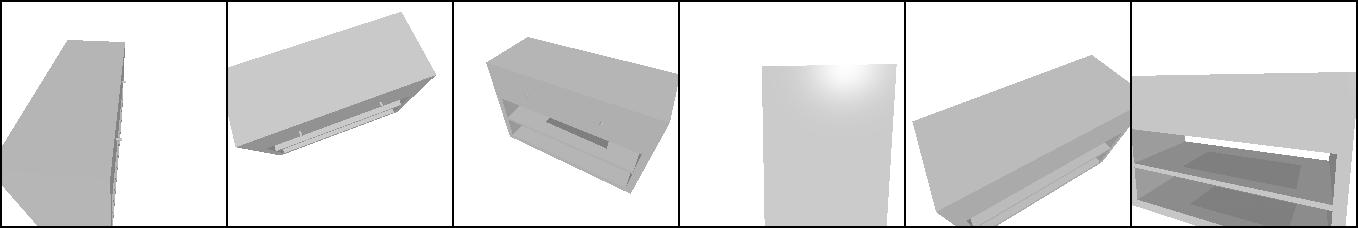} \\ \hline
 \textbf{spherical:} \hspace{10pt} &\includegraphics[trim= 4cm 2.7cm 4cm 2.2cm , clip, width = 0.135\linewidth]{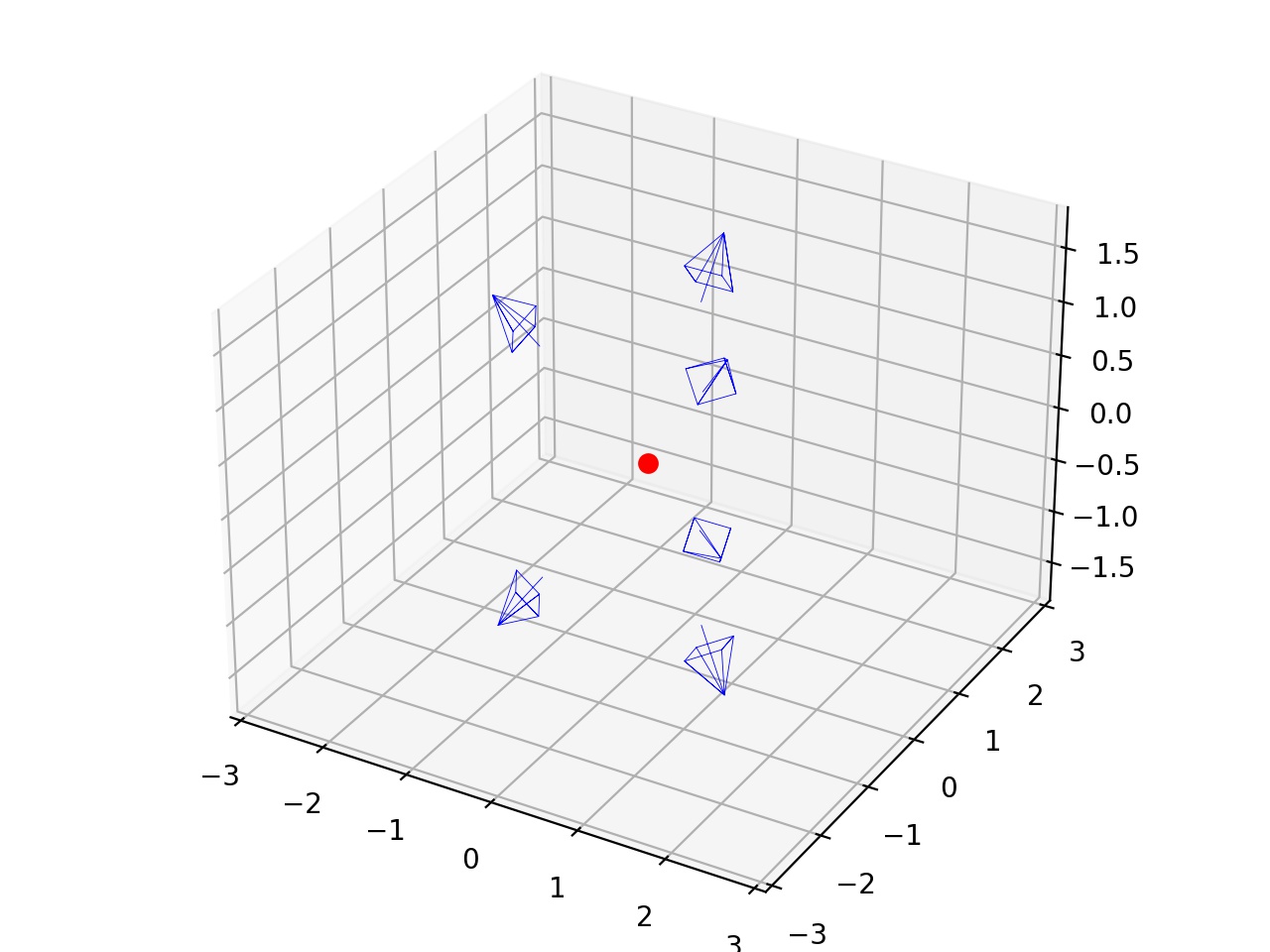} &
\includegraphics[width = 0.7\linewidth]{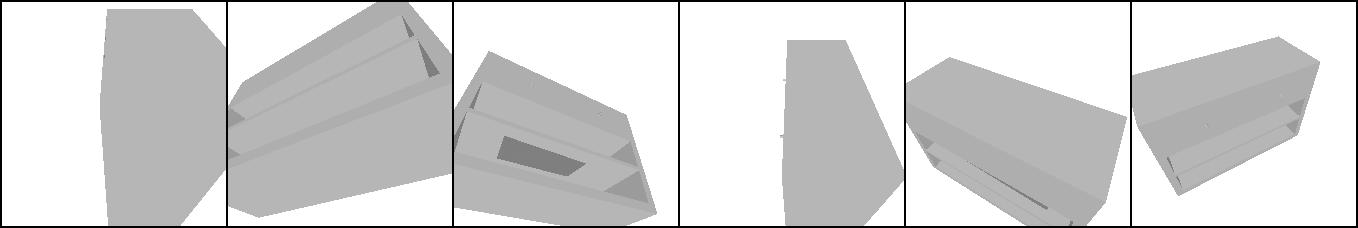} \\ \hline
 \textbf{MVTN-spherical:}  & \includegraphics[trim= 4cm 2.7cm 4cm 2.2cm , clip, width = 0.135\linewidth]{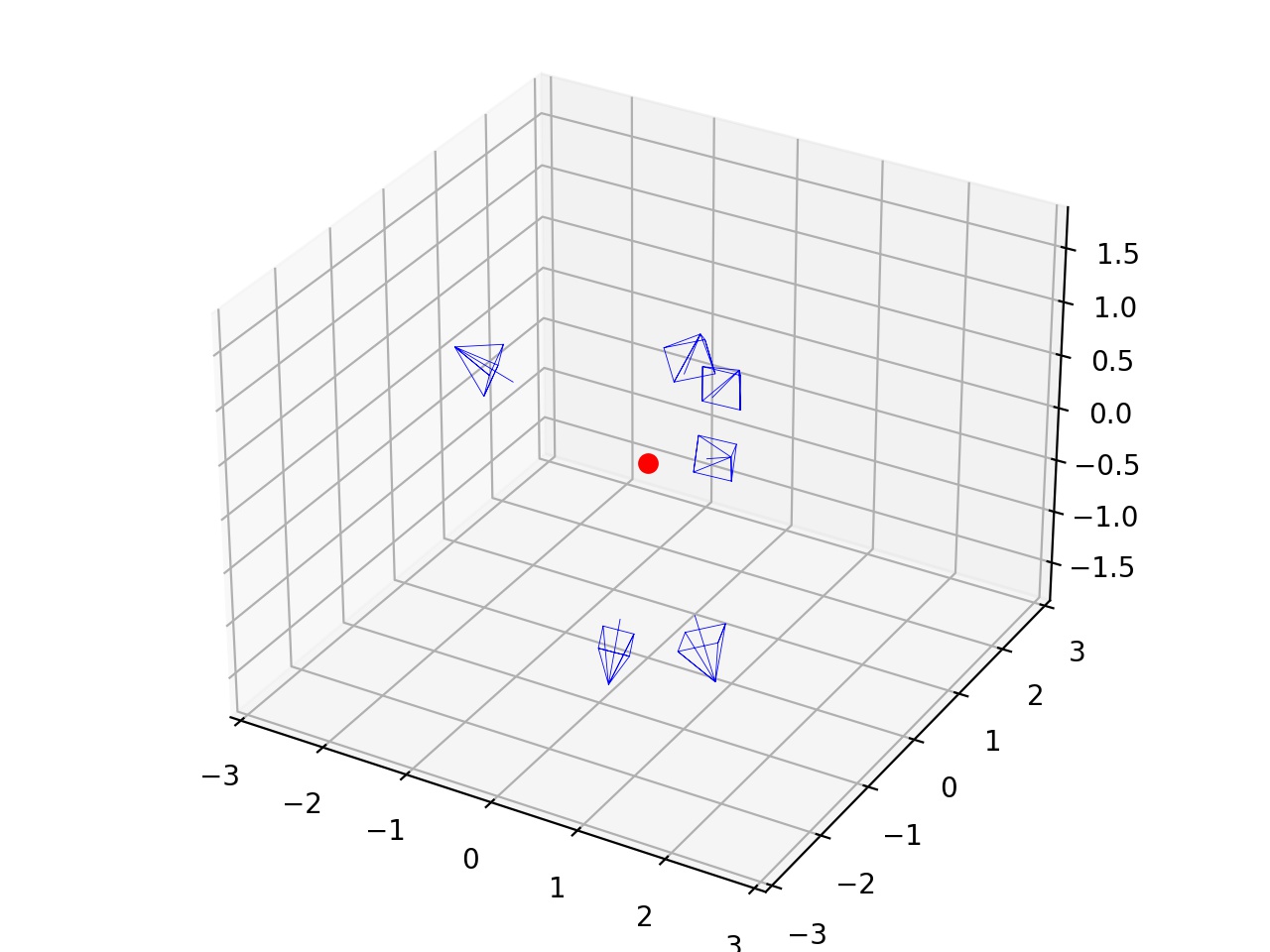} &
\includegraphics[width = 0.7\linewidth]{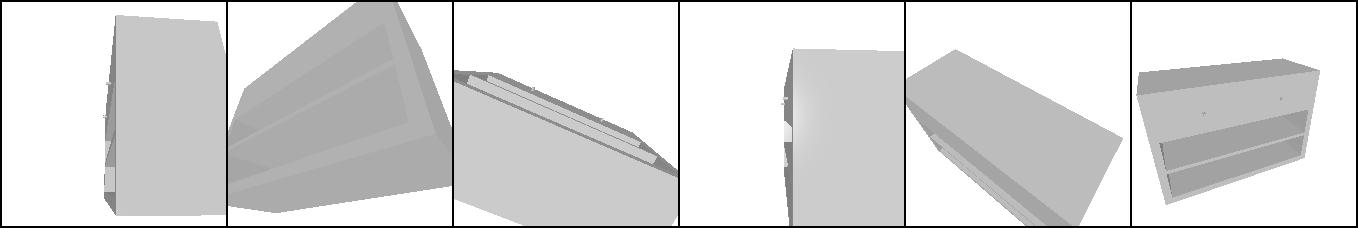} \\
\bottomrule
\end{tabular}
\vspace{4pt}
\caption{\small \textbf{Qualitative Examples for MVTN predicted views (II)}: The view setups commonly followed  in the multi-view literature are circular \cite{mvcnn} or spherical \cite{mvviewgcn,mvrotationnet}. The red dot is the center of the object.  MVTN-circular/MVTN-spherical are trained to predict the views as offsets to these common configurations. Note that MVTN adjust the original views to make the 3D object better represented by the multi-view images.
}
    \label{fig:views-mvt-sup-2}
\end{figure*}

We visualize the distribution of predicted views by MVTN for specific classes in \figLabel{\ref{fig:distribution-sup}}. This is done to ensure that MVTN is learning per-instance views and regressing the same views for the entire class (collapse scenario). We can see that the MVTN distribution of the views varies from one class to another, and the views themselves on the same class have some variance from one instance to another. We also show specific examples for predicted views in \figLabel{\ref{fig:views-mvt-sup-1}, \ref{fig:views-mvt-sup-2}}. Here, we show both the predicted camera view-points and the renderings from these cameras. Note how MVTN shifts every view to better show the discriminative details about the 3D object.
To test that these views are per-instance, we average all the views predicted by our 4-view MVTN for every class and test the trained MVCNN on these fixed per-class views. In this setup, MVTN achieves 90.6\% on ModelNet40, as compared to 91.0\% for the per-instance views and 89\% for the fixed views. 

\subsection{Shape Retrieval Examples}
\begin{figure*} [h] 
\tabcolsep=0.03cm
\resizebox{0.98\linewidth}{!}{
\begin{tabular}{c|cccccccccc}

\includegraphics[width = 0.09090909090909091\linewidth]{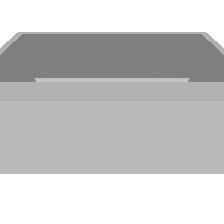} &
\includegraphics[width = 0.09090909090909091\linewidth]{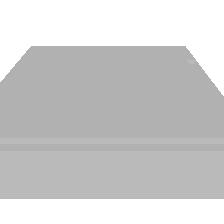}  &
\includegraphics[width = 0.09090909090909091\linewidth]{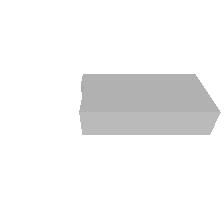}  &
\includegraphics[width = 0.09090909090909091\linewidth]{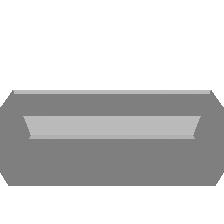}  &
\includegraphics[width = 0.09090909090909091\linewidth]{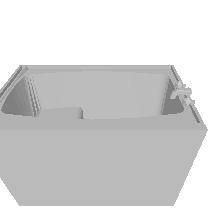}  &
\includegraphics[width = 0.09090909090909091\linewidth]{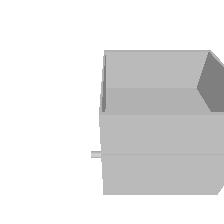}  &
\includegraphics[width = 0.09090909090909091\linewidth]{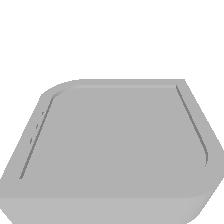}  &
\includegraphics[width = 0.09090909090909091\linewidth]{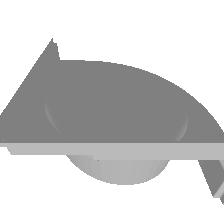}  &
\includegraphics[width = 0.09090909090909091\linewidth]{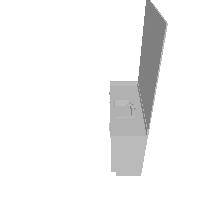}  &
\includegraphics[width = 0.09090909090909091\linewidth]{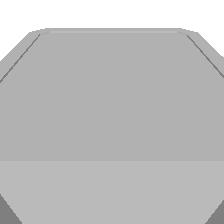}  &
\includegraphics[width = 0.09090909090909091\linewidth]{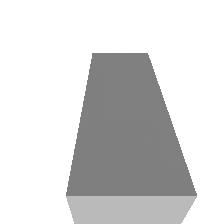}  \\

\includegraphics[width = 0.09090909090909091\linewidth]{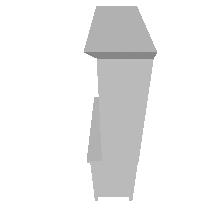} &
\includegraphics[width = 0.09090909090909091\linewidth]{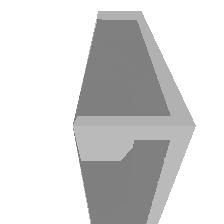}  &
\includegraphics[width = 0.09090909090909091\linewidth]{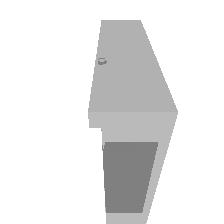}  &
\includegraphics[width = 0.09090909090909091\linewidth]{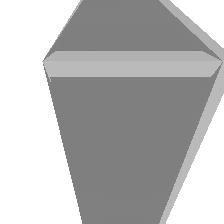}  &
\includegraphics[width = 0.09090909090909091\linewidth]{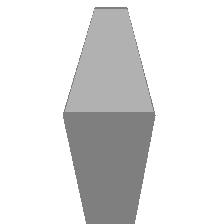}  &
\includegraphics[width = 0.09090909090909091\linewidth]{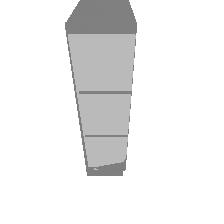}  &
\includegraphics[width = 0.09090909090909091\linewidth]{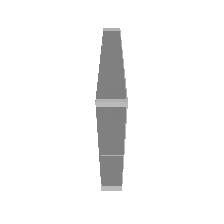}  &
\includegraphics[width = 0.09090909090909091\linewidth]{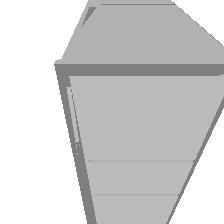}  &
\includegraphics[width = 0.09090909090909091\linewidth]{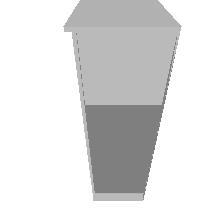}  &
\includegraphics[width = 0.09090909090909091\linewidth]{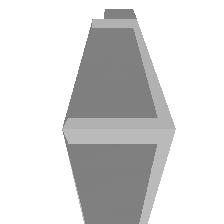}  &
\includegraphics[width = 0.09090909090909091\linewidth]{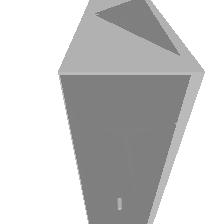}  \\

\includegraphics[width = 0.09090909090909091\linewidth]{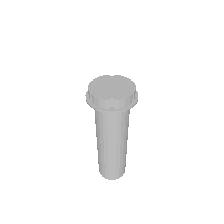} &
\includegraphics[width = 0.09090909090909091\linewidth]{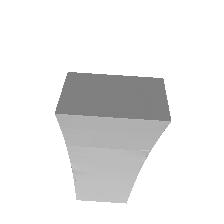}  &
\includegraphics[width = 0.09090909090909091\linewidth]{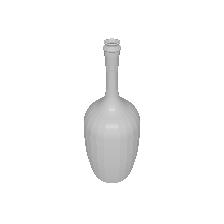}  &
\includegraphics[width = 0.09090909090909091\linewidth]{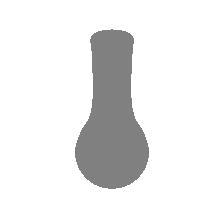}  &
\includegraphics[width = 0.09090909090909091\linewidth]{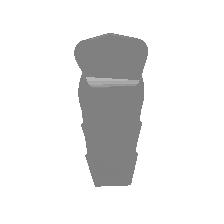}  &
\includegraphics[width = 0.09090909090909091\linewidth]{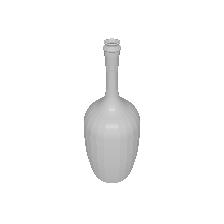}  &
\includegraphics[width = 0.09090909090909091\linewidth]{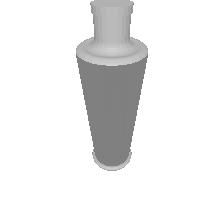}  &
\includegraphics[width = 0.09090909090909091\linewidth]{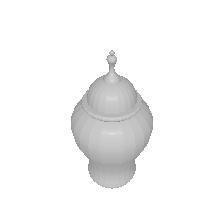}  &
\includegraphics[width = 0.09090909090909091\linewidth]{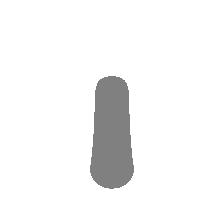}  &
\includegraphics[width = 0.09090909090909091\linewidth]{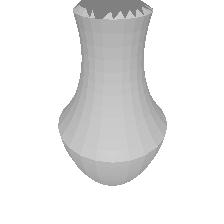}  &
\includegraphics[width = 0.09090909090909091\linewidth]{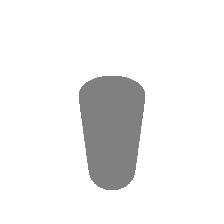}  \\

\includegraphics[width = 0.09090909090909091\linewidth]{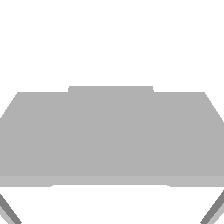} &
\includegraphics[width = 0.09090909090909091\linewidth]{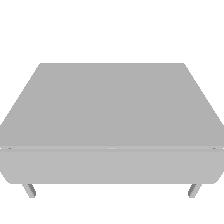}  &
\includegraphics[width = 0.09090909090909091\linewidth]{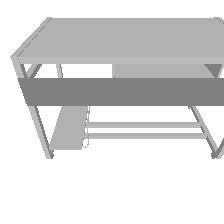}  &
\includegraphics[width = 0.09090909090909091\linewidth]{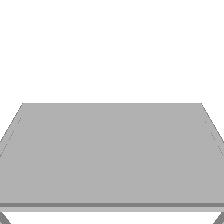}  &
\includegraphics[width = 0.09090909090909091\linewidth]{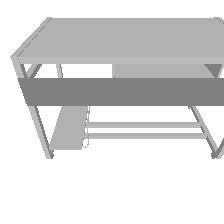}  &
\includegraphics[width = 0.09090909090909091\linewidth]{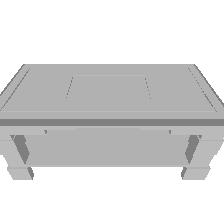}  &
\includegraphics[width = 0.09090909090909091\linewidth]{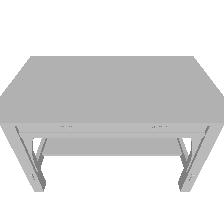}  &
\includegraphics[width = 0.09090909090909091\linewidth]{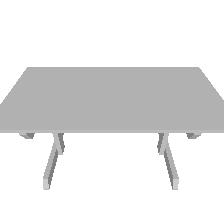}  &
\includegraphics[width = 0.09090909090909091\linewidth]{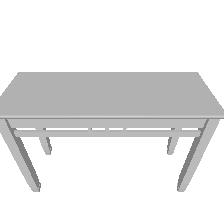}  &
\includegraphics[width = 0.09090909090909091\linewidth]{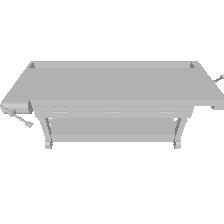}  &
\includegraphics[width = 0.09090909090909091\linewidth]{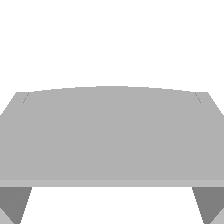}  \\

\includegraphics[width = 0.09090909090909091\linewidth]{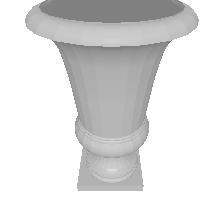} &
\includegraphics[width = 0.09090909090909091\linewidth]{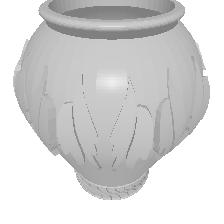}  &
\includegraphics[width = 0.09090909090909091\linewidth]{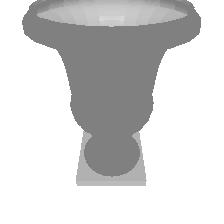}  &
\includegraphics[width = 0.09090909090909091\linewidth]{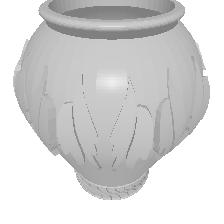}  &
\includegraphics[width = 0.09090909090909091\linewidth]{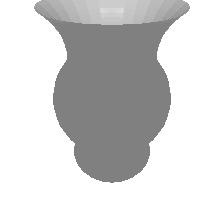}  &
\includegraphics[width = 0.09090909090909091\linewidth]{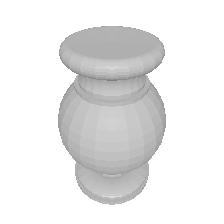}  &
\includegraphics[width = 0.09090909090909091\linewidth]{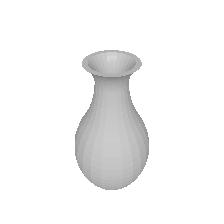}  &
\includegraphics[width = 0.09090909090909091\linewidth]{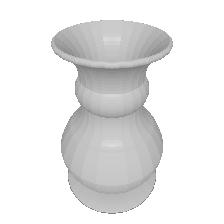}  &
\includegraphics[width = 0.09090909090909091\linewidth]{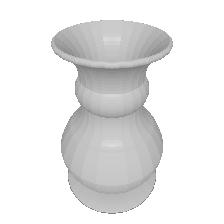}  &
\includegraphics[width = 0.09090909090909091\linewidth]{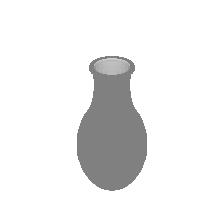}  &
\includegraphics[width = 0.09090909090909091\linewidth]{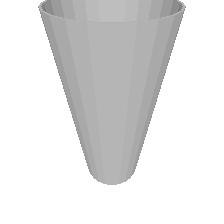}  \\

\includegraphics[width = 0.09090909090909091\linewidth]{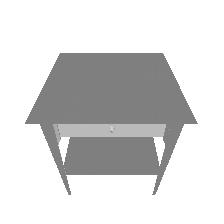} &
\includegraphics[width = 0.09090909090909091\linewidth]{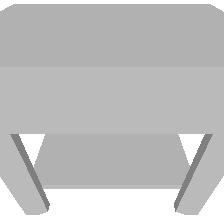}  &
\includegraphics[width = 0.09090909090909091\linewidth]{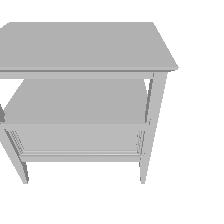}  &
\includegraphics[width = 0.09090909090909091\linewidth]{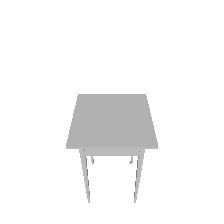}  &
\includegraphics[width = 0.09090909090909091\linewidth]{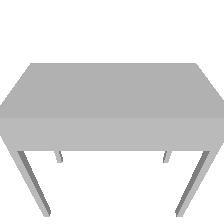}  &
\includegraphics[width = 0.09090909090909091\linewidth]{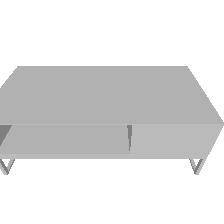}  &
\includegraphics[width = 0.09090909090909091\linewidth]{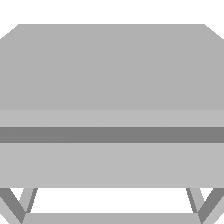}  &
\includegraphics[width = 0.09090909090909091\linewidth]{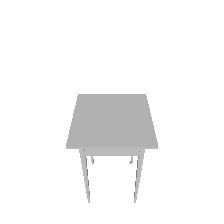}  &
\includegraphics[width = 0.09090909090909091\linewidth]{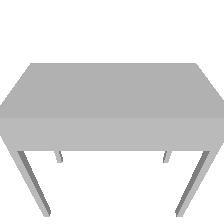}  &
\includegraphics[width = 0.09090909090909091\linewidth]{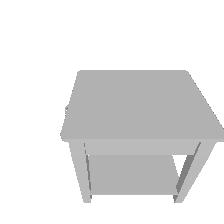}  &
\includegraphics[width = 0.09090909090909091\linewidth]{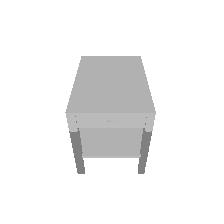}  \\

\includegraphics[width = 0.09090909090909091\linewidth]{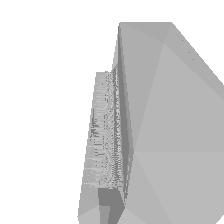} &
\includegraphics[width = 0.09090909090909091\linewidth]{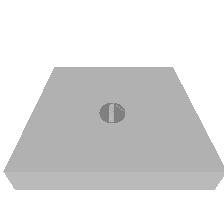}  &
\includegraphics[width = 0.09090909090909091\linewidth]{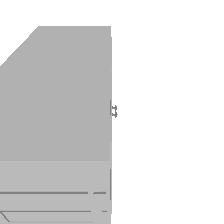}  &
\includegraphics[width = 0.09090909090909091\linewidth]{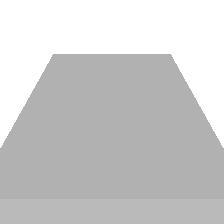}  &
\includegraphics[width = 0.09090909090909091\linewidth]{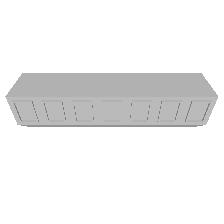}  &
\includegraphics[width = 0.09090909090909091\linewidth]{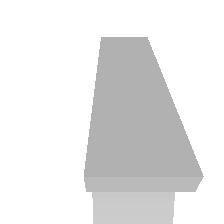}  &
\includegraphics[width = 0.09090909090909091\linewidth]{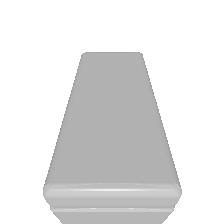}  &
\includegraphics[width = 0.09090909090909091\linewidth]{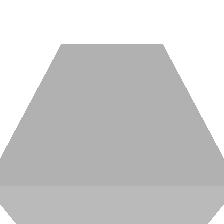}  &
\includegraphics[width = 0.09090909090909091\linewidth]{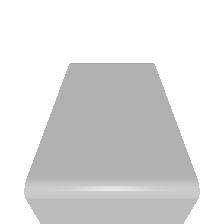}  &
\includegraphics[width = 0.09090909090909091\linewidth]{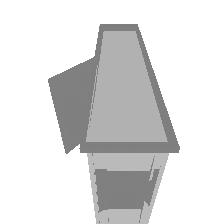}  &
\includegraphics[width = 0.09090909090909091\linewidth]{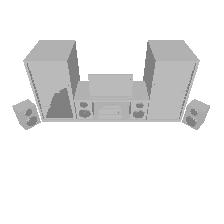}  \\

\includegraphics[width = 0.09090909090909091\linewidth]{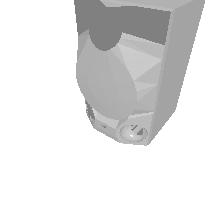} &
\includegraphics[width = 0.09090909090909091\linewidth]{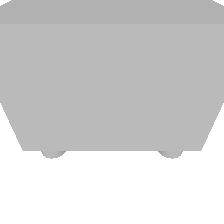}  &
\includegraphics[width = 0.09090909090909091\linewidth]{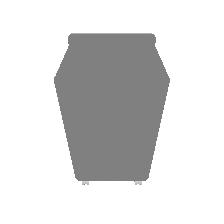}  &
\includegraphics[width = 0.09090909090909091\linewidth]{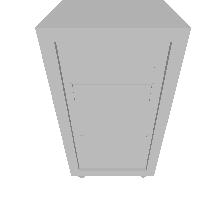}  &
\includegraphics[width = 0.09090909090909091\linewidth]{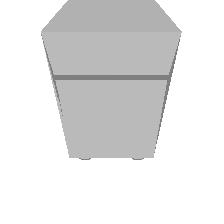}  &
\includegraphics[width = 0.09090909090909091\linewidth]{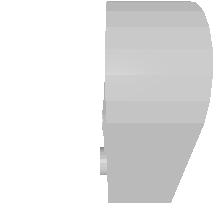}  &
\includegraphics[width = 0.09090909090909091\linewidth]{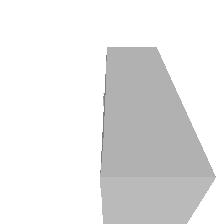}  &
\includegraphics[width = 0.09090909090909091\linewidth]{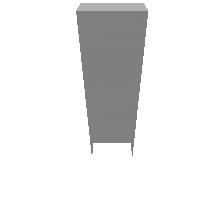}  &
\includegraphics[width = 0.09090909090909091\linewidth]{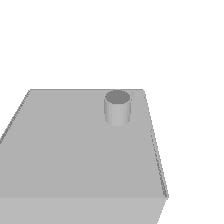}  &
\includegraphics[width = 0.09090909090909091\linewidth]{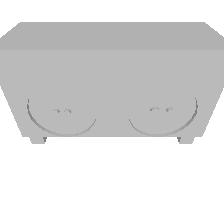}  &
\includegraphics[width = 0.09090909090909091\linewidth]{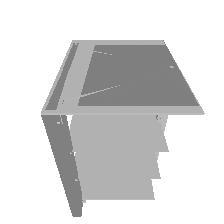}  \\

\includegraphics[width = 0.09090909090909091\linewidth]{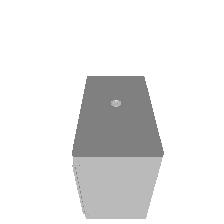} &
\includegraphics[width = 0.09090909090909091\linewidth]{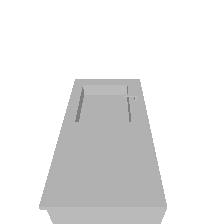}  &
\includegraphics[width = 0.09090909090909091\linewidth]{supimages/retrieval/6996.jpg}  &
\includegraphics[width = 0.09090909090909091\linewidth]{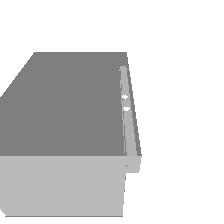}  &
\includegraphics[width = 0.09090909090909091\linewidth]{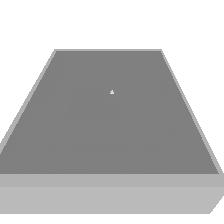}  &
\includegraphics[width = 0.09090909090909091\linewidth]{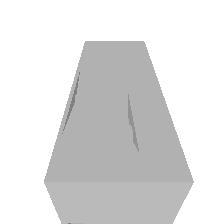}  &
\includegraphics[width = 0.09090909090909091\linewidth]{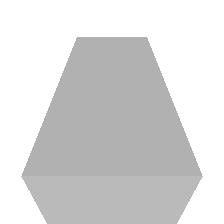}  &
\includegraphics[width = 0.09090909090909091\linewidth]{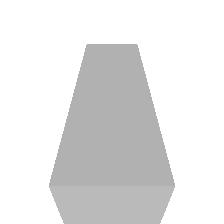}  &
\includegraphics[width = 0.09090909090909091\linewidth]{supimages/retrieval/4585.jpg}  &
\includegraphics[width = 0.09090909090909091\linewidth]{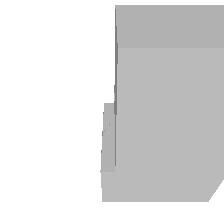}  &
\includegraphics[width = 0.09090909090909091\linewidth]{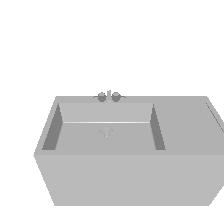}  \\

\includegraphics[width = 0.09090909090909091\linewidth]{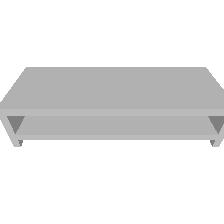} &
\includegraphics[width = 0.09090909090909091\linewidth]{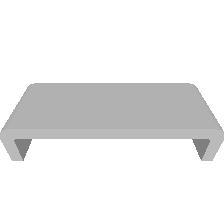}  &
\includegraphics[width = 0.09090909090909091\linewidth]{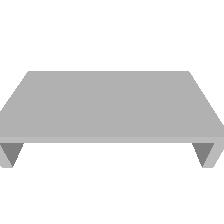}  &
\includegraphics[width = 0.09090909090909091\linewidth]{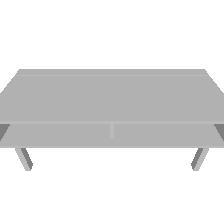}  &
\includegraphics[width = 0.09090909090909091\linewidth]{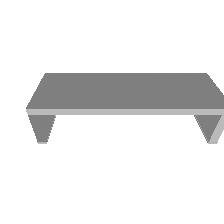}  &
\includegraphics[width = 0.09090909090909091\linewidth]{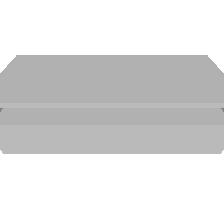}  &
\includegraphics[width = 0.09090909090909091\linewidth]{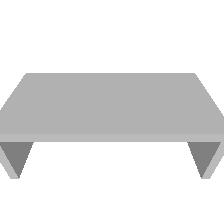}  &
\includegraphics[width = 0.09090909090909091\linewidth]{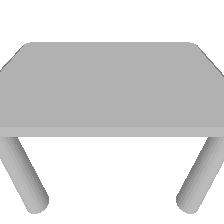}  &
\includegraphics[width = 0.09090909090909091\linewidth]{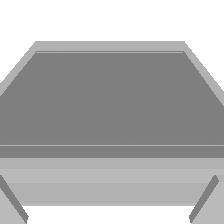}  &
\includegraphics[width = 0.09090909090909091\linewidth]{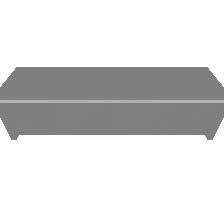}  &
\includegraphics[width = 0.09090909090909091\linewidth]{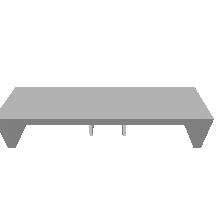}  \\

\includegraphics[width = 0.09090909090909091\linewidth]{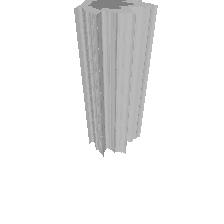} &
\includegraphics[width = 0.09090909090909091\linewidth]{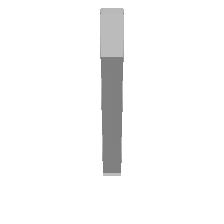}  &
\includegraphics[width = 0.09090909090909091\linewidth]{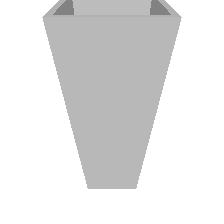}  &
\includegraphics[width = 0.09090909090909091\linewidth]{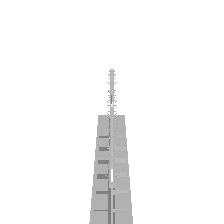}  &
\includegraphics[width = 0.09090909090909091\linewidth]{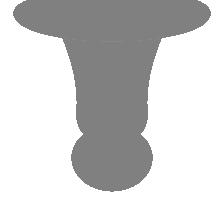}  &
\includegraphics[width = 0.09090909090909091\linewidth]{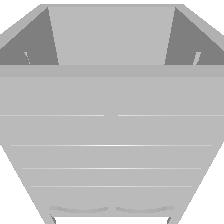}  &
\includegraphics[width = 0.09090909090909091\linewidth]{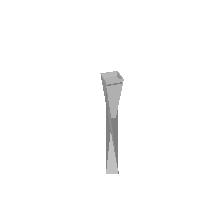}  &
\includegraphics[width = 0.09090909090909091\linewidth]{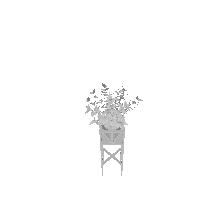}  &
\includegraphics[width = 0.09090909090909091\linewidth]{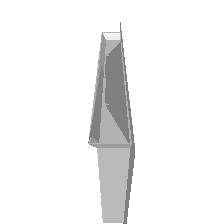}  &
\includegraphics[width = 0.09090909090909091\linewidth]{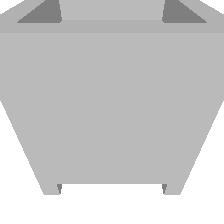}  &
\includegraphics[width = 0.09090909090909091\linewidth]{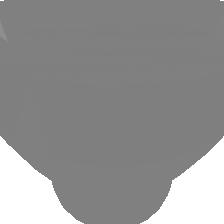}  \\ \hline
\end{tabular}
}
\vspace{2pt}
\caption{\small \textbf{Qualitative Examples for Object Retrieval}: \textit{(left):} we show some query objects from the test set. \textit{(right)}: we show top ten retrieved objects by our MVTN from the training set.
}
    \label{fig:imgs-retr-sup}
\end{figure*}

We show qualitative examples of our retrieval results using the MVTN-spherical with ViewGCN in \figLabel{\ref{fig:imgs-retr-sup}}. Note that the top ten retrieved objects for all these queries are positive (from the same classes of the queries). %

\end{document}